\documentclass[final,times,nonatbib]{elsarticle}
\usepackage[margin=2.35cm]{geometry}

\usepackage[T1]{fontenc}
\usepackage[english]{babel}
\usepackage{amsmath,amsfonts,amssymb}
\usepackage{graphicx}
\usepackage{booktabs}   %
\usepackage{multirow}
\usepackage{makecell}
\usepackage[export]{adjustbox}
\usepackage{placeins}
\usepackage{xcolor}
\usepackage{siunitx}
\usepackage{microtype}

\usepackage{acronym}

\usepackage{hyperref}
\hypersetup{
    colorlinks=true,
    allcolors={blue},
    pdftitle={Consensus Capture in Reference-Based UAV Video Stabilization: A Benchmark on Dynamic Urban Scenes},
    pdfauthor={Robert Fonod; Nikolas Geroliminis},
        pdfkeywords={UAV Video Stabilization; Image Registration; Homography Estimation;
                 Feature Matching; Oriented Bounding Box; Traffic Monitoring}
}

\usepackage[backend=biber,natbib=true,style=authoryear,sorting=ynt,
            maxcitenames=2,mincitenames=1,giveninits=true,dashed=false,
            uniquelist=false,sortcites=true]{biblatex}
\renewbibmacro*{begentry}{\ifentrytype{dataset}{\mkbibbrackets{dataset}\addspace}{}}

\usepackage{cleveref}
\crefname{figure}{Fig.}{Figs.}
\Crefname{figure}{Fig.}{Figs.}
\crefname{equation}{Eq.}{Eqs.}
\Crefname{equation}{Eq.}{Eqs.}
\crefname{table}{Table}{Tables}
\Crefname{table}{Table}{Tables}
\crefname{section}{Section}{Sections}
\Crefname{section}{Section}{Sections}
\crefname{appendix}{Appendix}{Appendices}
\Crefname{appendix}{Appendix}{Appendices}
\crefformat{appendix}{#2#1#3}
\Crefformat{appendix}{#2#1#3}
\crefmultiformat{appendix}{#2#1#3}{ and~#2#1#3}{, #2#1#3}{ and~#2#1#3}
\Crefmultiformat{appendix}{#2#1#3}{ and~#2#1#3}{, #2#1#3}{ and~#2#1#3}

\usepackage{pifont}
\newcommand{\cmark}{\ding{51}}
\newcommand{\xmark}{\ding{55}}

\graphicspath{{figures/}{diagrams/}}

\acrodef{A-KAZE}{accelerated-KAZE}
\acrodef{BB}{bounding box}
\acrodefplural{BB}[BBs]{bounding boxes}
\acrodef{BEV}{bird's-eye view}
\acrodef{BF}{brute force}
\acrodef{BRISK}{binary robust invariant scalable keypoints}
\acrodef{CLAHE}{contrast limited adaptive histogram equalization}
\acrodef{CTE}{corner transfer error}
\acrodef{CV}{computer vision}
\acrodef{DL}{deep learning}
\acrodef{FLANN}{fast library for approximate nearest neighbors}
\acrodef{FoV}{field of view}
\acrodef{GCP}{ground control point}
\acrodefplural{GCP}[GCPs]{ground control points}
\acrodef{GSD}{ground sample distance}
\acrodef{GT}{ground truth}
\acrodef{HBB}{horizontal bounding box}
\acrodefplural{HBB}[HBBs]{horizontal bounding boxes}
\acrodef{HEA}{homography estimation accuracy}
\acrodef{IoU}{intersection over union}
\acrodef{ITS}{intelligent transportation systems}
\acrodef{mAA}{mean average accuracy}
\acrodef{MI}{mutual information}
\acrodef{OBB}{oriented bounding box}
\acrodefplural{OBB}[OBBs]{oriented bounding boxes}
\acrodef{OCE}{object center error}
\acrodef{ORB}{oriented FAST and rotated BRIEF}
\acrodef{RANSAC}{random sample consensus}
\acrodef{RoI}{region of interest}
\acrodef{RSIFT}{root SIFT}
\acrodef{SAM}{Segment Anything Model}
\acrodef{SfM}{structure from motion}
\acrodef{SIFT}{scale-invariant feature transform}
\acrodef{SLAM}{simultaneous localization and mapping}
\acrodef{SNN}{second nearest neighbors}
\acrodef{SOTA}{state-of-the-art}
\acrodef{SSIM}{structural similarity}
\acrodef{UAV}{unmanned aerial vehicle}
\acrodefplural{UAV}[UAVs]{unmanned aerial vehicles}
\acrodef{UHD}{ultra-high-definition}
\acrodef{ZNCC}{zero-mean normalized cross-correlation}

\newcommand{\pixdom}{\Omega}                                 %
\newcommand{\Homog}{\mathbb{H}}                              %
\newcommand{\Fk}{F^{k}}                                      %
\newcommand{\Fref}{F^{\mathrm{ref}}}                         %
\newcommand{\Href}{\mathbf{H}^{k \rightarrow \mathrm{ref}}}           %
\newcommand{\Hest}{\widehat{\mathbf{H}}^{k \rightarrow \mathrm{ref}}} %
\newcommand{\Horacle}{\widetilde{\mathbf{H}}^{k \rightarrow \mathrm{ref}}} %
\newcommand{\Hrefinv}{\mathbf{H}^{\mathrm{ref} \rightarrow k}}
\newcommand{\Hestinv}{\widehat{\mathbf{H}}^{\mathrm{ref} \rightarrow k}}
\newcommand{\Horacleinv}{\widetilde{\mathbf{H}}^{\mathrm{ref} \rightarrow k}}
\newcommand{\Hgcp}{\mathbf{H}^{k \rightarrow \mathrm{ref}}_{\mathrm{GCP}}}
\newcommand{\fail}{\bot}                                     %
\newcommand{\eps}{\varepsilon}
\newcommand{\Reg}{\mathcal{R}}                               %
\newcommand{\Valid}{\mathcal{V}}                             %
\newcommand{\Veh}{\mathcal{M}}                               %
\newcommand{\Ground}{\mathcal{G}}                            %
\newcommand{\Boxes}{\mathcal{B}}                             %
\newcommand{\Pipe}{\mathcal{P}}                              %
\newcommand{\Corners}{\mathcal{C}}                           %
\newcommand{\apply}[2]{#1 \cdot #2}
\DeclareMathOperator{\dehom}{\pi}
\DeclareMathOperator{\CTEop}{CTE}
\DeclareMathOperator{\HEAop}{HEA}
\DeclareMathOperator{\IoUop}{IoU}
\DeclareMathOperator{\OCEop}{OCE}

\DeclareMathOperator{\ZNCCop}{ZNCC}
\DeclareMathOperator{\MIop}{MI}
\DeclareMathOperator*{\argmin}{arg\,min}

\newcommand{\suppfile}[1]{{\appto\UrlBreaks{\do\_}\nolinkurl{#1.csv}}}
\makeatletter
\def\ps@pprintTitle{%
 \let\@oddhead\@empty
 \let\@evenhead\@empty
 \let\@oddfoot\@empty
 \let\@evenfoot\@empty
}
\makeatother

\begin{document}

\begin{frontmatter}

\title{Consensus Capture in Reference-Based UAV Video Stabilization:
       A Benchmark on Dynamic Urban Scenes}

\author[epfl]{Robert Fonod\corref{cor1}}
\ead{robert.fonod@ieee.org}
\author[epfl]{Nikolas Geroliminis}

\affiliation[epfl]{organization={\'Ecole Polytechnique F\'ed\'erale de Lausanne (EPFL),
                                 Urban Transport Systems Laboratory (LUTS)},
                   addressline={Station 18},
                   city={CH-1015 Lausanne},
                   country={Switzerland}}

\cortext[cor1]{Corresponding author.}

\begin{abstract}
Reference-frame stabilization registers every frame of a hovering-drone video onto one reference
frame, keeping object positions comparable. The registration pipelines it relies on are selected
on feature-matching benchmarks in which only the camera moves. Over a busy intersection, vehicles
moving together supply mutually consistent correspondences that can draw the robust homography
estimate toward their motion: consensus capture, which inlier statistics cannot flag. A two-part
benchmark measures it and the object masking meant to prevent it. A static protocol
scores ten pipelines under exactly known homographies on 4K aerial frames, tuned on a
site-disjoint split at a matched resolution and feature budget. A dynamic protocol scores them at an offline and a
real-time budget on hovers selected from the video alone to span the amount and
coherence of moving foreground, plus a stationary control. It measures background alignment, object
position error and the share of a moving group's displacement an estimate follows, against a
pseudo-ground-truth reference audited with manually annotated ground control points, which also
score every candidate. On static scenes, tuned AKAZE, RootSIFT, SIFT and BRISK lead, retuning the
matcher thresholds raises the mean average accuracy (mAA) by at most \num{0.0062}, and masking costs
less than \num{0.005}. Unmasked ORB is captured on \num{28} of \num{304} real-time dynamic pairs
and on \num{6} offline; its real-time captured estimates have a median corner error
of \SI{18.8}{px} yet typical inlier counts and ratios. Masking removes capture for every pipeline and
leaves the leading pipelines unchanged. Code, data and results are released.
\end{abstract}


\end{frontmatter}

\section{Introduction}
\label{sec:introduction}

Reference-frame stabilization is the registration model used when object positions must stay
consistent over a long hover, during which wind keeps shifting and rotating the view: every frame is aligned directly onto one
fixed reference frame, so that error does not accumulate and a position measured in frame $k$ is
comparable with one measured minutes earlier. Trajectory extraction from high-altitude
\ac{UAV} video is the application studied here, and traffic monitoring supplies the footage, but
the requirement is shared by any analysis that compares moving objects across time in image or map
coordinates, such as river surface velocimetry~\citep{ljubicic2021comparison}. The estimate is a homography, and the pipeline that produces it (feature detector,
descriptor, matcher, robust estimator, and, where objects are annotated, a masking strategy that
withholds them from feature detection) is assembled from components each of which has been
benchmarked extensively~\citep{jin2021image,efe2021effect,barath2023homography}.

\begin{figure}[t]\centering
  \includegraphics[width=\textwidth]{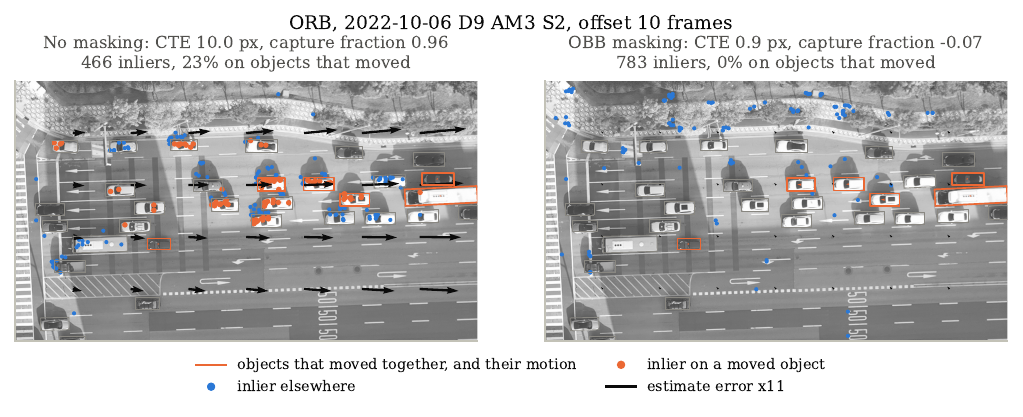}
  \caption{Consensus capture on real footage. Unmasked ORB at the real-time budget (left) registers
  a frame ten frames (\SI{0.33}{s}) after the reference frame of case S2 (\Cref{tab:survey-cases}).
  Both panels show the same zoomed-in crop of the $3840 \times 2160$ frame, centered on the largest
  moving group; orange boxes
  are the vehicles that moved together between the two frames, with their motion. The estimate
  registers the group back onto where it stood (capture fraction \num{0.96}), so its error, drawn
  as black arrows on a grid and magnified by the factor in the legend, carries the platoon's motion
  across the whole crop (\SI{10.0}{px} corner error against the reference estimator), while \SI{23}{\percent} of its \num{466} inliers
  lie on the moving vehicles (orange dots). The same pipeline under oriented-box masks (right)
  keeps \num{783} inliers, none on those vehicles, at a corner error of \SI{0.9}{px}. Both are re-registered
  with the stored parameters and seed of the real-time campaign (\Cref{sec:capture-results}).}
  \label{fig:teaser}
\end{figure}

Those benchmarks evaluate correspondence accuracy on scenes without independently moving content:
planar surfaces, landmark facades, building walls. The only source of disagreement between their
two views is the camera. A hovering drone over a busy intersection presents the opposite case. The
camera barely moves, and dozens of objects in the frame move without it, many of them together: a
platoon released by a green light translates as a unit, and every vehicle in it supplies
correspondences that agree with one another under a single transform, the one describing their
motion. If those correspondences win the consensus, the robust estimator does not fail. It solves
the problem it was posed, returning a homography that is internally consistent and well supported
but describes the vehicles rather than the road (\Cref{fig:teaser}).

We call this failure \emph{consensus capture}, and it has a consequence for evaluation before it
has one for method design. Inlier count, inlier ratio, and reprojection error measure the
self-consistency of a fit, and a coherently moving foreground is self-consistent by construction.
Where it wins, these quantities are no longer merely weak proxies for accuracy, as they are already
known to be on static scenes~\citep{jin2021image,barath2023homography}; they become
anti-informative, because the moving-object consensus can be the cleaner of the two available
fits. An evaluation that reads them cannot detect capture. What is needed scores the alignment of
the static world directly, makes a moving-object consensus observable, and relates it to how much
of the frame moves and how coherently. To our knowledge no matching or registration benchmark provides
this: none combines
independently moving content with reference geometry (\Cref{sec:related}), so a risk flagged for
airborne video~\citep{kirchhof2006detection} has gone unmeasured.

Capture matters most where trajectories are used one at a time. Macroscopic quantities such as flow
and density average a transient offset away, but microscopic analyses read individual trajectories
and their derivatives: car following, lane changes~\citep{kim2026lane}, per-vehicle noise
emission~\citep{espadaler2023noise}, and surrogate safety measures such as
time-to-collision, post-encroachment time and the deceleration rate to avoid a
crash~\citep{abdelaty2023advances,espadaler2025swarm}, which drone trajectory
datasets now provide or are built to support~\citep{zheng2024citysim, FONOD2025105205}.
Under reference-frame registration each frame is registered on its own, so nothing ties a captured estimate to those of neighboring frames: its error
follows the group's motion and is present only while the group holds the consensus. Speeds and
accelerations are differences of positions over fractions of a second, so an error that varies from
frame to frame in this way is amplified in them~\citep{punzo2011assessment}. The accuracy these
datasets aim for is correspondingly tight: the inD dataset requires a positioning error below
\SI{0.1}{\meter}~\citep{bock2020ind}, under two pixels at the \ac{GSD} of the footage studied here
(\Cref{sec:problem}), whereas the captured estimates of \Cref{sec:capture-results} have a median
corner error of \SI{18.8}{px}, about a meter on the ground, against the reference estimator. Since
the estimates pulled toward the group keep typical inlier statistics, a pipeline captured on a few
frames passes their error, unflagged, to every trajectory in those frames. Several of these datasets are also recorded lower
than the \SIrange{140}{150}{\meter} studied here: inD from up to \SI{100}{\meter} at about
\SI{4}{\centi\meter} per pixel~\citep{bock2020ind} and the intersections of CitySim from
\SI{120}{\meter}~\citep{zheng2024citysim}; a view that covers less ground can give a vehicle group
of the same size a larger share of the frame and of its features.

We provide such an evaluation in two parts. Part~I is a static protocol under exactly known
homographies and photometric degradation, on 4K frames from a multi-drone experiment over Songdo,
South Korea: one from each of the \num{29} viewpoints from which its ten drones recorded \num{20}
intersections. It establishes what each pipeline can do when nothing moves and which of its
settings matter, extending the campaign of \citet{FONOD2025105205} with paired trials,
per-pipeline tuning on a site-disjoint split, and a matched resolution and feature budget. Part~II is a dynamic
protocol on nine hovers over busy intersections, each registered onto a reference frame selected
from the video alone to span the amount and coherence of
moving foreground, plus a stationary control that separates the removal of moving objects from the removal
of texture. It runs every pipeline at an offline and a real-time point of Part~I's budget grid.
Real footage has no true homography, so Part~II measures background alignment outside the annotated
objects, the position error a trajectory inherits, the direction of the error relative to the
dominant motion, and the share of a moving group's displacement an estimate follows, which
measures capture where it happens. Its pseudo-ground-truth reference estimator is audited against
manually annotated \acp{GCP} (image positions of static ground points, not surveyed coordinates),
which also score every candidate directly, so that the
masking comparison does not depend on the reference estimator's own masking. Masking with
\acp{HBB} and \acp{OBB} is a controlled axis throughout, since the shape of a mask could
matter in either direction (\Cref{sec:masking}).

On hovers over Songdo intersections, the measurement finds capture. At the real-time budget,
unmasked ORB is captured in all five cases that present a coherently moving group, the four with
much moving foreground and the stationary control, on \num{28} of the \num{304} pairs that carry
one, though \num{22} of them in a single case. The \num{17} within the first second are pulls toward
the group; \num{10} of the \num{11} beyond it are breakdowns of the estimate, which unmasked ORB also
shows with no group in view. At the offline budget capture is rarer (unmasked ORB, \num{6} pairs in
two cases). Estimates pulled toward a group cannot be told apart by their inlier statistics: their
inlier counts and ratios match those of the same pipeline at the nearest offsets of the same case,
at both budgets. Masking removes capture for every pipeline at both budgets while leaving
the leading pipelines, SIFT, RootSIFT and AKAZE, where they were; for real-time ORB, whose feature
budget it also returns to the ground, the captured pairs can account for at most a quarter of the gain. \acp{HBB}
and \acp{OBB} do not differ at these densities and orientations, where the two geometries
withhold nearly the same pixels.
Averaged over all nine cases, four of which never present a moving group, the masking benefit is
either not significant or smaller than the repeatability of the \acp{GCP}
(\Cref{sec:masking-results}). The conclusions are bounded by the
moving share of these cases; the protocol is not, and it applies unchanged to footage from lower
altitude or over denser traffic, where that share is larger.

\subsection{Contributions}
\label{sec:contributions}

\begin{enumerate}
  \item A benchmark for the complete registration pipeline of reference-based UAV stabilization in
        dynamic urban scenes: a static protocol under exactly known homographies
        (\Cref{sec:static}) and a dynamic protocol on real hovering-drone footage at an offline and a
        real-time compute budget (\Cref{sec:dynamic}). It covers the pipelines a deployed
        stabilization library exposes, as a measurement of that system rather than a leaderboard
        of state-of-the-art matchers (\Cref{sec:pipelines}).
  \item An evaluation methodology for dynamic scenes, where self-consistency statistics are
        unreliable: background-alignment measures that need no ground-truth homography,
        object-level errors that bound the position error a trajectory inherits, two measures that
        make consensus capture observable (the direction of the error and the share of a moving
        group's displacement an estimate follows), and a \ac{GCP} audit that measures the error of the
        pseudo-ground truth and supplies an independent comparison (\Cref{sec:metrics}), under a
        pre-specified analysis plan (\Cref{sec:analysis-plan}).
  \item A case-selection procedure that measures from the video alone how much of the foreground
        moves, how coherently, and when, and chooses the cases and their reference frames to span
        these conditions, under which capture can arise; its
        measurements are validated against published vehicle trajectories
        (\Cref{sec:dynamic-protocol}).
  \item A measurement of consensus capture and of object-aware masking on real footage: at the
        real-time budget unmasked ORB is captured in every case that presents a coherent group, its
        pull toward the group is invisible to inlier statistics, and capture disappears under
        masking at detection (\Cref{sec:capture-results,sec:masking-results}).
  \item The public release of the benchmark implementation~\citep{stabilo_benchmark} and of its
        data~\citep{stabilo_benchmark_data} (annotated frames, cases, \acp{GCP} and every
        campaign's results), from which every benchmark number is regenerated by a command. Also
        released are \texttt{hbb2obb}~\citep{hbb2obb}, the SAM-assisted conversion of \acp{HBB}
        to \acp{OBB} that makes \ac{OBB} masks available at benchmark scale, and Songdo
        Vision OBB~\citep{songdo_vision_obb}, the \ac{OBB} version of Songdo
        Vision~\citep{songdo_vision}, together with the human-verified oriented reference against
        which the conversion is validated (\Cref{sec:songdo-vision-obb,sec:appendix-obb}).
\end{enumerate}

\section{Related work}
\label{sec:related}

\paragraph{Local features and matching}
Correspondence estimation has been surveyed repeatedly as it moved from hand-crafted operators to
learned ones~\citep{ma2021survey}. The pipelines
benchmarked here span three generations of that work. Hand-crafted detectors and descriptors, from
SIFT~\citep{lowe2004sift} and its RootSIFT normalization~\citep{arandjelovic2012rootsift} to the
binary ORB~\citep{rublee2011orb} and BRISK~\citep{leutenegger2011brisk} and the nonlinear
scale spaces of KAZE~\citep{alcantarilla2012kaze} and AKAZE~\citep{alcantarilla2013akaze}, remain
the default in deployed stabilization pipelines, because they need no training data, no
accelerator, and no assumption about the imagery. Learned detectors and descriptors replace one or
both stages with a network trained for repeatability and matchability, among them
SuperPoint~\citep{detone2018superpoint}, DISK~\citep{tyszkiewicz2020disk},
DeDoDe~\citep{edstedt2024dedode}, Key.Net~\citep{barroso2019keynet} and
XFeat~\citep{potje2024xfeat}; learned matchers such as LightGlue~\citep{lindenberger2023lightglue}
replace the nearest-neighbor search and its ratio test with attention over the two keypoint sets;
and detector-free methods such as LoFTR~\citep{sun2021loftr} drop keypoints and match dense features
directly. Dense matchers such as RoMa~\citep{edstedt2024roma} go further and estimate a warp and
its certainty for every pixel, from which correspondences are sampled.
A stabilization pipeline is chosen as a whole and its components interact (a matcher accepts only
certain descriptors, a detector-free method has no keypoint budget), so the comparison of
\Cref{sec:static-results} spans all three. Two properties of this literature shape the protocol.
Orientation estimation is not universal, so several learned models are effectively upright, which
is why rotation is swept, as \citet{tareen2018comparative} sweep it for the classical detectors,
which they find unequally invariant to it; and the input resolution of
a learned model is a memory bound rather than a free choice, which is why the compute budget is an
axis of the benchmark.

\paragraph{Direct homography regression}
A parallel line of work regresses the transform from an image
pair~\citep{detone2016homographynet}. \citet{zhang2020caudhn} observe that a loss computed uniformly
over the image cannot keep moving or non-planar objects from contributing to it, and learn a
content mask that rejects those regions: the problem measured here, approached by learning what to
ignore instead of by annotating it. These methods lie outside the feature-based family deployed in the trajectory extraction systems of
\Cref{sec:related-traffic}.

\paragraph{Robust estimation and dominant structure}
RANSAC~\citep{fischler1981ransac} and its modern variants~\citep{barath2020magsac,barath2018graphcut}
make a homography recoverable from contaminated correspondences. That a coherent structure of
outliers can capture the consensus is known~\citep{stewart1997bias}, and \citet{chum2005degensac}
detect a dominant plane inside the estimator; \Cref{sec:why-not-inliers} explains why that defense
does not transfer to a coherently moving group. Filters that prune matches before the estimator rest
on a similar vote: \citet{jiang2018hierarchical} keep matches whose motion agrees in direction and
length, \citet{zhou2024unified} build local motion consensus into a learned matcher, and the motions
of a coherently moving group agree as well.

\paragraph{Scenes with more than one structure}
Multi-model fitting accepts that a scene may contain several coherent structures and recovers them
jointly, by clustering correspondences~\citep{toldo2008jlinkage,wong2011adelaide} or by fitting
instances progressively inside the robust estimator~\citep{barath2019progressivex}.
Vehicles crossing an intersection are such a
second structure, and AdelaideRMF~\citep{wong2011adelaide} does contain independently moving
content. These methods are evaluated by how well correspondences are partitioned among structures,
not by the accuracy of one recovered transform, and AdelaideRMF supplies no reference homography for
the background. Recovering several homographies also leaves open which one describes the ground,
which correspondence counts cannot decide when the moving structure is the larger. Multi-model
fitting is a remedy that changes the estimator; what is missing first is a measurement of the
failure under the estimators actually deployed.

\paragraph{Benchmarks for matching and homography estimation}
Comparative studies of classical detectors and descriptors~\citep{tareen2018comparative} and
HPatches~\citep{balntas2017hpatches} established the vocabulary this field still uses, and
\citet{schonberger2017comparative} set the pattern of judging hand-crafted and learned features on a
common downstream task. Three benchmarks set the
methodological standard we adopt. \citet{jin2021image} argue that local features must be judged by
a downstream task rather than by intermediate proxies, and show that carefully tuned classical
pipelines remain competitive with learned ones, a conclusion \citet{efe2021effect} reach on HPatches
by optimizing a single parameter; \citet{barath2023homography} contribute a large-scale homography
benchmark with an explicit tuning-and-test protocol. Their shared lesson is that a comparison is
only as meaningful as the care taken to place every method on equal footing
(\Cref{sec:comparability}). None of them measures registration accuracy on a scene whose content
moves independently of the camera (\Cref{tab:benchmarks}), because their reference geometry,
\ac{SfM} or manually labeled correspondences on static structure, must exclude independently moving
objects to work at all. Registration benchmarks that do model change place it between
acquisitions~\citep{sun2025nothing}. Pseudo-ground truth has a further cost: \citet{brachmann2021limits} show
that the algorithm generating it can reorder the methods evaluated against it, which is why the
reference estimator of Part~II is audited against \acp{GCP} (\Cref{sec:gcp-audit}).

\begin{table}[htbp]
  \centering
  \caption{Benchmarks for feature matching, homography estimation and stabilization, and the property
           this work adds. ``Independent motion'' asks whether any part of the scene moves between the two
           views independently of the camera.}
  \label{tab:benchmarks}
  \small
  \begin{tabular}{@{}llll@{}}
    \toprule
    Benchmark & Content & Reference geometry & Independent motion \\
    \midrule
    HPatches~\citep{balntas2017hpatches} & planar surfaces          & manual homography     & \xmark \\
    \citet{efe2021effect}                & planar surfaces          & manual homography     & \xmark \\
    IMC-PT~\citep{jin2021image}          & landmark photos          & SfM poses       & \xmark \\
    HEB~\citep{barath2023homography}     & landmark photos          & SfM planes      & \xmark \\
    AdelaideRMF~\citep{wong2011adelaide} & multi-structure pairs    & none (memberships)    & \cmark \\
    \citet{ljubicic2021comparison}       & river videos             & none (static features) & \cmark \\
    \midrule
    This work, Part~I                    & UAV images, intersections & known by construction & \xmark \\
    This work, Part~II                   & UAV videos, intersections & audited pseudo-GT     & \cmark \\
    \bottomrule
  \end{tabular}
\end{table}

\paragraph{Suppressing moving objects before geometry is estimated}
Excluding independently moving content before a geometric model is fitted is established practice
in dynamic \ac{SLAM}. DynaSLAM~\citep{bescos2018dynaslam} and DS-SLAM~\citep{yu2018dsslam} segment
likely-moving classes and withhold those regions from tracking and mapping. Both report that suppression keeps
the estimate usable in populated scenes; what is missing is a measurement. Those systems are
evaluated end to end, by trajectory error on indoor sequences with a handful of walking people or
on driving sequences, which does not say how much of the frame the objects may cover before the consensus shifts to them, which box
geometry is worth its cost, or what the failure looks like without suppression. A hovering drone
over an intersection is the opposite regime: a nearly static camera and dozens of vehicles moving
in coherent streams. Isolating the effect needs the masking strategy varied as a controlled axis
while everything else is held fixed (\Cref{sec:masking-results}), and a metric that does not read
the objects themselves (\Cref{sec:photometric-metrics}).

\paragraph{Video stabilization as a different objective}
Stabilization also names work whose goal is a pleasant camera path rather than a fixed frame of
reference: smoothing the path under geometric constraints~\citep{grundmann2011l1}, over a mesh of
local homographies~\citep{liu2013bundled}, or by learning the warp~\citep{yu2020videostab}. Its
measures (cropping ratio,
distortion, a frequency-domain stability score) follow from a perceptual objective, and none asks
where a pixel actually is. A smoothed path may drift from any particular frame, which is
inadmissible when object positions are compared across a recording, and a mesh of local warps can
absorb independent object motion instead of rejecting it. Moving objects enter this literature as a
threat to the global motion estimate, met by adaptive RANSAC~\citep{choi2009robust}, clustered motion
vectors~\citep{wu2021robust} or foreground trajectories~\citep{zhao2019robust}, and scored by the
similarity of the output frames, not against reference geometry.

\paragraph{Registration and stabilization of aerial video}
Registering airborne video to a fixed reference is long established in
photogrammetry~\citep{kumar2001aerial}. The remote sensing literature on image registration
concentrates on matching across modalities, sensors and acquisition times~\citep{geng2025feature},
with structural or phase-congruency descriptors built to be insensitive to radiometric
change~\citep{ye2017hopc,li2020rift} and networks trained on large multimodal
datasets~\citep{wu2026mapglue}, so transient content is a nuisance to be removed rather than a
property to be measured.

\paragraph{Moving traffic in airborne registration}
Vehicle measurement from the air rests on this
registration~\citep{reinartz2006traffic,kariminejadasl2006optical}, and moving traffic is a recognized
hazard to it. \citet{kirchhof2006detection} warn that many objects moving in the same direction, as on
dense roads, can survive RANSAC and bias the homography, and exclude points at detected motion.
\citet{kariminejadasl2014sequential} rule out direct SIFT with RANSAC against the first image of a
helicopter sequence, in part because of the moving vehicles, and chain consecutive homographies
instead. \citet{ljubicic2021comparison} compare seven stabilization tools on \ac{UAV} river footage,
where the water moves and the banks do not, by residual motion at static features. None of these
measures capture or what it costs in position.

\paragraph{Aerial detection benchmarks and trajectory datasets}
Aerial object detection has well-established
benchmarks~\citep{ding2022dota,zhu2022visdrone}, DOTA made \acp{OBB} standard for
aerial objects, and UAVid~\citep{lyu2020uavid} separates moving from static cars in 4K \ac{UAV} video
for semantic segmentation. They evaluate \emph{recognition}, not the registration that must precede
any use of object positions when the camera moves. Drone-based trajectory datasets show the dependency most
clearly: pNEUMA~\citep{barmpounakis2020pneuma}, inD~\citep{bock2020ind},
CitySim~\citep{zheng2024citysim} and Songdo Traffic~\citep{FONOD2025105205} all extract positions from
hovering footage after a registration step, whose error georeferenced
positions~\citep{novikov2024vehicle} inherit directly, and the intersections they are recorded over are where a
moving-object consensus is most likely.

\paragraph{UAV stabilization for traffic monitoring}
\label{sec:related-traffic}
The work closest to this paper is \citet{FONOD2025105205}, which extracts georeferenced vehicle
trajectories from hovering-drone recordings of urban intersections and stabilizes each frame onto a
reference frame with an object-masked homography, releasing that stage as
\texttt{stabilo}~\citep{stabilo} and its parameter study as
\texttt{stabilo\nobreakdash-optimize}. The vehicles are masked to keep
correspondences on moving objects out of the fit, the failure this paper names consensus capture,
but the pipeline was selected on static scenes under synthetic warps, so neither that failure nor
what masking does about it was measured on the recordings. That is the gap addressed here: the static campaign is
extended (\Cref{sec:static}) and paired with a dynamic one on the footage the pipeline is deployed
on (\Cref{sec:dynamic}), where the masking axis can be measured rather than assumed.

\section{Problem setting}
\label{sec:problem}

\paragraph{Scene and platform assumptions}
Recordings come from hovering drones at approximately constant altitude
(\SIrange{140}{150}{\meter}) in \ac{BEV} at 4K resolution ($3840 \times 2160$), so inter-frame
motion is small and wind-induced. The \acf{GSD} is about \SI{5.5}{\centi\meter} per pixel: the
scale of the homographies that register hovers of this experiment onto its georeferenced
orthophoto~\citep{FONOD2025105205} ranges from \num{0.0548} to \SI{0.0564}{\meter} per pixel over
eleven hovers. A pixel of registration error is thus about \SI{5.5}{\centi\meter} on the ground,
and the \SI{5}{px} upper limit of the accuracy interval below is about \SI{28}{\centi\meter}.
Every pixel quantity in this paper converts at that rate.
The stabilization model is a reference-frame homography, which holds the static scene in place
over a hover exactly only where that scene is planar. Above-ground structures (buildings, poles,
vegetation) depart from the plane in proportion to how far the platform translates; a hovering
platform translates little but does drift, so the evaluation excludes buildings and large tree
canopy, the largest such structures, by polygons drawn on each reference frame (\Cref{sec:photometric-metrics}) and scores
the rest of the static scene. The deployed regime is a
hovering platform, so the geometric perturbation a pipeline has to recover in practice is a few
pixels of drift and a fraction of a degree of yaw per frame relative to the reference, growing
slowly over a recording. Part~I samples a wider regime than that, so that a
pipeline's operating limits are measured, and \Cref{sec:static}
states where the two regimes differ.

\subsection{Images, transforms, and the registration model}
\label{sec:notation}

Let $\pixdom = \{0,\dots,W-1\} \times \{0,\dots,H-1\}$ be the pixel domain of a recording and let
$\Fk : \pixdom \to \mathbb{R}$ denote its $k$-th frame, with $\Fref$ a designated \emph{reference
frame} (one fixed frame per case, chosen by the case selection of \Cref{sec:dynamic-protocol}). Write
$\dehom(x_1,x_2,x_3) = (x_1/x_3,\, x_2/x_3)$ for the projection from homogeneous to Euclidean coordinates, and let
$\Homog$ be the group of planar homographies, that is, the real projective linear group
$\mathrm{PGL}(3,\mathbb{R})$ of non-singular $3 \times 3$ matrices identified up to a non-zero
scalar~\citep{hartley2004multiple}. \Cref{tab:notation} collects the notation. A homography $\mathbf{H} \in \Homog$ acts on a
point $\mathbf{x} \in \mathbb{R}^2$ by
\begin{equation}
  \apply{\mathbf{H}}{\mathbf{x}} \;=\; \dehom\!\left( \mathbf{H} \begin{bmatrix} \mathbf{x} \\ 1 \end{bmatrix} \right),
  \label{eq:action}
\end{equation}
which is defined whenever the third homogeneous coordinate is non-zero. A degenerate or
non-finite estimate drives that coordinate to zero, and \Cref{sec:failures} explains why the
resulting non-finite coordinates are propagated rather than suppressed.

\paragraph{Direct-to-reference registration}
Each frame $\Fk$ is aligned to $\Fref$ by a homography $\Href \in \Homog$, estimated
\emph{directly} for every $k$; the registered frame samples $\Fk$ through its inverse:
\begin{equation}
  \Fref(\mathbf{x}) \;\approx\; \Fk\!\left( \apply{\Hrefinv}{\mathbf{x}} \right), \qquad \mathbf{x} \in \pixdom .
  \label{eq:registration}
\end{equation}
Sequential accumulation, $\Href = \prod_{j=1}^{k} H^{j \to j-1}$, is \emph{not} used. Its error grows
with $k$ by construction, and it is not the algorithm deployed in reference-frame
stabilization~\citep{FONOD2025105205}. \Cref{eq:registration} rests on the ground-plane assumption, which
above-ground structure violates.

\subsection{Pipelines under study}
\label{sec:pipelines}

A \emph{pipeline} is the complete chain that turns an image pair into a transform. We write it as a
tuple
\begin{equation}
  \Pipe \;=\; (\, \mathrm{det},\ \mathrm{desc},\ \mathrm{match},\ \mathrm{rob},\ \mu \,),
  \label{eq:pipeline}
\end{equation}
whose components are, in order, the feature detector, the descriptor, the matcher, the robust estimator, and
the masking strategy $\mu \in \{\textsf{none}, \textsf{hbb}, \textsf{obb}\}$ of
\Cref{sec:masking}. Applied to a pair it yields
\begin{equation}
  \Pipe\!\left( \Fref,\, \Fk,\, \Boxes^{\mathrm{ref}},\, \Boxes^{k} \right)
    \;\in\; \Homog \cup \{\fail\},
  \label{eq:pipeline-map}
\end{equation}
where $\Boxes^{\mathrm{ref}}$ and $\Boxes^{k}$ are the object annotations of the reference frame and
of frame $k$ (\Cref{sec:masking}) and $\fail$ denotes
a failure to produce an estimate at all. Admitting $\fail$ explicitly, rather than substituting the
identity, is what makes the failure rate a reported quantity instead of a silent bias
(\Cref{sec:failures}). We write $\Hest$ for a successful estimate.

The five axes of \Cref{eq:pipeline} are evaluated \emph{jointly}. Component-wise evaluation is the
norm in the matching literature, but the quantity a stabilization system delivers is the transform,
and the interactions between the axes, particularly between $\mu$ and $\mathrm{rob}$, are what
this benchmark is designed to expose.

The components come from two libraries: the classical detectors and descriptors and every robust
estimator are OpenCV's, and the learned detectors, descriptors and matchers are
kornia's~\citep{riba2020kornia}. Both are driven through
\texttt{stabilo}~\citep{stabilo}, the stabilization library of \citet{FONOD2025105205}, which
this benchmark evaluates and does not modify. Which implementation supplies a component is part of
what is compared, for the reason given below.

\paragraph{Selection of pipelines}
The detector axis holds ten pipelines. Five are classical, matched by brute force with a ratio
test: ORB~\citep{rublee2011orb}, SIFT~\citep{lowe2004sift},
RootSIFT~\citep{arandjelovic2012rootsift}, BRISK~\citep{leutenegger2011brisk} and
AKAZE~\citep{alcantarilla2013akaze}. Four are learned and sparse: DISK~\citep{tyszkiewicz2020disk},
DeDoDe~\citep{edstedt2024dedode} and Key.Net~\citep{barroso2019keynet} (KeyNet in the tables),
each matched by LightGlue~\citep{lindenberger2023lightglue}, and XFeat~\citep{potje2024xfeat},
matched by brute force because the library's LightGlue has no weights for it. One is detector-free,
LoFTR~\citep{sun2021loftr}, which has no keypoint budget to set. The deployed library exposes one
more detector, KAZE~\citep{alcantarilla2012kaze}, which is left out: AKAZE is its accelerated
variant, and it is the slowest classical detector by a wide margin. Otherwise, the set is everything the library exposes,
which is the inclusion criterion: the object of study is the system a trajectory-extraction pipeline actually
runs and the alternatives a practitioner can select between by configuration alone, not the state
of the art in matching at large. Methods outside
that system, such as SuperPoint~\citep{detone2018superpoint} and the dense matchers of
\Cref{sec:related}, fall outside this scope; nothing in the protocol depends on the method, and it
applies to them unchanged.

\paragraph{Robust estimators are identified by implementation, not by name}
The $\mathrm{rob}$ axis sweeps nine OpenCV method codes spanning seven algorithms:
LMEDS~\citep{rousseeuw1984lmeds}, RANSAC~\citep{fischler1981ransac}, RHO~\citep{bazargani2018rho},
GC-RANSAC~\citep{barath2018graphcut}, PROSAC~\citep{chum2005prosac}, MAGSAC++~\citep{barath2020magsac},
and LO-RANSAC~\citep{chum2003loransac} in three configurations that differ in local-optimization
iterations and in whether they run in parallel, the last four within OpenCV's implementation of the
USAC framework~\citep{raguram2013usac}. They are reported under OpenCV's flag names with the
algorithm in parentheses, because the three LO-RANSAC configurations are separate entries a
practitioner selects from, and because two of the flags are widely described as DEGENSAC,
including in the documentation of the library evaluated here, although neither carries the
dominant-plane check of \citet{chum2005degensac} into a homography estimate
(\Cref{sec:why-not-inliers}). Following \citet{barath2023homography}, a method is fixed by its
implementation and version rather than by its published name (\Cref{sec:appendix-parameters}).

\paragraph{Masking acts on feature detection}
The order is: detection or segmentation $\rightarrow$ mask construction $\rightarrow$ feature
detection $\rightarrow$ description $\rightarrow$ matching $\rightarrow$ robust estimation. No
keypoint inside a mask reaches matching, so metrics defined on ``removed features'' or on a
``dynamic inlier ratio'' are not applicable here, and are not used. Where the mask enters the
detector differs between implementations, and it decides what masking can do under a feature
budget; every detector is used with the masking its implementation provides, unmodified. ORB,
BRISK and AKAZE detect outside the mask and then keep their strongest responses up to the budget,
so the keypoints a mask removes are replaced by others. SIFT, as OpenCV implements it, keeps its
strongest responses over the whole frame and then drops those inside the mask; the learned
detectors accept no mask, so the library drops their selected keypoints that fall inside it. For
both, masking can only remove keypoints.

\begin{table}[htbp]
  \centering
  \caption{Notation used throughout. Transform superscripts give the direction of the mapping, so
           $\Hrefinv = \bigl( \Href \bigr)^{-1}$.
           Matrices and vectors are bold and scalars italic, so the homography $\mathbf{H}$ is
           distinct from the frame height $H$; frame dimensions are $H \times W$ and object box
           extents $w \times h$, following the usual image and detection conventions.}
  \label{tab:notation}
  \begin{tabular}{@{}ll@{}}
    \toprule
    Symbol & Meaning \\
    \midrule
    $\pixdom$, $H \times W$          & pixel domain of a recording, and its size \\
    $\Fk$, $\Fref$                   & frame $k$; the reference frame \\
    $\Homog$                         & group of planar homographies, $\mathrm{PGL}(3,\mathbb{R})$ \\
    $\apply{\mathbf{H}}{\mathbf{x}}$, $\dehom$ & action of $\mathbf{H}$ on a Euclidean point; homogeneous projection \\
    $\Href$, $\Hrefinv$              & true transform current $\to$ reference, and its inverse \\
    $\Hest$                          & a candidate pipeline's estimate of $\Href$ \\
    $\Horacle$                       & the reference estimator (pseudo-ground truth) \\
    $\Hgcp$                          & transform fitted to the GCPs of an audited frame \\
    $\fail$                          & failure to produce an estimate \\
    $\Pipe$, $\mu$                   & a pipeline and its masking strategy \\
    $\Boxes^{k}$, $\mathbf{q}_j$, $\bar{\mathbf{q}}_j$ & object annotations of frame $k$; corners and center of box $j$ \\
    $\Corners$, $\mathbf{c}_i$, $\mathbf{d}_i$         & frame corners and their transfer displacements \\
    $\Reg$, $\Valid$, $\Veh$, $\Ground$ & evaluation region; its validity, object, and static-structure terms \\
    $\tau$, $\rho$                   & object dilation and border erosion, in pixels \\
    $\mathbf{u}$, $\bar\theta$, $\kappa$      & dominant motion axis, its angle, and its concentration \\
    $e_\parallel$, $e_\perp$, $A$    & error along and across that axis; their alignment ratio \\
    $\eps$, $\eps_{\mathrm{floor}}$  & HEA threshold; GCP-audited equivalence margin \\
    $\delta$                         & the equivalence margin of a comparison (\Cref{sec:analysis-plan}) \\
    \bottomrule
  \end{tabular}
\end{table}

\section{Evaluation methodology}
\label{sec:metrics}

This section shows why the internal statistics of a robust fit cannot rank pipelines on a dynamic
scene and defines what replaces them: the metrics, the treatment of failures, comparability, the
audit of the reference estimator, and the analysis plan.

\subsection{Why internal-consistency metrics are insufficient}
\label{sec:why-not-inliers}

Inlier count, inlier ratio, and reprojection error measure the internal consistency of a fit: how
well the correspondences the estimator retained agree with the model fitted to them. On static
scenes they are already known to be weak proxies for accuracy. \citet{barath2023homography}
report that the inlier count depends on image resolution, on the inlier threshold and on the
estimator, so that MAGSAC returns far more inliers than its competitors at comparable pose
accuracy, and in the tabulation of \citet[Table~13]{jin2021image} the method with the highest inlier
ratio, R2D2 at \SI{52.9}{\percent}, is outperformed downstream by RootSIFT at \SI{39.6}{\percent}.

The claim here is stronger, and it follows from the scene rather than from the estimator. A
homography needs four correspondences; a group of objects moving together, such as a stream of
vehicles crossing an intersection, supplies hundreds that are mutually consistent under a single
transform, the one describing their motion. Their coverage ranges from a small fraction of a
high-altitude view to a large share of a low-altitude view over congested lanes, but the estimator
weighs their share of the correspondences, not of the area, so even a small coverage can cast many
votes. Outliers that form a coherent structure of their own bias a robust estimate toward a
compromise between the structures or capture it outright~\citep{stewart1997bias}, and a busy
intersection produces that situation persistently. Wherever the moving group wins, every
internal-consistency check is satisfied by the wrong answer: these statistics are not merely
uninformative but \emph{anti-informative}. They cannot tell whether it has won;
\Cref{sec:capture-results} measures that for the cases studied here.

The estimator cannot detect this failure on its own. \citet{chum2005degensac} detect a scene
dominated by a single plane with a test inside the estimator, because that case \emph{is} a
degeneracy: the data underconstrain the fundamental matrix, and the deficiency is visible in the
sample. A homography fitted to coherently moving objects is fully determined and over-supported; it
differs from the intended solution only in which part of the scene it describes, which no individual
correspondence encodes. None of the estimators compared here contains a check that could apply:
OpenCV runs DEGENSAC only for the fundamental matrix, and the only tests reaching a homography are
conditions on the minimal sample (no three collinear points, unchanged orientation), which a stream
of vehicles over several lanes satisfies. The information needed, which image content is static, is
external to the estimation problem, so the failure has to be handled by the protocol.

These quantities are therefore recorded as \emph{diagnostics}, to be explained rather than
optimized, and never rank a pipeline. What replaces them is the rest of this section: agreement
with an external geometry where one exists (\Cref{sec:geometric-metrics}), direct measurement of
background alignment where one does not (\Cref{sec:photometric-metrics}), and two measures that make
a moving-object consensus observable, the direction of the error (\Cref{sec:motion-axis}) and the
share of a moving group's displacement an estimate follows (\Cref{sec:capture-measure}).

\subsection{Geometric metrics}
\label{sec:geometric-metrics}

\paragraph{Corner transfer error}
Let $\Corners = \{\mathbf{c}_1,\dots,\mathbf{c}_4\}$ be the frame corners, $\mathbf{c}_1=(0,0)$, $\mathbf{c}_2=(W,0)$, $\mathbf{c}_3=(W,H)$,
$\mathbf{c}_4=(0,H)$. Given a reference transform $\mathbf{H}$ and an estimate $\widehat{\mathbf{H}}$ of its inverse direction, a
corner sent forward through $\mathbf{H}$ and back through $\widehat{\mathbf{H}}$ should return to where it started; the
residual is the error. Define the per-corner displacement
\begin{equation}
  \mathbf{d}_i \;=\; \apply{\widehat{\mathbf{H}}}{\left( \apply{\mathbf{H}}{\mathbf{c}_i} \right)} \;-\; \mathbf{c}_i,
  \qquad i = 1,\dots,4,
  \label{eq:cte-vectors}
\end{equation}
and summarize it by
\begin{equation}
  \CTEop_{\mathrm{mean}} = \tfrac{1}{4}\textstyle\sum_i \lVert \mathbf{d}_i \rVert, \qquad
  \CTEop_{\max} = \max_i \lVert \mathbf{d}_i \rVert, \qquad
  \CTEop_{\mathrm{rms}} = \bigl( \tfrac{1}{4}\textstyle\sum_i \lVert \mathbf{d}_i \rVert^2 \bigr)^{1/2}.
  \label{eq:cte}
\end{equation}
The \ac{CTE} is measured in pixels, is invariant to the parameterization of the homography, and
weights the frame periphery, where a small angular error produces the largest displacement. We keep
the displacement \emph{vectors} of \Cref{eq:cte-vectors} and not only their norms, because their
directions carry the evidence used in \Cref{sec:motion-axis}.

Two directions must be distinguished. In Part~I the perturbation $\mathbf{H}$ maps reference $\to$ warped and
the pipeline estimates the inverse, so \Cref{eq:cte-vectors} applies directly. In Part~II the
reference estimator $\Horacle$ and the candidate $\Hest$ both map current $\to$ reference, so one is
inverted first and the \emph{disagreement} between two same-direction transforms is
\begin{equation}
  \mathbf{d}_i \;=\; \apply{\Hest}{\left( \apply{\Horacleinv}{\mathbf{c}_i} \right)} \;-\; \mathbf{c}_i .
  \label{eq:cte-disagreement}
\end{equation}

\paragraph{Homography estimation accuracy}
Following~\citet{detone2018superpoint} and \citet{efe2021effect}, we aggregate over a set of $N$ trials
by the \acf{HEA}: the fraction of trials that succeed within a pixel budget $\eps$,
\begin{equation}
  \HEAop_{\eps} \;=\; \frac{1}{N} \sum_{n=1}^{N}
    \mathbf{1}\!\left[\, \Pipe_n \neq \fail \;\wedge\; \CTEop^{(n)}_{\mathrm{mean}} \le \eps \,\right].
  \label{eq:hea}
\end{equation}
A failed trial counts in the denominator but never in the numerator, so declining to answer cannot
raise accuracy; unlike a mean \ac{CTE}, the rate is not dominated by one catastrophic trial.

\paragraph{Integrating the accuracy curve}
Any single $\eps$ is arbitrary: tuning can make a method win at one threshold, yielding several
simultaneous states of the art~\citep{efe2021effect}. Following \citet{efe2021effect},
\citet{jin2021image} and \citet{barath2023homography}, we summarize
$\HEAop_{\eps}$ by the area under its curve, the \acf{mAA} over a threshold interval,
\begin{equation}
  \mathrm{mAA}\bigl[\eps_{\mathrm{lo}}, \eps_{\mathrm{hi}}\bigr] \;=\;
    \frac{1}{\eps_{\mathrm{hi}} - \eps_{\mathrm{lo}}}
    \int_{\eps_{\mathrm{lo}}}^{\eps_{\mathrm{hi}}} \HEAop_{\eps} \; \mathrm{d}\eps
  \;=\; \frac{1}{N}\sum_{n=1}^{N}
    \frac{\eps_{\mathrm{hi}} - \mathrm{clip}\bigl( \CTEop^{(n)}_{\mathrm{mean}},\,
          \eps_{\mathrm{lo}},\, \eps_{\mathrm{hi}} \bigr)}
         {\eps_{\mathrm{hi}} - \eps_{\mathrm{lo}}},
  \label{eq:maa}
\end{equation}
where the right-hand form, which is what we evaluate, is exact because each trial enters
$\HEAop_\eps$ as a step in $\eps$; a failed trial's error is taken as infinite. Each trial
contributes linearly in its error inside the interval and nothing beyond it: past a few pixels an
estimate is wrong, and how wrong carries little information.

Both parts report $\mathrm{mAA}[0, 5]$ px as the headline quantity; the cited benchmarks integrate
over other ranges and units, so their values are not comparable with these. In Part~II the
integral is evaluated per case and averaged with equal weight per case, the sampling unit. The
reference estimator's audited residual (\Cref{sec:gcp-audit}) does not raise the lower limit,
since integrating from it would still order two candidates a tenth of a pixel apart above it. It is
the equivalence margin of the reference-relative pixel metrics (\Cref{sec:analysis-plan}), mAA
keeps its margin of \num{0.01}, and $\mathrm{mAA}[\eps_{\mathrm{floor}}, 5]$ is reported beside the
headline as a sensitivity check.

\paragraph{Object-level agreement}
Frame corners describe the transform globally; the annotated objects describe where the content of
interest actually lands. For each annotated object $j$ let $\mathbf{q}_j \in \mathbb{R}^{4\times 2}$ hold
the corners of its box as the reference geometry places them in the reference frame, and
$\widehat{\mathbf{q}}_j$ the corners of the same box as the candidate places them, each corner
transformed on its own. In Part~I these are a box of $\Boxes^{\mathrm{ref}}$ and its round trip
through \Cref{eq:cte-vectors}; in Part~II they are a box of $\Boxes^{k}$ carried into the reference
frame by $\Horacle$ and by $\Hest$. We define two per-object errors,
\begin{equation}
  \IoUop_j \;=\; \IoUop\!\left( \mathbf{q}_j,\, \widehat{\mathbf{q}}_j \right),
  \qquad
  \OCEop_j \;=\; \bigl\lVert \bar{\widehat{\mathbf{q}}}_j - \bar{\mathbf{q}}_j \bigr\rVert,
  \label{eq:object-errors}
\end{equation}
where $\IoUop(\cdot,\cdot)$ is the ratio of intersection to union area of two convex
quadrilaterals and $\bar{\mathbf{q}}_j = \tfrac{1}{4}\sum_{i} \mathbf{q}_{j,i}$ is the center of box $j$.
$\IoUop_j$ is threshold-free and bounded in $[0,1]$; the \acf{OCE} is measured in pixels and
converts to meters through the \ac{GSD} of \Cref{sec:problem}. Over the objects of one frame, each is summarized by its mean
(the headline value), standard deviation, quartiles and extremes; the minimum of $\IoUop_j$ and the
maximum of $\OCEop_j$ expose a registration that is accurate on average but fails at one edge of the
frame, which a mean hides.

The two errors answer different questions. $\IoUop_j$ depends on object size as well as on
registration: a displacement that barely changes the overlap of a bus removes much of a
motorcycle's overlap. $\OCEop_j$ does not, and it is the quantity a trajectory-extraction pipeline
inherits, since such a pipeline carries each detected box into the reference frame corner by corner
and reads the object's position off the carried quadrilateral. Because $\mathbf{q}_j$ and
$\widehat{\mathbf{q}}_j$ carry the same box, the detector's localization error cancels, and
\ac{OCE} isolates what registration adds to a trajectory.

Both errors use the \acp{OBB} for every pipeline, whatever its masking strategy $\mu$, so no
candidate is scored on a geometry its own strategy chose; each $\widehat{\mathbf{q}}_j$ is compared as the quadrilateral it becomes, with no rectangle refitted
and no axis-aligned hull, which would conflate registration with orientation.

\subsection{Reference-free background-alignment metrics}
\label{sec:photometric-metrics}

On real footage no true homography exists, so alignment is measured directly: $\Fref$ is compared
with the warped current frame $\Fk \!\circ\! \Hestinv$ over an evaluation region that excludes the annotated objects and raised structures.

\paragraph{The evaluation region}
Write $S \oplus D_r$ and $S \ominus D_r$ for the morphological dilation and erosion of a pixel set
$S$ by a disc $D_r$ of radius $r$~\citep{soille2004morphological}. Let $\Veh^{\mathrm{ref}}$ be the
pixels covered by the reference frame's annotated objects and $\Veh^{k}$ those of the current
frame's, both rasterized in reference coordinates. The region is
\begin{equation}
  \Reg \;=\; \Bigl(
      \Valid \,\cap\, \overline{\Veh^{\mathrm{ref}} \oplus D_{\tau}}
             \,\cap\, \overline{\Veh^{k} \oplus D_{\tau}}
             \,\cap\, \Ground \Bigr) \ominus D_{\rho},
  \label{eq:region}
\end{equation}
where $\Valid = \{\mathbf{x} \in \pixdom : \apply{\Hestinv}{\mathbf{x}} \in \pixdom\}$ is the set of pixels
receiving real data under the warp, less a two-pixel fringe of interpolated values, $\Ground$ is the
frame less polygons drawn by hand over the buildings, other raised structures and large contiguous
tree canopy of the reference frame, $\tau$ is a constant dilation in pixels,
and the erosion by $D_\rho$ keeps the assembled region a margin away from every one of its
boundaries: the frame edge, the edge of the warped data, and the dilated objects. Each term is
necessary: without $\Valid$ a homography is rewarded for pushing empty area into frame; without the
object terms a busier scene scores worse for every pipeline; without $\Ground$ the parallax of
roofs and canopy under the platform's drift is charged to the pipeline; without the erosion, pixels at those
boundaries, where the warp interpolates and the masks meet the background, enter the metric. The
polygons are drawn once per case, before any of its results are read, and depend on no pipeline.

Two properties keep \Cref{eq:region} from being tunable by the pipeline under test. The region
is common to all candidates: $\Veh^{k}$ is placed into reference coordinates by the reference
estimator's $\Horacle$, never by the candidate's own $\Hest$, and only $\Valid$ is
candidate-specific, because it must describe the image that candidate produced. And the dilation
$\tau$ belongs to the protocol: it is a fixed pixel count, independent of the margin by which a
pipeline grows its \emph{feature-suppression} masks (\Cref{sec:masking}). $|\Reg| / |\pixdom|$ is recorded
with every measurement, and a measurement whose region falls below a minimum area fraction is
scored as the identity warp (\Cref{sec:failures}). Part~I computes no reference-free measure.

\paragraph{Four complementary measures}
Over $\Reg$ we compute \acf{SSIM}~\citep{wang2004ssim}, \acf{ZNCC},
\acf{MI}~\citep{viola1997mutualinformation,maes1997mutual}, and a magnitude-weighted agreement of
gradient orientation (GOS), a normalized form of the gradient terms used in multimodal
registration~\citep{pluim2000gradient,haber2006gradient}. Writing $A = \Fref$ and $B = \Fk \!\circ\! \Hestinv$ restricted to $\Reg$, with
means $\bar A, \bar B$ and standard deviations $\sigma_A, \sigma_B$,
\begin{equation}
  \ZNCCop(A,B) \;=\; \frac{1}{|\Reg|\,\sigma_A \sigma_B}
    \sum_{\mathbf{x} \in \Reg} \bigl( A(\mathbf{x})-\bar A \bigr)\bigl( B(\mathbf{x})-\bar B \bigr) \;\in\; [-1,1],
  \label{eq:zncc}
\end{equation}
\begin{equation}
  \MIop(A,B) \;=\; \sum_{a,b} p_{AB}(a,b) \,
    \log_2 \frac{p_{AB}(a,b)}{p_A(a)\,p_B(b)},
  \label{eq:mi}
\end{equation}
in bits, with $p_{AB}$ the joint intensity histogram over $\Reg$ ($64 \times 64$ bins) and $p_A, p_B$
its marginals, and
\begin{equation}
  \mathrm{GOS}(A,B) \;=\;
    \frac{\sum_{\mathbf{x} \in \Reg} w(\mathbf{x}) \,
          \cos\bigl( \angle \nabla A(\mathbf{x}) - \angle \nabla B(\mathbf{x}) \bigr)}
         {\sum_{\mathbf{x} \in \Reg} w(\mathbf{x})} \;\in\; [-1,1],
  \qquad
  w(\mathbf{x}) = \bigl( \lVert \nabla A(\mathbf{x})\rVert \, \lVert \nabla B(\mathbf{x})\rVert \bigr)^{1/2},
  \label{eq:gos}
\end{equation}
with Sobel gradients. The weighting makes \Cref{eq:gos} usable on aerial asphalt: flat regions,
whose gradient direction is arbitrary, contribute nothing, and road markings and curbs dominate;
the geometric mean of the two magnitudes keeps a few high-contrast edges from outweighing the rest
quadratically.

The four differ in what they tolerate: \ac{ZNCC} an affine change of intensity, \ac{MI} any
one-to-one remapping, \Cref{eq:gos} any monotonically increasing change, and \ac{SSIM}, which
compares local luminance, contrast and structure separately, the least photometric change of the
four. Agreement among them is evidence that a measured alignment is not an artifact of one
measure's blind spot.

\subsection{Decomposing error along the dominant motion axis}
\label{sec:motion-axis}

The mechanism studied here leaves a geometric signature: a homography dragged by a moving-object
consensus has its error \emph{along} the direction of motion, whereas ordinary estimation noise has no
preferred direction. Measuring that signature makes capture testable.

\paragraph{Recovering the motion axis}
Elongated objects such as vehicles travel along their own long axis, so the orientation
distribution of the \acp{OBB} recovers the direction of flow without any correspondence between
frames, which matters because the pairs here reach thousands of frames apart and no tracker links
their objects. Let $\theta_j$ be the
orientation of the long axis of box $j$ of the reference frame and $a_j$ its area. Orientations are \emph{axial} data: a box
rotated by $\pi$ is the same box, so the mean is taken over doubled angles and halved
back~\citep{mardia1999directional}. With the weighted resultant
\begin{equation}
  R \;=\; \frac{\sum_j a_j \, e^{\,2 i \theta_j}}{\sum_j a_j},
  \qquad
  \bar\theta = \tfrac{1}{2}\arg R,
  \qquad
  \kappa = \lvert R \rvert \in [0,1],
  \label{eq:motion-axis}
\end{equation}
the dominant axis is the unit vector $\mathbf{u} = (\cos\bar\theta, \sin\bar\theta)$ and $\kappa$ is its
\emph{concentration}.

\paragraph{Decomposing the error}
With $\mathbf{u}^{\perp} = (-\sin\bar\theta, \cos\bar\theta)$ and the corner displacements $\mathbf{d}_i$ of
\Cref{eq:cte-disagreement}, define
\begin{equation}
  e_{\parallel} = \tfrac{1}{4}\textstyle\sum_i \lvert \mathbf{d}_i \cdot \mathbf{u} \rvert,
  \qquad
  e_{\perp} = \tfrac{1}{4}\textstyle\sum_i \lvert \mathbf{d}_i \cdot \mathbf{u}^{\perp} \rvert,
  \qquad
  A \;=\; \frac{e_{\parallel}}{e_{\parallel} + e_{\perp}} \;\in\; [0,1].
  \label{eq:motion-alignment}
\end{equation}
$A \approx 0.5$ is isotropic error; $A \to 1$ is error almost entirely along the motion axis, the
signature of a moving-object consensus; and $A \to 0$ is error across it, which that mechanism
cannot produce. Error without a preferred direction need not give $A = 0.5$ at four corners,
however: a small rotation about the frame center displaces each corner perpendicular to its
diagonal and a small scale change along it, so on a $16{:}9$ frame with $\mathbf{u}$ along an image
axis they give $A$ of about \num{0.36} and \num{0.64}. The level of $A$ is therefore read against the identity transform, whose error is the
platform's own motion, and the evidence for capture is an increase of $A$ without masking over
$A$ with masking on the same pairs, not its level.

The bounded form is used instead of the bare ratio $e_{\parallel}/e_{\perp}$, which diverges
exactly in the case of interest ($e_{\perp} = 0$) and would drop the strongest evidence from every
aggregate. At a crossroads with two busy perpendicular approaches the doubled angles cancel and
$\kappa$ falls toward zero, so the decomposition is suppressed when $\kappa < \kappa_{\min} = 0.3$
rather than reporting an artifact. \Cref{eq:motion-axis} also needs \emph{oriented} annotations,
since an \ac{HBB} carries no orientation.

\subsection{Measuring capture directly}
\label{sec:capture-measure}

The direction of the error is a signature of capture read at the four frame corners, far from the
objects that cause it. Capture can also be measured where it happens. The reference frame's objects
are found again in frame $k$ by the procedure the case selection uses (\Cref{sec:dynamic-protocol}):
each box is compared by appearance with the boxes of similar size in frame $k$ and accepted only as
a mutual best match. With $\mathbf{y}_j$ the center of the match of reference box $j$, its
displacement with the platform's motion removed by the reference estimator is
\begin{equation}
  \mathbf{m}_j \;=\; \apply{\Horacle}{\mathbf{y}_j} \;-\; \bar{\mathbf{q}}_j .
  \label{eq:object-motion}
\end{equation}
A box has \emph{moved} when $\lVert \mathbf{m}_j \rVert$ exceeds a tenth of the median object width:
clear of the apparent motion of a stationary object, which comes from box placement and the
reference estimator's error, and less than a car covers between consecutive frames at urban speeds.
Moved objects \emph{move together} with moved object $i$ when their displacements differ from
$\mathbf{m}_i$ by at most a tenth of the median object width or a quarter of
$\lVert \mathbf{m}_i \rVert$, whichever is larger: the members of one stream drift apart in
proportion to how far it moved, while opposing or crossing streams never agree. The largest set by
area that moves together with one of its members is the \emph{group}, $G$, the structure a single
wrong homography could fit. The \emph{capture fraction} of a candidate is the share of the group's
displacement it follows,
\begin{equation}
  \lambda \;=\; \operatorname*{median}_{j \in G} \;
  \frac{-\bigl( \apply{\Hest}{\mathbf{y}_j} - \apply{\Horacle}{\mathbf{y}_j} \bigr) \cdot \mathbf{m}_j}
       {\lVert \mathbf{m}_j \rVert^{2}} .
  \label{eq:capture-fraction}
\end{equation}
An estimate that agrees with the reference estimator where the group lies has $\lambda = 0$; one
that registers the group back onto where it stood in the reference frame, which is what a
homography fitted to the group does, has $\lambda = 1$; values between are a compromise fit, pulled
toward the group without registering it. A measurement is \emph{captured} when $\lambda \geq 0.1$:
the estimate has moved, where the group lies, by at least a tenth of the group's displacement. That
lies above what a transform reaches without following the group: the identity, which uses no
features, reaches \num{0.07} from the platform's motion alone, and the reference estimator's audited
error (\Cref{sec:oracle}) amounts to \num{0.08} on the smallest displacement of any group,
\SI{5}{px} (\Cref{tab:realtime_masking-capture}). Capture is reported descriptively, with the counts at
$\lambda \geq 0.25$, $0.5$ and $1$ beside it (files \suppfile{offline_masking_capture} and
\suppfile{realtime_masking_capture}; the CSV files named in this paper are released with its
data~\citep{stabilo_benchmark_data}). $\lambda$ is defined on pairs whose group holds at
least four objects, and only where the pipeline returned an estimate. Beside it, every measurement
records where the estimate's inliers lie: the share on annotated objects and the share on objects
that moved. The second undercounts what moves with the objects, since a vehicle's cast shadow lies
outside its box; $\lambda$ does not depend on it. These quantities are diagnostics, like the inlier
statistics they are read against, and never rank a pipeline.

\subsection{Treatment of failures}
\label{sec:failures}

A registration pipeline can fail outright: too few matches survive filtering, or the robust
estimator returns a degenerate transform. Implementations commonly substitute the identity, the
stabilization library evaluated here among them~\citep{stabilo}, which keeps a campaign running but
hides the failure and biases every metric, because the identity is a \emph{plausible} answer for a
nearly stationary drone. We therefore admit $\fail$ explicitly in \Cref{eq:pipeline-map} and detect
it from what the estimator returns: an absent inlier count, which every fallback leaves, or a
missing or non-finite matrix. An estimate exactly equal to the identity is not a failure, since
between near-identical frames the estimator returns exactly the identity with thousands of
inliers, and that answer is correct. A finite estimate that drives the third homogeneous
coordinate of \Cref{eq:action} to zero at a corner yields non-finite coordinates, which
\Cref{eq:hea,eq:maa} score as a miss; clamping that division, as some implementations do, would
score a collapsing estimate as merely inaccurate.

The reference-free measures need a different treatment, because an undefined value drops out of a
mean: a pipeline that failed on a third of its pairs would otherwise report its background
alignment over the two thirds it solved. A failed pair is therefore scored with the identity warp,
the transform the pipeline actually delivered, and so is a pair whose evaluation region is too
small to be read; declining to answer costs exactly what delivering the identity costs.

A configuration that exhausts the benchmark machine's memory is \emph{not} a failure. It is
reported as not available and its partial measurements are discarded, because a sparser scene needs less memory and is also easier to
register, so the subset that fit would flatter it. Counting it as a failure would make a score depend
on the machine's memory.

\subsection{Placing pipelines on equal footing}
\label{sec:comparability}

A benchmark that compares ten pipelines at one shared set of matcher settings measures the
settings as much as the pipelines. This is the central methodological finding of the benchmarks
reviewed in \Cref{sec:related}, and we adopt their remedies.

\paragraph{Two hyperparameters are tuned per pipeline}
The ratio-test threshold and the inlier threshold of the robust estimator are both known to be
decisive and to have method-specific optima: the first thresholds a descriptor distance
distribution, a property of the descriptor, and the second a reprojection residual, a property of
keypoint localization accuracy. \citet{efe2021effect} obtain their entire
result by varying the first; \citet[Sec.~5.3]{jin2021image} find its optimum differs per feature.
They are swept \emph{jointly} rather than sequentially (the inlier threshold alone where the
matcher applies no ratio test: LightGlue and LoFTR), following
\citet{barath2023homography}: a looser ratio test admits more outliers, which moves the inlier
threshold that best separates them, so the sequential optimum need not be the joint one. Selection uses \Cref{eq:maa}, the quantity the benchmark ranks on, because the threshold maximizing
correspondence-level accuracy is not the one maximizing homography accuracy: a robust estimator
turns a surplus of looser matches into a better fit~\citep{efe2021effect}. It is also
conservative: a pipeline keeps the values its preset carries unless the best grid point beats them
on mAA by more than the equivalence margin of \num{0.01} (\Cref{sec:analysis-plan}), because where
the optimum is flat the best grid point only moves the values within a set the data cannot tell
apart; Part~II starts from Part~I's values in the same way.

\paragraph{The remaining estimator parameters are fixed, on published grounds}
Of the three parameters every robust estimator here exposes, the confidence level, the inlier
threshold and an iteration cap, only the threshold is tuned. \citet[Sec.~5.1]{jin2021image} ``find
the confidence value to be the least sensitive'' of the three and fix it at $1 - 10^{-6}$, the value
used here. The iteration cap is a \emph{time} budget rather than an accuracy parameter, and the
benchmarks we follow treat it as one: \citet{barath2023homography} fix it at $1000$ while
grid-searching the thresholds and sweep it only to draw a time--accuracy curve, and
\citet{jin2021image} set it from a per-pair time budget. We fix it at $5000$, where it is not the
binding constraint, and report wall-clock time so that no pipeline gains accuracy through unreported
computation. All remaining parameters are held at documented values across pipelines
(\Cref{sec:appendix-parameters}).

\paragraph{Tuning and evaluation scenes are disjoint}
The sweep runs on scenes held out from every reported result, a split the benchmark's tooling
enforces: a threshold chosen on the data a number is reported on measures how much tuning freedom
each pipeline was granted, and \citet{barath2023homography} name the lack of such a split as a
drawback of earlier homography datasets. The split is by \emph{site}, not by frame: the Part~I set
holds eight intersections recorded by more than one drone, near-duplicate views of the same road
layout, markings and parked vehicles, so a frame-wise split would leak the evaluation scenes into
the tuning half. Every third site in alphabetical order goes to tuning,
which reproduces the split from the file names alone. Of the \num{29} scenes, \num{9} from
\num{7} sites tune and \num{20} from \num{13} sites evaluate. Part~II keeps the split: its
evaluation cases are drawn from evaluation sites and its own tuning cases from tuning sites. Tuning is run with masking
disabled and the same parameters are used under every masking mode, so no mode is tuned into its
own best configuration; because the deployed configuration is masked, this is the conservative
choice for the masking comparison, and any advantage measured for masking is measured with
parameters tuned against it.

\paragraph{The compute budget is fixed before anything is tuned}
Two further parameters, the downsampling ratio and the feature budget, decide how much work a
pipeline does rather than how it uses it, and the shipped presets do not hand every pipeline the
same amount: the learned models' ratios are memory bounds. \citet{jin2021image} report every
method at matched keypoint budgets for this reason, and \citet{barath2023homography} report
accuracy against runtime. We therefore sweep both parameters over every pipeline on the tuning
scenes first, under masking disabled and under \ac{OBB} masking, and read off a single matched
operating point: the lowest budget at which every pipeline that can reach it is within the
equivalence margin of its own saturation under both masking conditions. Masking removes the highest-contrast
content in the frame, so the saturating budget need not be the same with and without it. If no
budget on the grid meets the criterion, every pipeline runs at the largest one and its shortfall
from saturation is reported. The thresholds are then selected at that operating point, so that
what is tuned is what is run; a pipeline that cannot reach it on the benchmark machine is reported
as such rather than scored at a lower budget. Part~II runs at two other points of the same grid, one for offline and one for real-time
registration (\Cref{sec:dynamic-protocol}); its thresholds are tuned again at each and stated in
pixels of the downsampled frame, so that they keep their size relative to keypoint localization as
the frame shrinks.

\subsection{Auditing the reference estimator}
\label{sec:gcp-audit}

Part~II has no true homography, so a strong pipeline is used as a \emph{reference estimator}
$\Horacle$ (\Cref{sec:oracle}). This is pseudo-ground truth: every Part~II number is a comparison
against an estimate that has its own error, and one drawn from the same family as the candidates
cannot be assumed neutral between them~\citep{brachmann2021limits}. We measure that error against
independent control points rather than assuming it negligible.

A set of unambiguous ground-plane \acp{GCP} is chosen by hand in $\Fref$ and located in a subset of
sampled frames, as described below. For a label $\ell$ present in both, with
$\mathbf{p}^{\mathrm{ref}}_{\ell}$ and $\mathbf{p}^{k}_{\ell}$ its annotated positions, the residual
is
\begin{equation}
  r_{\ell} \;=\; \bigl\lVert \apply{\Horacle}{\mathbf{p}^{k}_{\ell}} - \mathbf{p}^{\mathrm{ref}}_{\ell} \bigr\rVert ,
  \label{eq:gcp-residual}
\end{equation}
and the upper percentile of its distribution pooled over frames is a \emph{resolution floor}
\begin{equation}
  \eps_{\mathrm{floor}} \;=\; Q_{0.95}\bigl( \{ r_{\ell} \} \bigr).
  \label{eq:floor}
\end{equation}
Differences between candidates smaller than $\eps_{\mathrm{floor}}$ lie within the reference
estimator's own uncertainty and are not interpreted as evidence.

Because the \acp{GCP} depend on no pipeline, \Cref{eq:gcp-residual} applies equally to a candidate
$\Hest$, and this is the comparison the masking result rests on. The reference runs with
\ac{OBB} masking, the configuration expected to be most accurate, so a candidate run under the
same masking shares its feature population (the same background pixels, the same exclusions, the
same movers the detector missed), and its error is positively correlated with the reference's: its
disagreement with the reference understates its true error, while an unmasked candidate shares less
and carries more of the reference's error on top of its own. A masking comparison read off
reference-relative numbers is therefore biased toward the reference's own masking mode, and for
the reference's own detector under \ac{OBB} masking it is circular by construction;
\Cref{eq:floor} measures the reference's absolute error on the audited points but says nothing
about that covariance. The
\acp{GCP} break the dependence. On every audited frame we fit a homography to the
points themselves,
\begin{equation}
  \Hgcp \;=\; \argmin_{\mathbf{H} \in \Homog} \sum_{\ell}
    \bigl\lVert \apply{\mathbf{H}}{\mathbf{p}^{k}_{\ell}} - \mathbf{p}^{\mathrm{ref}}_{\ell} \bigr\rVert^2 ,
  \label{eq:gcp-fit}
\end{equation}
by the normalized direct linear transform with a nonlinear refinement of the reprojection
error~\citep{hartley2004multiple}, from at least \num{12} points spread to the frame periphery: the
number and placement of control points govern accuracy in UAV
photogrammetry~\citep{sanzablanedo2018gcp}, and a homography fitted to central points is poorly
determined exactly where the \ac{CTE} is read. Its error depends on no pipeline. Every candidate is
scored against $\Hgcp$ on the audited frames with the metrics of \Cref{sec:geometric-metrics}, and
those scores, with the reference-free measures, are the primary evidence for the masking
comparison; the reference-relative scores over all pairs remain the evidence for behavior with
temporal distance and for ranking detectors within one masking mode, where whatever the reference's \ac{OBB} masking
favors is common to every candidate compared. The tables of these scores flag SIFT and RootSIFT, which share the reference's keypoint
detector and whose reference-relative scores are therefore the least independent of it.

\paragraph{Placement of the points}
Points are placed in a magnified panel with an exact pixel mapping. In $\Fref$ they are chosen and annotated
manually, on the road surface, clear of every object and of the excluded structure. In each audited
frame every point is first proposed by an image fit that shares nothing with the candidates or the
reference estimator: the point's $\num{81} \times \num{81}$-pixel neighborhood in $\Fref$, with the objects of both
frames filled with the background mean, is aligned to the frame by an affine enhanced correlation
coefficient fit~\citep{evangelidis2008ecc} started from the platform motion of the case selection
(\Cref{sec:dynamic-protocol}), and the proposal stands only if the fit correlates at \num{0.8} or
better and the fit back from it returns within half a pixel of the reference point. The annotator
sees each proposal beside the point's reference neighborhood and accepts it or places the point
manually instead; a dense local fit and a sparse whole-frame homography fail in different ways, so their
agreement is evidence about the estimator, and the review guards against the failures of the fit
itself (a neighboring feature, an occlusion, a point that left the frame). A point without a
standing proposal is placed by hand in a panel that opens near where $\Horacle$ predicts it,
displaced by a hidden offset of a few pixels, so that a click following the panel rather than the
feature can only inflate the residual. On two cases the audited frames were also placed entirely
by hand (\Cref{tab:gcp-proposal-validation}): pooled over both, the unaided clicks lie a median of
\SI{0.44}{px} (\SI{0.82}{px} at the \num{95}th percentile) from the proposals and \SI{0.46}{px}
(\SI{0.91}{px}) from $\Horacle$'s predictions, while the proposals lie \SI{0.06}{px}
(\SI{0.32}{px}) from the predictions; on each case the proposals lie closer to the predictions than
either lies to the clicks, which separates the click noise from the estimator's error. The
\ac{GCP} repeatability is measured by annotating the audited frames of two cases a second time in a
separate session without proposals, the reference frame included; it is the annotation floor,
reported beside \Cref{eq:floor} so the two are not conflated. The residual is reported per
temporal-offset bin as well as pooled, because the reference's error plausibly grows with offset as
shadows move and content turns over; the first of the four audit bins holds offsets of one to three
frames, at which a moving group is still shared with the reference, so that a captured estimate can
be scored against the \acp{GCP} as well.

Finally, the evaluation region of \Cref{eq:region} places the current frame's objects with
$\Horacle$, so a reference error of $\eps$ misplaces the object exclusion by $\eps$. The dilation
$\tau$ absorbs that only while $\tau > \eps_{\mathrm{floor}}$, so the reference-free family uses the
reference only to place the exclusion, which still covers the objects while that holds; the
implementation checks the audited residual against $\tau$ and warns when the condition fails. A
candidate's own error is not absorbed: a misregistered frame carries its objects off their
exclusion, and their pixels enter its own score.

\subsection{Analysis plan}
\label{sec:analysis-plan}

The plan below was fixed before the campaigns it governs ran, and the implementation applies it when it
renders a table, so that no number is bolded, ranked, or called a difference by a rule chosen
after the outcome was seen. Like the comparisons of learning methods over several data
sets~\citep{demsar2006statistical}, it treats the independent unit, not the measurement, as the
sample, and uses paired nonparametric tests over it.

\paragraph{Units and clusters}
In Part~I one measurement is a (scene, trial) pair. The perturbation depends on the scene and the
trial only, never on the pipeline, so pipelines are compared on \emph{paired} per-trial
differences. The independent unit is the site, of which the evaluation set has \num{13}: two
frames of one intersection from two drones show the same static scene and are not independent
evidence. In Part~II one
measurement is a frame pair under one pipeline and one masking mode. Masking is a within-pair
factor, since all three modes see the same pair, so the masking effect is a paired difference per
pair. The independent unit is the case, of which there are nine, and frame pairs within a case
are never treated as independent. The nine lie at seven intersections, two of them hovered over in
two sessions; a case is defined by the traffic at its reference frame, which two sessions do not
share, so the case rather than the intersection is the unit.

\paragraph{Outcomes}
Part~I ranks on $\mathrm{mAA}[0,5]$ of \Cref{eq:maa}. The Part~II masking claim has one primary
contrast, \ac{OBB} masking against none for the average pipeline (per case, the mean paired
benefit over the pipelines measured on every outcome in every case), on five outcomes: the \ac{CTE}
against $\Hgcp$ on the audited frames, the reference-free \ac{ZNCC} and gradient-orientation
agreement over the evaluation region on all pairs, and \ac{IoU} and \ac{OCE} against $\Hgcp$ on the
audited frames. The contrast is claimed when at least one outcome favors \ac{OBB} masking after a
Holm adjustment~\citep{holm1979simple} across the five and none favors no masking at the
unadjusted level; the second condition can only withhold a claim. The outcomes differ in sensitivity: those against $\Hgcp$ see eight audited
frames per case, the reference-free ones every pair, so requiring all five to agree would tie the
claim to the least sensitive. Secondary are the same contrast per pipeline, and horizontal against
none and oriented against horizontal, both for the average pipeline and per pipeline. The mechanism claim has three pre-specified
contrasts, each formed within a case and then read across
cases: the oriented-against-none benefit increases with the coverage of the current frame; it is
smaller in pairs whose foreground did not move than in pairs whose foreground did, at matched
coverage (\Cref{sec:dynamic-protocol}); and the motion alignment $A$ of \Cref{eq:motion-alignment}
is larger without masking than with \ac{OBB} masking. Each is read for the average pipeline, as
the primary contrast is, with a Holm adjustment across its outcomes (\ac{ZNCC} and
gradient-orientation agreement for the first two, $A$ for the third), and is supported when an
outcome's interval lies above zero at an adjusted $p$ of at most \num{0.05}; the per-pipeline
readings are secondary. Capture itself is reported descriptively, per pipeline, masking mode and
offset bin: the share of pairs with a moving group on which the estimate is captured
(\Cref{eq:capture-fraction}), the cases in which that happens, and the inlier count and ratio of
captured estimates beside those of the others. It depends on the traffic each case happens to
carry at its reference frame, so it describes these cases and is not tested as a population rate.

\paragraph{Estimates and intervals}
Every comparison reports the paired difference as an effect size: the Hodges--Lehmann
estimate~\citep{hodges1963estimates} of the per-cluster paired differences, with a
\SI{95}{\percent} percentile bootstrap interval~\citep{davison1997bootstrap} obtained by resampling
sites in Part~I and cases in Part~II, with \num{10000} draws. Every outcome, mAA
included, is computed per cluster and differenced there, so an interval on mAA is an interval on
the cluster-level paired difference. The test is an exact sign-flip permutation~\citep{good2005permutation} of the
per-cluster paired differences: $2^{13}$ flips in Part~I and $2^{9} = 512$ in Part~II, so the
smallest two-sided $p$ attainable in Part~II is $2/512 = 0.0039$. After the Holm adjustment
across five outcomes, a masking effect is detectable at the \num{0.05} level only if it has the
same sign in all nine cases, or in eight with the dissenting case the smallest in magnitude
($4/512 \times 5 = 0.039$).
The design detects a consistent effect and nothing weaker. The $2 \times 2$ over moving foreground
and coherence, with two cases per cell, is reported descriptively, without intervals or an
interaction test; the within-case mechanism contrasts test its prediction.

\paragraph{Criterion for a difference}
A difference is claimed only when the bootstrap interval excludes zero \emph{and} the point
estimate exceeds an equivalence margin $\delta$, the smallest difference worth acting
on~\citep{lakens2017equivalence}. In Part~I the ground truth is exact and
$\delta = 0.01$ in mAA, one percent of its $[0,1]$ range, so that two thousand paired trials
cannot declare a difference no reader would act on. In Part~II, reference-relative pixel metrics
use the audited residual, $\delta = \eps_{\mathrm{floor}}$, which for a comparison pooled over the
cases is the largest of the per-case residuals; pixel metrics against $\Hgcp$ use the \num{95}th
percentile of the \ac{GCP} repeatability, the larger of the two repeat sessions'; and \ac{IoU} and the
reference-free measures use no margin, since their units carry no floor, and are reported with
intervals alone.

\paragraph{Multiplicity and presentation}
Part~I has one primary outcome, mAA, and Part~II one primary contrast. Secondary contrasts are reported with intervals and
Holm-adjusted $p$-values (per pipeline, across the pipelines) and mark the differences that
pass the rule above, but carry no claim. Every table bolds the best value in a column and every value not
distinguishable from it under the rule above, and says so in its caption; columns that are
diagnostics are never bolded. Bolding applies the rule without a multiplicity adjustment and is
descriptive: it establishes neither equivalence within the bold set nor an adjusted difference
outside it. The estimate, interval and adjusted $p$ behind each of those
decisions are reported in \Cref{sec:appendix-plan}. The correlation between inlier ratio and alignment quality
(\Cref{fig:inlier-vs-quality}) is computed within each pipeline and case and its distribution is
shown, never a single pooled coefficient, because estimators differ systematically in how many
inliers they report at equal accuracy~\citep{barath2023homography} and a pooled correlation would
measure that mix.

\section{Masking configurations and their annotations}
\label{sec:masking}

Three configurations are evaluated, and the complete benchmark is repeated for each: no masking,
\ac{HBB} masks, and \ac{OBB} masks. Under the latter two, the annotated boxes of both frames are
rasterized into a binary mask that the feature detector receives, so that no keypoint is detected
on an annotated object; an \ac{HBB} is rasterized as an axis-aligned rectangle and an \ac{OBB} as a quadrilateral. Each box is first scaled about its center by $1 + m$ with the mask margin
$m = \num{0.15}$, to absorb annotation slack, shadow and motion blur at the object boundary. That
margin belongs to the pipeline and is a hyperparameter of it, distinct from the constant dilation
$\tau$ of the evaluation region in \Cref{eq:region}, which belongs to the protocol. The \ac{HBB}
is the detector's and the \ac{OBB} is fitted to the segmented vehicle, so the two differ in
extent as well as in shape and neither need contain the other; the suppressed-area gap between the
two geometries is reported for every scene set.

\paragraph{Rationale for two box geometries}
Whether the shape of a mask matters for registration is not obvious in advance, and plausible
mechanisms point both ways. An \ac{HBB} around a vehicle turned off the image axes also
covers the road beside it, two wedges of asphalt that at an intersection often hold lane markings,
stop lines and curb edges, the static texture the estimator most needs; the wedges grow with the
angle, to about twice the vehicle's own area for a \num{40}$\times$\num{10} box at
\SI{45}{\degree}, and in a queue of turning vehicles adjacent \acp{HBB} merge over the
markings between them. \acp{OBB} keep that texture. Conversely, what moves with a vehicle is
not confined to its body: its cast shadow, its motion blur and the reflections off its windows
travel with it, lie partly outside a tight \ac{OBB}, and supply correspondences that agree with
its motion, and the extra area of an \ac{HBB} covers part of them. Which effect dominates, if
either, is an empirical question, so the box geometry is a controlled axis of both parts, and the
margin sweep of \Cref{sec:masking-results} compares the two geometries at equal suppressed area, so
that a difference can be told apart from the area each removes. \acp{OBB} are also what the
motion axis of \Cref{sec:motion-axis} and the object-level errors of \Cref{sec:geometric-metrics}
are read from, whatever the masking.

The annotations are the same for both parts and are produced outside the benchmark, once per image
set, by the detector released with \texttt{geo-trax}~\citep{geotrax} for the \acp{HBB} and
by the conversion of \Cref{sec:appendix-obb} for the oriented ones. Neither model runs inside the
benchmark loop: a campaign re-reads the same frames tens of thousands of times, and two pipelines
are only comparable if they were handed identical masks, which a per-run detector call cannot
promise. Each label set carries a provenance record naming the command, the settings, the package
versions and the checkpoint checksums that produced it, because annotations are not
bit-reproducible across accelerators and the published annotations, not the command, are the
reproducible artifact.

\subsection{Oriented annotations and the Songdo Vision OBB release}
\label{sec:songdo-vision-obb}

\ac{OBB} masks require an oriented annotation for every object in every frame the benchmark
touches, orders of magnitude beyond what can be drawn by hand. They are obtained from the
horizontal annotations by prompting SAM-family segmentation models~\citep{kirillov2023sam,ravi2025sam2}
with each box and fitting a minimum-area rotated rectangle~\citep{freeman1975minimum} to the
resulting mask~\citep{hbb2obb}; the procedure and its validation against a human-verified oriented
reference are given in \Cref{sec:appendix-obb}. Inside the benchmark the conversion runs over the
vehicle annotations of the scenes and cases under test, the only use on which any number in this
paper depends. Applied to the whole of Songdo Vision~\citep{songdo_vision} at the selected
configuration it yields \num{274190} \acp{OBB} over \num{5419} frames, one per \ac{HBB}
and carrying its class and identifier, released as Songdo Vision
OBB~\citep{songdo_vision_obb} with the \emph{human-verified} reference used
to select and validate the conversion: \num{4245} \acp{OBB} over \num{50} frames from six
recordings (\Cref{sec:appendix-obb-gt}). None of those \num{50} frames appears in Songdo Vision, as
checked against all \num{5419} by checksum and by decoded pixels, but they come from the same six
flights. That is a weaker guarantee than the site-level split of \Cref{sec:comparability}, and the
appropriate one here: the conversion is selected by \ac{IoU} against a human-verified reference, a quantity
that involves neither the pipelines nor their metrics, so scene overlap cannot tune it toward a
favorable stabilization result; it can only flatter the reported conversion fidelity, which
\Cref{sec:appendix-obb} therefore states as a tuned-set estimate.

\paragraph{Fallback boxes}
Where the segmentation finds no usable mask inside a box, the source \ac{HBB} is emitted
unchanged, which would dilute any difference between the two mask geometries. At the selected
configuration that share is \num{0} wherever it has been measured: among the \num{4245} boxes of
the reference set, the \num{274190} of the released conversion, and, read from the per-object
confidence written beside each label file (exactly \num{0} marks a fallback), the \num{2311} boxes
of Part~I's \num{29} scenes, the \num{60213} of Part~II's nine cases and the \num{44397} of its five
tuning cases (\num{136} frames each).

\section{Part I: static benchmark}
\label{sec:static}

\subsection{Protocol}
\label{sec:static-protocol}
\begin{figure*}[htbp]\centering
  \includegraphics[width=\textwidth]{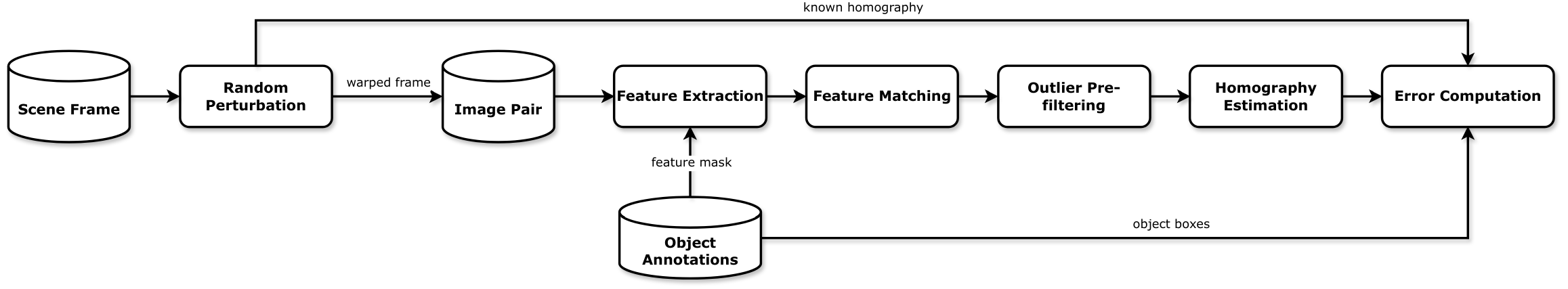}
  \caption{Part I. A scene frame is warped by a known geometric and photometric perturbation, and
  the pipeline registers the warped copy back onto the original. The perturbation is the ground
  truth, so no separate reference estimate is involved. The object annotations enter twice and for
  different purposes: as the feature mask handed to the pipeline, and as the object boxes whose
  object-level errors are scored.}\label{fig:static-protocol}
\end{figure*}

\Cref{fig:static-protocol} shows the protocol. Known homographies (translation, rotation about
the frame center, uniform scale, and a small projective term) and a chain of photometric
degradations (brightness, saturation, blur, haze, sensor noise, and JPEG compression) are applied
to static aerial frames, and the pipeline registers the degraded, warped frame back onto the
original, so the ground truth is exact. Trials are \emph{paired}: the perturbation of a (scene,
trial) does not depend on the pipeline, so every pipeline faces the identical sequence of
perturbations, and every sampled parameter is recorded with its trial. The ranges
(\Cref{tab:perturbation}) are those of the published campaign~\citep{FONOD2025105205}: translation up to
\SI{10}{\percent} of each frame dimension, rotation up to \SI{15}{\degree}, scale within
\SI{5}{\percent}, and a small projective term. This is wider than the deployed hover regime of
\Cref{sec:problem}, so that operating limits are measured;
a second configuration at deployment scale (rotation within \SI{2}{\degree}, translation within
\SI{2}{\percent}, scale within \SI{1}{\percent}, no projective term, the same photometric chain)
is run for the detector axis, and rotation is swept separately beyond the main range.

The masks for the warped copy are the scene's boxes carried through the known homography, since
the warped frame has no annotation of its own. An \ac{OBB} stays a quadrilateral; an
\ac{HBB} for the warped frame is the axis-aligned envelope of the carried \emph{\ac{OBB}}, which is what a detector would draw around the rotated object, rather than of the carried
\ac{HBB}, which would inflate an envelope a second time by an amount growing with the
rotation.

\begin{figure}[htbp]\centering
  \includegraphics[width=\textwidth]{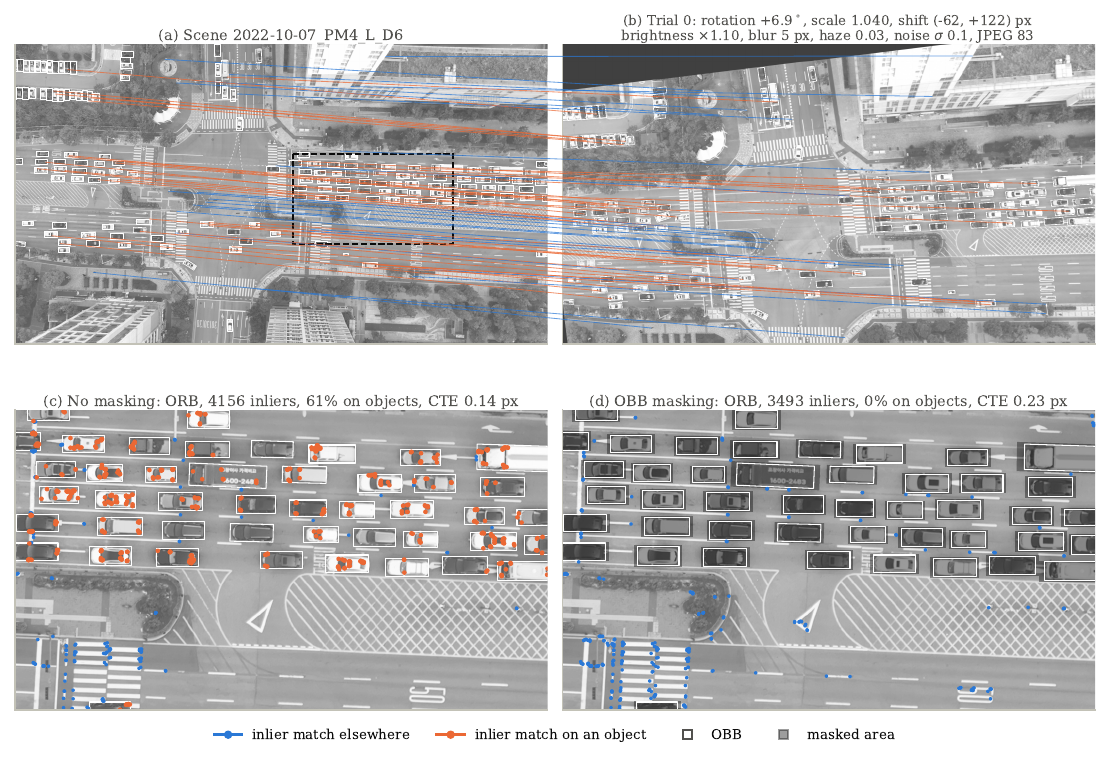}
  \caption{One trial of the detector campaign, registered again
with the campaign's stored parameters and seed. (a)~The evaluation scene with the most objects,
every \ac{OBB} outlined. (b)~Its perturbed copy in trial~0, with the sampled parameters above it and
a random sample of unmasked ORB's inlier matches drawn between the two frames. (c,\,d)~The dashed
crop of (a) with every ORB inlier unmasked and under \ac{OBB} masks, the masked area (each \ac{OBB}
scaled about its center by $1 + m$, \Cref{tab:fixed-parameters}) shaded; orange marks an inlier on an
annotated object, blue one elsewhere. Nothing in the scene moves, so the inliers on vehicles are
correct here; under masking ORB spends that share of its feature budget on the
ground.}\label{fig:static-trial}
\end{figure}

\Cref{fig:static-trial} shows one trial. Vehicles are among the most textured content of a road
scene, and without masking a large share of a pipeline's inliers lies on them. In Part~I that is
harmless, because the vehicles move with the ground; under masking, a detector that applies the
mask before its keypoint cap (\Cref{sec:pipelines}) spends that share of its budget on the ground
instead. In Part~II these are the correspondences a moving group can contribute.

The campaigns run in a fixed order on the two splits of \Cref{sec:comparability}. On the tuning
scenes: the compute-budget sweep, under masking disabled and \ac{OBB} masking, which fixes the
matched operating point; then the joint sweep of the ratio-test and inlier thresholds at that
point. On the evaluation scenes, at the selected values: the detector axis under the three masking
modes, the same axis at the untuned starting thresholds without masking, which measures the effect of
tuning on data it was not selected on, the detector axis again at deployment-scale perturbation,
the estimator axis at two fixed detectors, and the rotation sweep. Trial counts are \num{30} per scene for the tuning-side sweeps, \num{100} for the evaluation
campaigns and \num{50} for the rotation sweep.

\paragraph{Scene set and selection rule}
Part~I runs on \num{29} 4K frames from the hovering sessions of a multi-drone experiment over the
Songdo International Business District, South Korea: the static scene set of Section~3.5.2
of~\citet{FONOD2025105205}, byte-identical to the frames that campaign ran on, whose intersections
are shown in its Fig.~1. The annotation is regenerated rather than inherited, and the frames are
published for the first time with the accompanying data record~\citep{stabilo_benchmark_data}. All
\num{29} come from one session and are selected by a positional rule: the first frame of the first
hover of every drone-intersection pair flown in it. Each drone alternates between two or three
intersections during a session, and eight of the \num{20} intersections are covered in turn by more
than one drone, so the \num{10} drones record them from \num{29} viewpoints. The set is a census of
the session's viewpoints rather than a sample chosen for content: no frame was kept for being busy,
quiet, hard or easy, none was removed after the fact, and the rule reproduces the set from the file
names alone. \Cref{tab:detectors-scenes} describes the evaluation frames: object count and
covered area each vary by a factor of about four, so the masking cost is measured over a range of
densities, and the gap between the two area columns is how much more of the frame an \ac{HBB}
mask suppresses than an oriented one.

\begin{table}[htbp]
  \centering
  \caption{Object content of the 20 scenes the campaign ran on, as quantiles over scenes. Area is the fraction of the frame covered by the union of the boxes, for the oriented boxes and for the horizontal boxes they were converted from, before any mask margin; the difference between the two area columns is the additional area a horizontal mask suppresses.}
  \label{tab:detectors-scenes}
  \begin{adjustbox}{max width=\textwidth}
  \begin{tabular}{lrrr}
    \toprule
    Statistic & Objects & OBB area [\%] & HBB area [\%] \\
    \midrule
    minimum & 36 & 1.63 & 1.83 \\
    q1 & 54 & 2.35 & 2.78 \\
    median & 78 & 3.32 & 3.56 \\
    q3 & 92 & 4.47 & 4.83 \\
    maximum & 152 & 6.60 & 6.96 \\
    \bottomrule
  \end{tabular}
  \end{adjustbox}
\end{table}

\subsection{Results}
\label{sec:static-results}

Part~I asks which settings of a registration pipeline matter when nothing in the scene moves. The
results follow the settings in the order of their effect: the compute budget, the ratio-test and
inlier thresholds, and the robust estimator; then the ranking they produce at the operating point
and what masking costs. The estimator, deployment and rotation sweeps are given in full in
\Cref{sec:appendix-static}.

\begin{figure}[htbp]\centering
  \includegraphics[width=\textwidth]{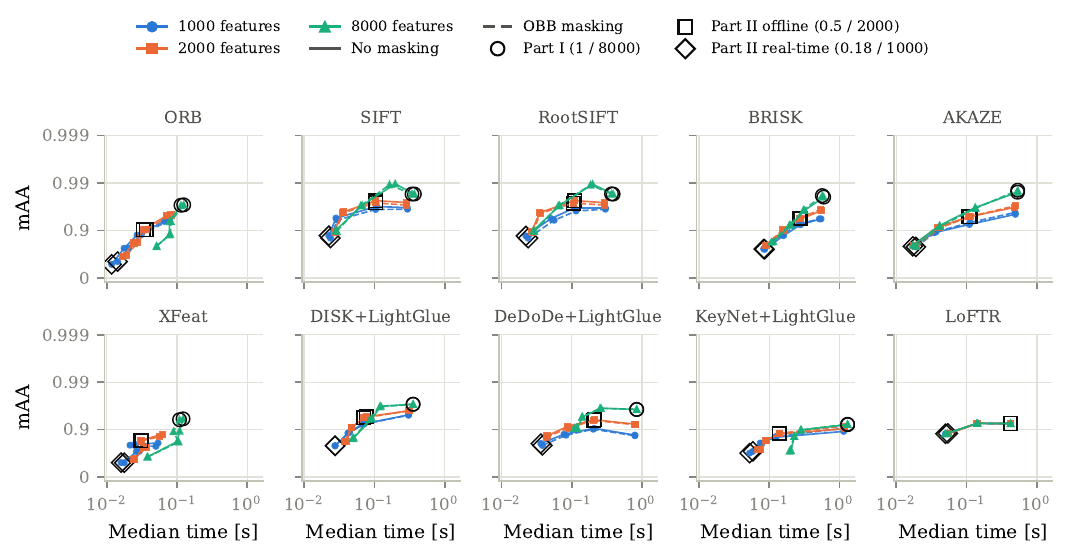}
  \caption{Accuracy against compute on the tuning scenes, at the
untuned starting thresholds. Each panel is one pipeline; a line joins the downsampling ratios
\num{0.18}, \num{0.30}, \num{0.50} and \num{1.0} at one feature budget (color and marker),
unmasked (solid) or under \ac{OBB} masks (dashed), so a split between the two lines is a masking
effect. A circle rings Part~I's operating point (\num{1.0}, \num{8000} features), a square
Part~II's offline one (\num{0.50}, \num{2000} features) and a diamond its real-time one, the
cheapest point of the grid (\num{0.18}, \num{1000} features). Time is the median registration time
per frame pair; mAA is drawn on a logarithmic scale of $1 - \text{mAA}$, so the last hundredths
before saturation stay visible. LoFTR has no point at full resolution, which exceeds the benchmark
GPU's memory.}\label{fig:budget-compute}
\end{figure}

\paragraph{The compute budget sets the accuracy}
The budget sweep crossed four downsampling ratios (\num{0.18}, \num{0.30}, \num{0.50}, \num{1.0})
with three feature budgets (\num{1000}, \num{2000}, \num{8000}) on the tuning scenes, at the
untuned starting thresholds, since it precedes the tuning (\Cref{fig:budget-compute}), and it has
the largest effect of the settings in this part. No budget met the criterion of \Cref{sec:comparability}, so every pipeline runs at the
grid's ceiling, full resolution with \num{8000} features, and the pipelines fall into two groups
below it. ORB, XFeat, KeyNet with LightGlue, BRISK and AKAZE are still gaining at the ceiling: from
half to full resolution at \num{8000} features their unmasked mAA rises by \num{0.035},
\num{0.048}, \num{0.023}, \num{0.018} and \num{0.017}, so a lower operating point would rank them
below their capacity. SIFT, RootSIFT, DISK and DeDoDe with LightGlue have saturated by half
resolution and move by less than the equivalence margin beyond it. LoFTR, which has no feature
budget, cannot reach the operating point: at full resolution it exhausts the \SI{24}{GB} of the
benchmark GPU (an NVIDIA GeForce RTX 4090) and is reported as N/A throughout Part~I; below that it
scores an mAA of \num{0.926} at ratios of \num{0.50} and \num{0.30} and \num{0.879} at its shipped
ratio of \num{0.18}. \ac{OBB} masking changes mAA by at most \num{0.065} at any grid point. The
largest change is ORB's gain at a ratio of \num{0.18} with \num{1000} features, which shrinks to
\num{0.017} at \num{2000} features and \num{0.004} at \num{8000}; the largest loss is XFeat's,
\num{0.034} at full resolution with \num{1000} features. Since nothing in these scenes moves, any
such gain is a matter of which features the budget keeps, a point \Cref{sec:masking-results}
returns to. At Part~II's real-time point, six of the ten pipelines (ORB, AKAZE, XFeat, SIFT,
RootSIFT and DISK with LightGlue) estimate a 4K frame's transform within one frame interval (\SI{33.4}{ms},
median over the tuning scenes) on the benchmark machine, with or without masking. The time covers
downsampling, feature extraction under the mask, matching and the robust fit, not decoding,
detection, box conversion or warping.

\paragraph{The matcher thresholds have little effect at the operating point}
The ratio test and the inlier threshold are the two parameters that earlier benchmarks found
decisive~\citep{efe2021effect,jin2021image}, and they are swept jointly for every pipeline
(\Cref{sec:comparability}), from untuned starting values: the ratio of \num{0.9} that the
stabilization library ships and an inlier threshold of \num{2} to \SI{3}{px} per detector, listed
with the selected values in \Cref{tab:tuned-parameters}. At the operating point the surfaces are
flat. All \num{80} grid points of SIFT, RootSIFT, ORB, BRISK and AKAZE lie within the equivalence
margin of the best, against \num{57} of XFeat's; LightGlue applies no ratio test, so its three
pipelines are swept over the threshold alone, and all \num{8} of their points are tied. Tuning
raises mAA over the starting values by \num{0.0062} for XFeat and by at most \num{0.0011} for every
other pipeline. The incumbents are the values each preset carries, and every pipeline keeps them,
since no grid point beats them by more than the margin; the largest gain on offer is
\num{0.0002}, for AKAZE at a ratio of \num{0.8}. At the kept ratio the inlier threshold moves mAA by at most \num{0.0002} for every
pipeline, so the data do not identify it. The estimator is the reason
(\Cref{fig:estimator-thresholds}): MAGSAC++, the operating estimator, marginalizes over the noise
scale instead of committing to one threshold. Below a ratio of \num{0.70} mAA stays flat for SIFT,
RootSIFT, AKAZE and BRISK and falls for ORB and XFeat, so the kept ratios lie on the plateau rather
than at an edge of it.

The selection carries over to the evaluation scenes. Unmasked, every pipeline's evaluation-set
mAA (\Cref{tab:detectors-static}) differs from its tuning-set mAA by at most \num{0.0025}, the
largest drop being KeyNet with LightGlue's, about a quarter of the equivalence margin, and the
order of the pipelines is unchanged except that SIFT and RootSIFT, \num{0.0001} apart on the tuning
scenes, swap places. Run at the starting values on the evaluation scenes
(file \suppfile{detectors_default_static}), the pipelines score lower by \num{0.0034} for XFeat and
differ by at most \num{0.0011} for the other eight, and the only change of rank is that SIFT and
RootSIFT, \num{0.0002} apart, swap places. At this operating
point, then, untuned values do not produce the reordering \citet{efe2021effect} report; the
protocol tunes every pipeline regardless, since whether the starting values are adequate cannot be
known before the sweep is run.

\paragraph{The estimator matters through its threshold}
The estimator axis is swept for ORB and SIFT, one binary and one floating-point descriptor. With the
detector fixed, the estimator matters less than the threshold it runs at
(\Cref{fig:estimator-thresholds}): with SIFT every estimator but RHO reaches an mAA of \num{0.983} to
\num{0.984} at its best threshold, and MAGSAC++ and LMEDS move by less than \num{0.0001} over the
whole range of $\theta$ for both detectors, since neither commits to one threshold. The inlier ratio
records the threshold rather than the estimate: for ORB it rises from about \num{0.17} to \num{0.81}
between $\theta = \SI{0.5}{px}$ and \SI{4}{px} whether mAA rises, falls or stays constant.

\paragraph{The ranking at the operating point}
\begin{table}[htbp]
  \centering
  \caption{Part I (static benchmark): accuracy under known homographies and photometric degradation. mAA integrates HEA$_{\varepsilon}$, the share of trials whose corner transfer error (CTE) stays within $\varepsilon$, over $\varepsilon \in [0, 5]$ px, a failed estimate counting as a miss; IoU and OCE (object center error) compare each object's box with its round trip through the known homography and the estimate. Bold marks the best value of each column and every value not distinguishable from it: a paired difference to the best, over the 13 sites, whose 95\% cluster-bootstrap interval includes zero or whose estimate is below the equivalence margin (0.01 for mAA). HEA$_{1}$, HEA$_{3}$ and HEA$_{5}$ are at least 0.999 in every row. No configuration failed on any trial. N/A marks configurations that could not be run on the benchmark machine: memory was exhausted before any measurement was taken, so the run was aborted rather than averaged over the subset of the data that happened to fit.}
  \label{tab:detectors-static}
  \small
  \setlength{\tabcolsep}{4pt}
  \begin{adjustbox}{max width=\textwidth}
  \begin{tabular}{llrrrrrr}
    \toprule
    Pipeline & Masking & mAA & CTE$_{50}$ [px] & CTE$_{90}$ [px] & IoU & OCE [px] & Time [s] \\
    \midrule
    AKAZE & HBB & \textbf{0.9845} & \textbf{0.070} & \textbf{0.129} & \textbf{0.9983} & \textbf{0.035} & 0.529 \\
    AKAZE & OBB & \textbf{0.9845} & 0.071 & \textbf{0.128} & \textbf{0.9983} & \textbf{0.034} & 0.528 \\
    AKAZE & none & \textbf{0.9839} & 0.075 & \textbf{0.130} & \textbf{0.9984} & \textbf{0.034} & 0.530 \\
    RootSIFT & none & \textbf{0.9830} & 0.082 & \textbf{0.125} & 0.9967 & 0.069 & 0.389 \\
    RootSIFT & OBB & \textbf{0.9828} & 0.083 & \textbf{0.126} & 0.9967 & 0.069 & 0.370 \\
    SIFT & none & \textbf{0.9828} & 0.084 & \textbf{0.125} & 0.9967 & 0.069 & 0.308 \\
    RootSIFT & HBB & \textbf{0.9827} & 0.083 & \textbf{0.127} & 0.9967 & 0.069 & 0.362 \\
    SIFT & OBB & \textbf{0.9826} & 0.083 & \textbf{0.128} & 0.9967 & 0.069 & 0.318 \\
    SIFT & HBB & \textbf{0.9825} & 0.084 & \textbf{0.129} & 0.9967 & 0.069 & 0.313 \\
    BRISK & OBB & \textbf{0.9811} & 0.087 & 0.145 & 0.9978 & 0.045 & 0.473 \\
    BRISK & HBB & \textbf{0.9811} & 0.088 & 0.146 & 0.9978 & 0.046 & 0.468 \\
    BRISK & none & \textbf{0.9799} & 0.094 & 0.153 & 0.9978 & 0.046 & 0.500 \\
    DISK+LightGlue & none & 0.9720 & 0.128 & 0.225 & 0.9949 & 0.104 & 0.357 \\
    DISK+LightGlue & OBB & 0.9715 & 0.131 & 0.232 & 0.9949 & 0.105 & 0.356 \\
    DISK+LightGlue & HBB & 0.9715 & 0.130 & 0.232 & 0.9949 & 0.105 & 0.355 \\
    ORB & none & 0.9703 & 0.141 & 0.217 & 0.9956 & 0.091 & 0.104 \\
    ORB & OBB & 0.9696 & 0.143 & 0.227 & 0.9955 & 0.094 & 0.118 \\
    ORB & HBB & 0.9695 & 0.143 & 0.227 & 0.9955 & 0.094 & 0.117 \\
    DeDoDe+LightGlue & none & 0.9626 & 0.178 & 0.294 & 0.9927 & 0.154 & 0.842 \\
    DeDoDe+LightGlue & OBB & 0.9622 & 0.179 & 0.296 & 0.9926 & 0.156 & 0.839 \\
    DeDoDe+LightGlue & HBB & 0.9621 & 0.179 & 0.298 & 0.9926 & 0.156 & 0.835 \\
    XFeat & none & 0.9458 & 0.254 & 0.414 & 0.9907 & 0.206 & 0.112 \\
    XFeat & OBB & 0.9434 & 0.266 & 0.425 & 0.9904 & 0.211 & 0.101 \\
    XFeat & HBB & 0.9434 & 0.270 & 0.426 & 0.9904 & 0.210 & \textbf{0.099} \\
    KeyNet+LightGlue & OBB & 0.9193 & 0.394 & 0.628 & 0.9825 & 0.351 & 1.304 \\
    KeyNet+LightGlue & none & 0.9189 & 0.396 & 0.632 & 0.9821 & 0.358 & 1.315 \\
    KeyNet+LightGlue & HBB & 0.9186 & 0.398 & 0.638 & 0.9824 & 0.353 & 1.300 \\
    LoFTR & all & \textit{N/A} & \textit{N/A} & \textit{N/A} & \textit{N/A} & \textit{N/A} & \textit{N/A} \\
    \bottomrule
  \end{tabular}
  \end{adjustbox}
\end{table}
\Cref{tab:detectors-static} reports the detector axis on the \num{20} evaluation scenes at the
selected values. AKAZE, RootSIFT, SIFT and BRISK form the leading tie set under all three masking
strategies: AKAZE scores highest, at \num{0.9839} unmasked and \num{0.9845} under either mask, and the
other three lie within the equivalence margin of it (file \suppfile{detectors_comparisons}). DISK with
LightGlue and ORB follow at \num{0.972} and \num{0.970} unmasked, \num{0.012} and \num{0.014}
behind with intervals clear of zero, and DeDoDe with LightGlue, XFeat and KeyNet with LightGlue
trail by \num{0.022} to \num{0.066}. In pixels the spread at the top is small: with every trial
inside the integration interval, mAA over $[0,5]$\,px is one minus the mean \ac{CTE}
divided by \SI{5}{px}, so the margin of \num{0.01} corresponds to \SI{0.05}{px} of mean error, and
the unmasked median error of the tie set lies between \SI{0.075}{px} and \SI{0.094}{px}. No
pipeline that ran fails on any of the \num{2000} trials or leaves the integration interval under
any masking strategy. Speed varies far more than accuracy. The unmasked median registration time spans
a factor of about thirteen: ORB and XFeat register a frame in \SIrange{0.10}{0.11}{s}, SIFT, DISK
with LightGlue and RootSIFT in \SIrange{0.31}{0.39}{s}, BRISK and AKAZE in
\SIrange{0.50}{0.53}{s}, and DeDoDe and KeyNet with LightGlue in \SIrange{0.84}{1.3}{s}. SIFT
reaches the tie set in less than three fifths of AKAZE's time, and ORB reaches a median error of
\SI{0.141}{px}, against \SI{0.075}{px} for AKAZE, in a fifth of its time.

\paragraph{Deployment regime and rotation}
At deployment-scale perturbation every pipeline scores higher, the leading tie set grows from four
pipelines to six and no pipeline moves by more than two places. The upright learned models collapse
beyond \SI{30}{\degree} of in-plane rotation (\Cref{fig:rotation-sweep}), well outside the
\SI{15}{\degree} of the main configuration. Both sweeps are in \Cref{sec:appendix-static}.

\paragraph{Cost of masking on static scenes}
The objects in a static scene move with the ground, so masking them removes correct
correspondences or, for a detector that refills its budget, exchanges them for others
(\Cref{fig:static-trial}), and a small cost is expected. This is the baseline against which the
dynamic results of \Cref{sec:masking-results} are attributed to object motion rather than to
masking as such. \ac{OBB} masking costs every pipeline less than the equivalence margin. The
largest loss in mAA is \num{0.0024}, for XFeat; for the remaining eight pipelines the change is
below \num{0.0013} in magnitude, and AKAZE, BRISK and KeyNet with LightGlue gain up to
\num{0.0012}. \ac{HBB} masking costs as little: \num{0.0024} for XFeat and at most \num{0.0008}
for the other eight pipelines, two of which gain. It scores below \ac{OBB} masking for every
pipeline but AKAZE, as the larger mask predicts (a median of \SI{3.56}{\percent} of the frame
against \SI{3.32}{\percent}, \Cref{tab:detectors-scenes}), but by at most \num{0.0007}, so at these
object densities the extra area an \ac{HBB} removes costs almost nothing, and a difference
between the two mask shapes in the dynamic benchmark is not explained by the area they remove
alone.

\section{Part II: dynamic benchmark}
\label{sec:dynamic}

\subsection{Protocol}
\label{sec:dynamic-protocol}
\begin{figure*}[htbp]\centering
  \includegraphics[width=\textwidth]{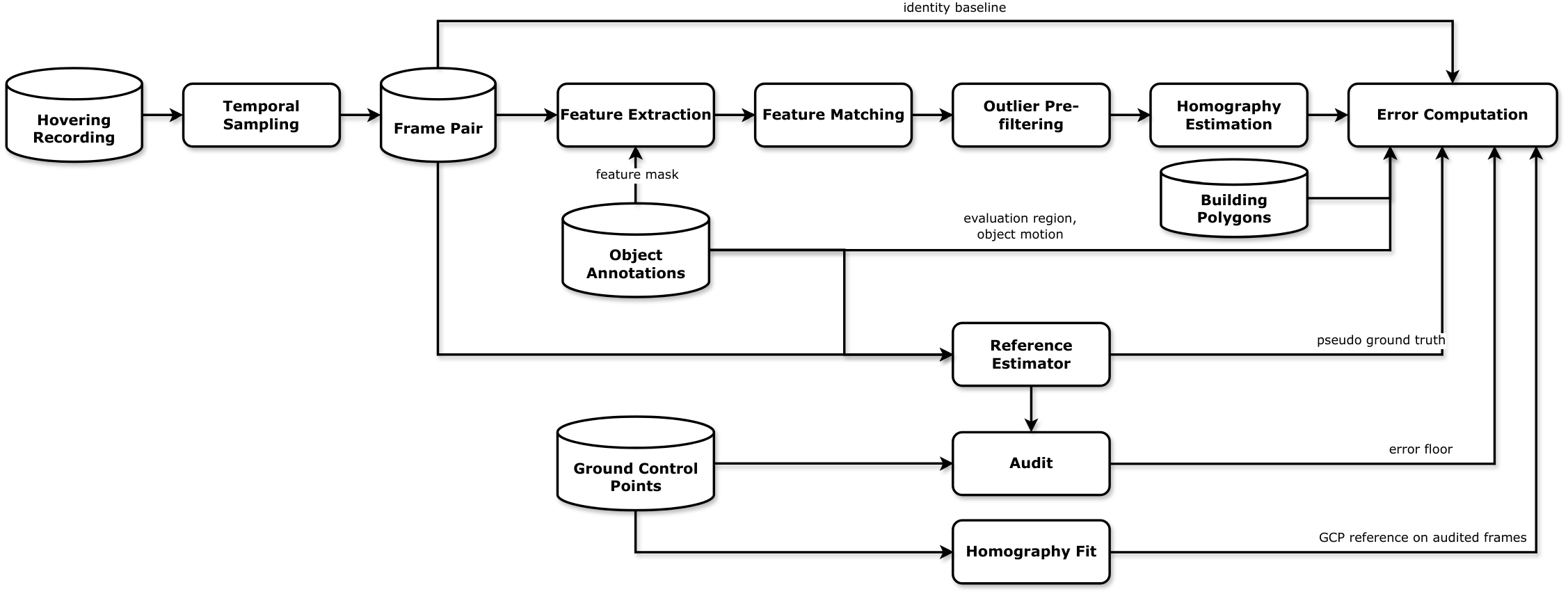}
  \caption{Part II. Every candidate registers the same frame pairs under each masking mode in turn,
  and the reference estimator registers them once, under \ac{OBB} masks. The reference
  estimator supplies pseudo-ground truth and its own error is measured against \acp{GCP}; that residual is the equivalence margin of the reference-relative pixel metrics. On the audited frames a homography fitted
  to the same points is an independent reference of its own, and the identity transform is scored on
  every pair as the unstabilized baseline. The object annotations enter as the feature mask and,
  placed by the reference estimator, as the evaluation region and as the object motion that the
  capture fraction is read against.}\label{fig:dynamic-protocol}
\end{figure*}

\Cref{fig:dynamic-protocol} shows the protocol and \Cref{tab:dynamic-parameters} its constants.
A case is one hover, the recording of one drone over one intersection cut to its stable part,
which lasts two to four minutes~\citep{FONOD2025105205}, and every sampled frame of it is registered
directly onto one reference frame, as the deployed pipeline does. The reference is a frame the
case selection picks inside the hover rather than its first: traffic at an intersection comes in
phases, and whether a group of vehicles moving together is in view at the reference decides whether
capture can occur at all.

Temporal offsets run up to \num{3600} frames (\SI{2}{\minute} at \num{29.97} frames per second) in
twelve bins one octave wide, \num{1}, \num{2}--\num{3}, \num{4}--\num{7}, \num{8}--\num{15}, and so
on to \num{1024}--\num{2047} and \num{2048}--\num{3600} frames, with \num{15} pairs drawn in each bin
that holds that many and every offset of the four shortest, \num{135} pairs per case, each weighing
the same. The spacing follows the mechanism. Capture needs objects present in both frames that
moved together, and both conditions decay fast: in the Songdo Traffic
trajectories~\citep{songdo_traffic}, over all \num{20} intersections and \num{4} days, about
\SI{90}{\percent} of the vehicles in view at a given instant are still in view \SI{4}{\second}
later, \SI{64}{\percent} after \SI{16}{\second}, \SI{45}{\percent} after \SI{32}{\second},
\SI{22}{\percent} after \SI{64}{\second} and \SI{5}{\percent} after \SI{120}{\second} (lower bounds,
since the ends of a hover cut some stays short), and the members of a moving group drift apart
within seconds. Evenly spaced offsets would spend most pairs on frames that share no moving group
with the reference and average the effect away; the octaves resolve the shortest offsets, where a
group is still shared and still moves as one, and the results are read per bin.

Every pipeline of Part~I is a candidate, so that a ranking that changes between static and dynamic
footage can be observed rather than assumed away, and every candidate runs under all three masking
modes on every pair, so masking is a within-pair factor. SIFT, the reference estimator's own
detector, is among them, which makes the one pairing that is circular by construction
(\Cref{sec:gcp-audit}) explicit. The candidates run in two regimes, both points of Part~I's budget
grid (\Cref{fig:budget-compute}). The \emph{offline} regime, for registering a recorded video after
the flight, is half resolution with \num{2000} features, the operating point of the campaign this
work extends~\citep{FONOD2025105205}. The \emph{real-time} regime is the grid's cheapest point, a
downsampling ratio of \num{0.18} with \num{1000} features, at which six of the ten pipelines
estimate a frame's transform within one frame interval (\SI{33.4}{ms}) at the median on the
benchmark machine. In both, the
inlier threshold is given in pixels of the downsampled frame, so that it keeps its size relative to
keypoint localization. The regimes share the cases, the pairs, the reference estimator and the
\acp{GCP}; the mask-margin sweep and the repeated runs are made in the offline regime only. Two
further rows are scored on every pair without any pipeline: the \emph{identity}, which is the
unstabilized frame and the transform a failed pipeline delivers, and the reference estimator
itself.

\paragraph{Thresholds are tuned again in each regime}
The ratio-test and inlier thresholds are not carried over from Part~I, where they were selected
under synthetic warps of static scenes: moving objects change which correspondences are outliers,
and the inlier threshold in particular decides whether matches on a moving group join the
consensus. They are tuned in each regime by Part~I's procedure and selection rule, with Part~I's
values as the incumbents (a joint grid widened to \num{14} ratios from \num{0.4} to \num{0.99} and
\num{13} inlier thresholds from \num{0.1} to \SI{16}{px}, selected on mAA, masking disabled),
scored against the reference estimator on five tuning cases that the survey selects, one per cell
and one control, from the intersections of Part~I's tuning scenes, so that no tuning case shares an
intersection with an evaluation case. Offline, SIFT, BRISK and AKAZE move to the strictest corner of
the grid, $r = \num{0.40}$ and $\theta = \SI{0.1}{px}$, and XFeat to $r = \num{0.45}$ and
$\theta = \SI{8}{px}$, gaining \num{0.0102} to \num{0.0156} mAA on the tuning cases; in real time
ORB, AKAZE, XFeat, DISK+LightGlue and DeDoDe+LightGlue move, gaining \num{0.0100} to \num{0.1249};
the others keep Part~I's values
(\Cref{tab:offline_tuning-tuned-parameters,tab:realtime_tuning-tuned-parameters}). The selections at
an end of the grid are not followed past it: the offline ones agree with the reference estimator to
a median of \SIrange{0.08}{0.20}{px}, and the margin of \num{0.01} mAA corresponds to about
\SI{0.05}{px} of error, both well below the audited floor (\Cref{sec:gcp-audit}), so a stricter
setting could only gain agreement with the reference that the audit cannot resolve. The same holds
for AKAZE's $\theta = \SI{0.1}{px}$ in real time (\SI{0.45}{px}); ORB's real-time $r = \num{0.99}$
lies next to $r = 1$, where the ratio test no longer rejects anything. The thresholds selected without masking are then used under all
three masking modes, so a measured masking advantage is not inflated by the tuning.

\paragraph{Cases are selected from the video alone}
The hypothesis predicts \emph{when} capture is severe: where more of the foreground moves, and moves
coherently. Testing that needs footage spanning those conditions, so the nine cases are selected by
design, on the amount and the coherence of the moving foreground as measured from the video. The
candidates come from the hovers of one recording day at the evaluation sites of Part~I's split; the next day in a fixed order is added only if a cell cannot be filled.
Every hover is read at points \SI{30}{\second} apart, clear of its first and last seconds and with
the full \num{3600} frames after them, and each point is a candidate reference frame; a hover
supplies at most one case. Each candidate is read the way the benchmark reads it: as pairs of the
candidate reference with one later frame in each of the five shortest offset bins (\num{1},
\num{2}, \num{5}, \num{11} and \num{22} frames, the bins' geometric centers), since every benchmark
pair shares the reference and a moving group moves as one only briefly. In each pair the platform's
motion is removed by an affine enhanced correlation coefficient fit to the
background~\citep{evangelidis2008ecc}, with each frame's objects filled with the background mean,
and every object annotated in the reference frame is sought among the later frame's detections by
appearance: its patch, resampled along its \ac{OBB}, is compared by \ac{ZNCC}, and only a mutual best
match between boxes of similar size is accepted. A reference object is then found where it was,
found displaced (moved by more than a tenth of the median object width), or not found, and the
largest set of displaced objects moving together is the largest structure a single wrong homography
could fit, both as defined for the capture measure (\Cref{sec:capture-measure}). Averaged over the
offsets, the displaced share of the frame is how many votes the moving foreground can cast, and the
share of it moving as one structure, the \emph{coherence}, is how many it can cast for one model;
opposing or crossing streams have low coherence however aligned their boxes are. Distances are
measured in object widths, so the procedure does not depend on ground resolution, and nothing is
tracked: six annotated frames per candidate suffice. The design is a $2 \times 2$ split at the
pool medians (\Cref{tab:survey-cases}): a candidate is \emph{high moving} if its displaced share of
the frame is at or above the median and \emph{low moving} otherwise, and \emph{coherent} or
\emph{mixed} by its coherence in the same way. Each cell holds two cases, ranked by their margin
from the split boundaries, with at most one hover per site in a cell; an unfilled cell is reported,
never padded from its neighbor.

The selection reads video alone, which keeps it applicable to footage without trajectories, but
the trajectories published for the same recordings~\citep{songdo_traffic} allow its descriptors to
be checked. Applying the survey's definitions to them, in ground coordinates where the platform's
motion is absent by construction, gives an independent value of every descriptor for every
candidate (\Cref{tab:survey_validation}): the static, moving and coherently moving foreground
correlate with their trajectory values at $r = 0.97$, coherence at $r = 0.90$ and stationarity at
$r = 0.93$; the two place 81\% of the candidates in the same cell and 97\% of the quarter farthest
from the split boundaries, where the selection draws, and the nine selected cases all fall in the
cells the trajectories assign.

\begin{figure}[htbp]\centering
  \includegraphics[width=\textwidth]{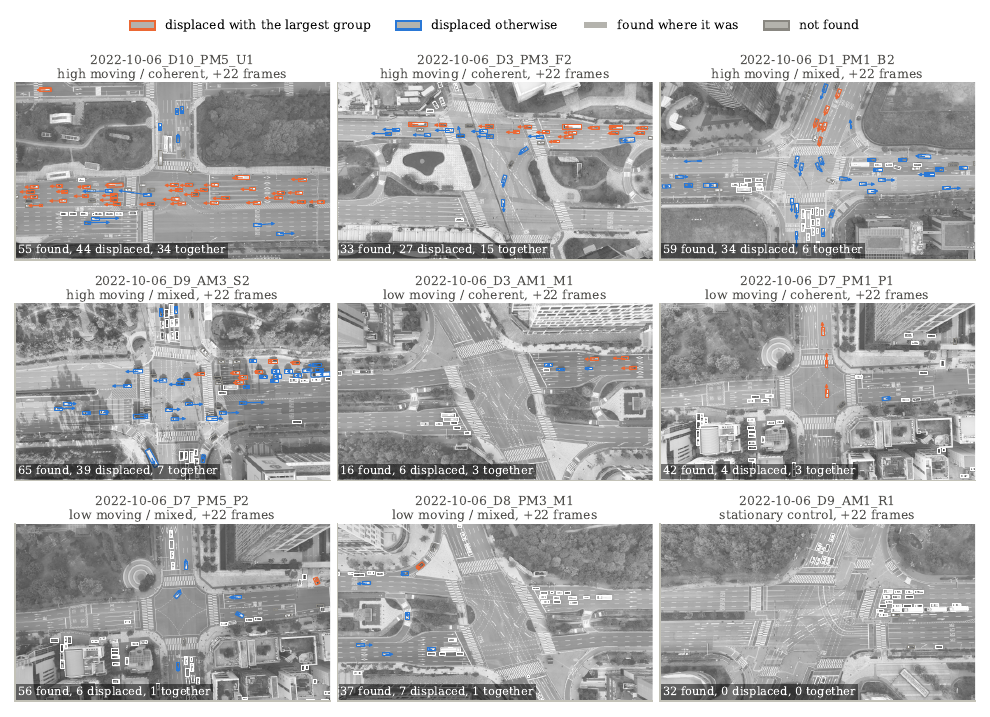}
  \caption{The nine cases of \Cref{tab:survey-cases} at their reference
frames, in cell order. Every object annotated in the reference frame is outlined by its \ac{OBB}
and colored by what it did \num{22} frames later, the longest offset the survey reads: displaced
with the largest group that moved together, displaced otherwise, found where it was, or not found.
Objects are found again as the survey finds them, with the reference estimator's transform as the
platform motion, as for the capture measure (\Cref{sec:capture-measure}); arrows are
displacements at true scale, and each panel counts the objects found, displaced and moving
together; the low-moving cases show few displaced objects by definition. Each panel is one pair, whereas the survey's descriptors average five offsets and remove
the platform motion by an affine fit to the background.}\label{fig:cases}
\end{figure}

\paragraph{The stationary control}
A queue at a red light is much foreground with little motion. If masking helps there as much as in a
flowing case, the effect comes from removing texture rather than moving objects, and the proposed mechanism
does not hold. The design therefore includes a stationary control case, a candidate whose reference
foreground mostly stays where it was (the unmoved share of the found foreground in the pool's top
decile, with at least the pool's median amount of it). Because one case gives one number, the
control is also defined at the level of the frame pair: the occupancy persistence between the
reference frame's objects and the current frame's objects placed in reference coordinates (the
intersection over union of the two footprints) is recorded with every measurement, so moving and
stationary foreground can be compared within one case, under the same background, which is the
comparison the texture confound needs. \Cref{tab:survey-cases} lists the nine cases with the
descriptors they were placed on, \Cref{fig:cases} shows each at its reference frame, and file
\suppfile{offline_masking_scenes} characterizes their frames in the terms of
\Cref{tab:detectors-scenes}.

\begin{table}[htbp]
  \centering
  \caption{Part II: the cases, selected from 831 candidate reference frames in 370 hovers of 2022-10-06. Candidates are points every 900 frames inside each hover, clear of its first and last seconds, and a case's reference frame is its selected point; a hover supplies at most one case. Each candidate is read on pairs of its reference frame with the frames 1, 2, 5, 11 and 22 frames later. Moving: the share of the frame covered by reference-frame objects displaced by more than 0.1 median object widths; Coherence: the share of that foreground moving with one of them (within 0.1 widths or 25\% of its displacement, whichever is larger); Stationarity: the share of the found foreground that stayed in place; each averaged over the offsets. Cells split at the pool medians (moving 0.67\%, coherence 0.42); coherence is undefined below 4 displaced objects, which left 72 candidates unclassified, and the control comes from the top decile of stationarity. Reference: the reference frame's index in its recording; Objects: its detections. Every cell was filled.}
  \label{tab:survey-cases}
  \small
  \begin{tabular}{llrrrrr}
    \toprule
    Case & Cell & Reference & Moving [\%] & Coherence & Stationarity & Objects \\
    \midrule
    2022-10-06\_D10\_PM5\_U1 & high moving / coherent & 1170 & 1.94 & 0.80 & 0.25 & 67 \\
    2022-10-06\_D3\_PM3\_F2 & high moving / coherent & 259 & 1.43 & 0.72 & 0.17 & 40 \\
    2022-10-06\_D9\_AM3\_S2 & high moving / mixed & 300 & 1.89 & 0.22 & 0.48 & 76 \\
    2022-10-06\_D1\_PM1\_B2 & high moving / mixed & 240 & 1.48 & 0.22 & 0.50 & 70 \\
    2022-10-06\_D7\_PM1\_P1 & low moving / coherent & 300 & 0.17 & 0.74 & 0.90 & 44 \\
    2022-10-06\_D3\_AM1\_M1 & low moving / coherent & 237 & 0.18 & 0.59 & 0.79 & 18 \\
    2022-10-06\_D7\_PM5\_P2 & low moving / mixed & 1200 & 0.24 & 0.23 & 0.89 & 62 \\
    2022-10-06\_D8\_PM3\_M1 & low moving / mixed & 1162 & 0.26 & 0.25 & 0.84 & 38 \\
    2022-10-06\_D9\_AM1\_R1 & stationary control & 2100 & 0.00 & -- & 1.00 & 34 \\
    \bottomrule
  \end{tabular}
\end{table}

\subsection{Reference estimator and its audited error}
\label{sec:oracle}

A \emph{reference estimator} is built from the leading pipelines of Part~I under \ac{OBB}
masking (\Cref{tab:detectors-static}): AKAZE, RootSIFT, SIFT and BRISK cannot be told apart on mAA,
and SIFT, the most widely used of the four, is taken. It runs at full resolution with a feature
budget of \num{20000} keypoints, a larger mask margin and a stricter estimator
(\Cref{tab:fixed-parameters}), and its ratio-test and inlier thresholds are tuned as every Part~I
pipeline is, on the static tuning scenes against their known homographies, at these settings. The
surface is flat, all \num{70} grid points within the equivalence margin of the best, so SIFT's
values, $r = \num{0.70}$ and $\theta = \SI{0.5}{px}$, are kept, at an mAA of \num{0.9856}
(\Cref{tab:oracle_tuning-tuned-parameters}). Its homographies $\Horacle$ are cached once, so every
candidate is scored against identical geometry. It runs masked because that is the configuration
expected to be most accurate; \Cref{sec:gcp-audit} states the consequence, which is why the masking
result is read primarily off the \acp{GCP} and the reference-free measures. Choosing
the reference on Part~I's ranking assumes what the paper questions elsewhere, that static accuracy
transfers to dynamic scenes; it is admissible only because the reference's accuracy on the dynamic
footage is then measured against independent control points rather than inferred.

Ground-plane \acp{GCP} are annotated in the reference frame and in eight sampled frames per case,
two in each of four offset bins (\num{1}--\num{3}, \num{4}--\num{31}, \num{32}--\num{511} and
\num{512}--\num{3600} frames), the first at the offsets where a moving group is still shared with
the reference, and the residual of $\Horacle$ against them is reported per offset bin and pooled,
with the \ac{GCP} repeatability beside it.

\begin{table}[htbp]
  \centering
  \caption{Part II: the reference estimator against ground control points, per case. Frames: the audited frames besides the reference frame; Points: the ground control points matched in them; Res.$_{50}$ and Res.$_{95}$: median and 95th percentile of the distance between a point's annotated position and its reference-frame point carried by the reference estimator, over the case and per offset bin (frames); Rep.$_{95}$: 95th percentile of the GCP repeatability, where a second session exists; $<$\,0.5\,px: the share of the points placed by hand (\Cref{tab:gcp-proposals}) within half a pixel of the reference estimator's prediction, near which the annotation panel opens under a hidden offset. The last row pools every case; the largest per-case Res.$_{95}$ is the equivalence margin of the pooled reference-relative comparisons.}
  \label{tab:gcp-audit}
  \small
  \setlength{\tabcolsep}{4pt}
  \begin{adjustbox}{max width=\textwidth}
  \begin{tabular}{llrrrrrrrrrr}
    \toprule
    Case & Cell & Frames & Points & Res.$_{50}$ [px] & Res.$_{95}$ [px] & Res.$_{95}$ 1--3 & Res.$_{95}$ 4--31 & Res.$_{95}$ 32--511 & Res.$_{95}$ 512--3600 & Rep.$_{95}$ [px] & $<$\,0.5\,px \\
    \midrule
    2022-10-06\_D1\_PM1\_B2 & high moving / mixed & 8 & 120 & 0.08 & 0.29 & 0.13 & 0.17 & 0.24 & 0.46 & -- & 67\% \\
    2022-10-06\_D10\_PM5\_U1 & high moving / coherent & 8 & 119 & 0.05 & 0.65 & 0.62 & 0.55 & 0.47 & 0.73 & 0.93 & 62\% \\
    2022-10-06\_D3\_AM1\_M1 & low moving / coherent & 8 & 119 & 0.05 & 0.38 & 0.07 & 0.11 & 0.17 & 0.81 & -- & 33\% \\
    2022-10-06\_D3\_PM3\_F2 & high moving / coherent & 8 & 119 & 0.05 & 0.16 & 0.07 & 0.34 & 0.12 & 0.18 & -- & 100\% \\
    2022-10-06\_D7\_PM1\_P1 & low moving / coherent & 8 & 118 & 0.11 & 0.24 & 0.19 & 0.15 & 0.23 & 0.34 & -- & 50\% \\
    2022-10-06\_D7\_PM5\_P2 & low moving / mixed & 8 & 117 & 0.03 & 0.12 & 0.04 & 0.07 & 0.09 & 0.28 & -- & 100\% \\
    2022-10-06\_D8\_PM3\_M1 & low moving / mixed & 8 & 116 & 0.05 & 0.40 & 0.13 & 0.26 & 0.29 & 0.81 & -- & 33\% \\
    2022-10-06\_D9\_AM1\_R1 & stationary control & 8 & 119 & 0.05 & 0.22 & 0.17 & 0.34 & 0.16 & 0.23 & 1.00 & 71\% \\
    2022-10-06\_D9\_AM3\_S2 & high moving / mixed & 8 & 117 & 0.12 & 0.45 & 0.39 & 0.25 & 0.46 & 0.59 & -- & 80\% \\
    all cases &  & 72 & 1064 & 0.06 & 0.38 & 0.28 & 0.26 & 0.27 & 0.53 & 1.00 & 61\% \\
    \bottomrule
  \end{tabular}
  \end{adjustbox}
\end{table}

\begin{figure}[htbp]\centering
  \includegraphics[width=\textwidth]{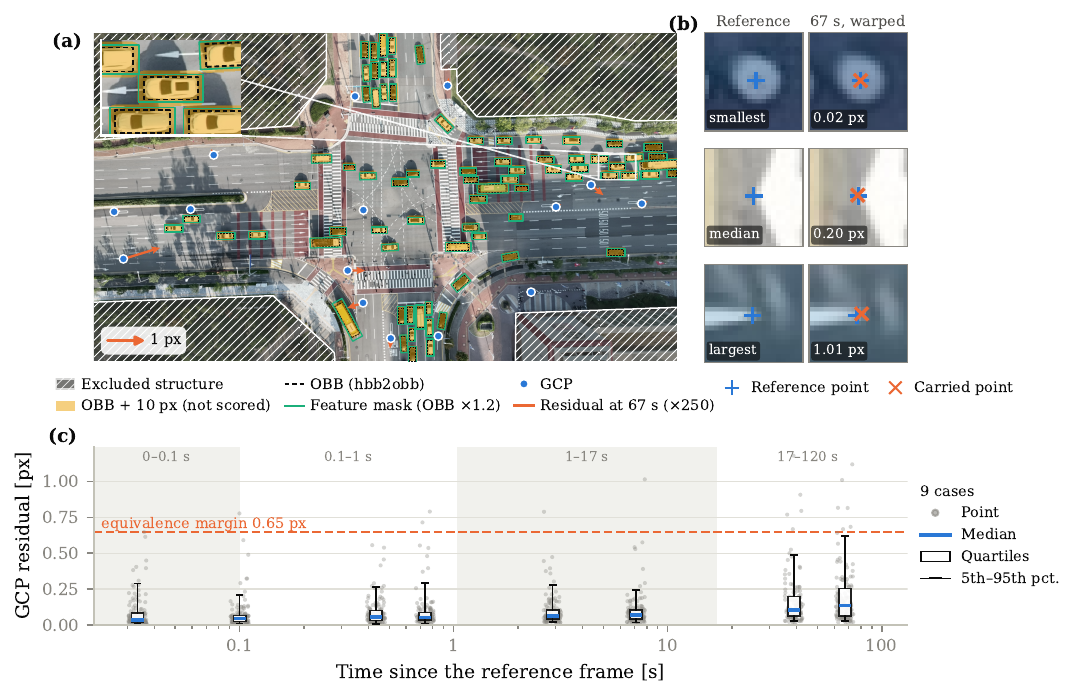}
  \caption{Audit of the reference estimator against \acp{GCP}. (a)~Reference frame of case 2022-10-06\_D9\_AM3\_S2 (\Cref{tab:gcp-audit}) with its
\acp{GCP}, the static structure excluded from the reference-free measures (hatched), every
object's \ac{OBB} as produced by hbb2obb (dashed), the reference estimator's feature mask (solid:
each \ac{OBB} scaled about its center by $1 + m$, \Cref{tab:fixed-parameters}) and the \ac{OBB}
dilated by $\tau$, which the reference-free measures exclude (shaded); the inset magnifies one
object at the frame's resolution. Each arrow is a point's residual at the case's longest audited
offset, drawn \num{250} times its length. (b)~The points with the smallest, median and largest of
those residuals in \num{21}\,$\times$\,\num{21}~px windows: the reference frame beside the audited
frame resampled into reference coordinates by $\Horacle$, with the reference point~($+$) and the
carried point~($\times$), whose distance is the residual. (c)~Every audited point of every case
against its time offset, pooled over the cases at each audited offset (box: quartiles and median;
whiskers: 5th and 95th percentiles), the four audit offset bins shaded alternately; the dashed line
is the equivalence margin $\eps_{\mathrm{floor}}$.}\label{fig:gcp-audit}
\end{figure}

Over \num{1064} points on the \num{72} audited frames of the nine cases, the reference estimator's
residual has a median of \SI{0.06}{px} and a \num{95}th percentile of \SI{0.38}{px}
(\SI{2.1}{\centi\meter} on the ground), ranging over the cases from \num{0.12} to \SI{0.65}{px}
(\Cref{tab:gcp-audit,fig:gcp-audit}). The largest of these is $\eps_{\mathrm{floor}}$, the
equivalence margin of every reference-relative pixel metric pooled over the cases, and each case's
own value starts its $\mathrm{mAA}[\eps_{\mathrm{floor}}, 5]$. Per offset bin the \num{95}th
percentile runs from \SI{0.28}{px} at \num{1}--\num{3} frames to \SI{0.53}{px} at
\num{512}--\num{3600}. The \ac{GCP} repeatability, from the repeat sessions on two cases, is
\SI{1.00}{px} at the \num{95}th percentile, and \SI{61}{\percent} of the points placed by hand lie
within half a pixel of the hidden prediction, the check of \Cref{sec:gcp-audit} that the clicks did
not follow the panel (\Cref{tab:gcp-proposals} counts how each point was placed). The residual can
lie below the repeatability because each later frame's point is placed by following the reference
point's neighborhood, so the error of the reference click moves the point and $\Horacle$'s
prediction alike and drops out; the repeatability is the distance between two unaided clicks of one
feature, noise that only the points placed by hand carry. The floor is far below the dilation
$\tau = \SI{10}{px}$, so the exclusions placed with the reference still cover the objects. On the audited
frames every candidate is also scored against $\Hgcp$ of \Cref{eq:gcp-fit}, which is the primary
geometric evidence for the masking comparison
(\Cref{fig:regime-summary,tab:offline_masking-gcp,tab:realtime_masking-gcp}).

\subsection{Consensus capture}
\label{sec:capture-results}

Capture is read four ways: directly, as a capture fraction of at least \num{0.1} on a pair with a
moving group (\Cref{sec:capture-measure}); as an error of many pixels against the \acp{GCP} where an audited frame falls on such a pair; as error concentrated along the direction of
motion; and as all three receding under masking.

\begin{figure}[htbp]\centering
  \includegraphics[width=\textwidth]{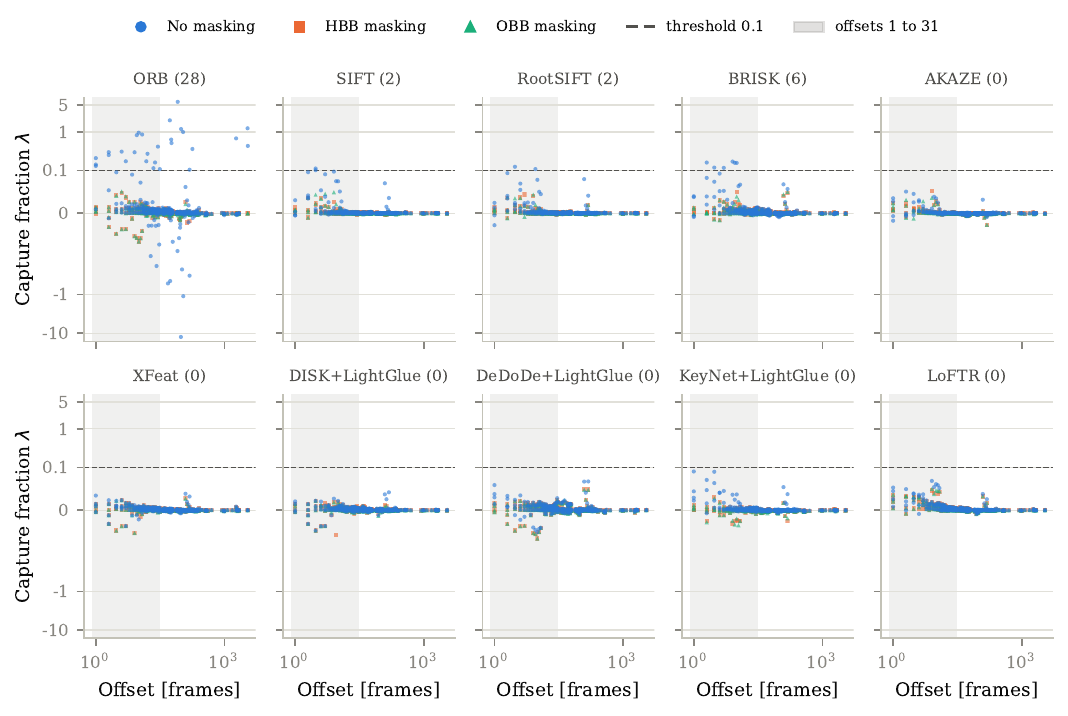}
  \caption{The capture fraction $\lambda$ (\Cref{eq:capture-fraction}) of every real-time
  measurement on a pair with a moving group, against the offset: one panel per pipeline, its title
  counting the captured measurements, and a marker per masking mode. The dashed line is the capture
  threshold of \num{0.1} and the shaded band the survey's offsets of \num{1} to \num{31} frames.
  The axis is linear within the threshold and logarithmic beyond it, so a partial pull, a complete
  capture ($\lambda \approx 1$) and a breakdown far from zero are all visible. The offline regime
  is in \Cref{fig:capture-scatter-offline}, the captured share per offset bin in
  \Cref{fig:capture-by-offset}.}\label{fig:capture-scatter}
\end{figure}

\begin{table}[htbp]
  \centering
  \caption{Part II (real-time regime at downsample ratio 0.18 and 1000 features): consensus capture over the 9 cases, the configurations captured on at least one pair. A pair counts when at least four of the reference frame's objects moved together; an estimate on it is captured when it follows at least 10\% of that group's displacement ($\lambda \geq 0.1$). Cases: cases with a capture; CTE$_{50}$: median corner error of the captured against the reference estimator, and CTE$^{\mathrm{GCP}}_{50}$ against the ground control points on the Audited frames; Inliers and Ratio: median inlier count and ratio of the captured, with their median percentile among the same configuration's estimates at the same case and offset bin (0.5 typical, 1 the most); On moving: median share of their inliers on objects that moved. Descriptive; nothing is bolded. The other 26 configurations are captured on no pair, with a largest $\lambda$ of 0.09. No masked estimate exceeds $\lambda = 0.05$. The identity transform reaches $\lambda = 0.07$, and the smallest displacement of any group is 5\,px. Every configuration, per offset bin: \texttt{realtime\_masking\_capture\_by\_offset.csv}. Every row counts 304 such pairs.}
  \label{tab:realtime_masking-capture}
  \small
  \setlength{\tabcolsep}{4pt}
  \begin{adjustbox}{max width=\textwidth}
  \begin{tabular}{llrrrrrrrrrrrr}
    \toprule
    Pipeline & Masking & Captured & Share & Cases & $\lambda_{\max}$ & CTE$_{50}$ [px] & Audited & CTE$^{\mathrm{GCP}}_{50}$ [px] & Inliers & Inlier rank & Ratio & Ratio rank & On moving \\
    \midrule
    ORB & none & 28 & 9.2\% & 5 & 6.03 & 18.8 & 4 & 4.3 & 290 & 0.58 & 0.16 & 0.58 & 13\% \\
    SIFT & none & 2 & 0.7\% & 2 & 0.11 & 0.9 & 2 & 1.0 & 642 & 0.50 & 0.87 & 0.50 & 0\% \\
    RootSIFT & none & 2 & 0.7\% & 1 & 0.12 & 0.9 & 1 & 1.0 & 564 & 0.69 & 0.85 & 0.44 & 0\% \\
    BRISK & none & 6 & 2.0\% & 1 & 0.17 & 1.6 & 1 & 1.2 & 464 & 0.81 & 0.88 & 0.69 & 3\% \\
    \bottomrule
  \end{tabular}
  \end{adjustbox}
\end{table}

\paragraph{Capture by pipeline and case}
Without masking, ORB is captured on \num{28} of the \num{304} pairs that carry a moving group in the
real-time regime (\num{9.2}\%), in five cases, and on \num{6} of them offline (\num{2.0}\%), in two
(\Cref{fig:capture-scatter,tab:realtime_masking-capture}; offline in
\Cref{fig:capture-scatter-offline} and file \suppfile{offline_masking_capture}). In real time BRISK (\num{6}
pairs), SIFT and RootSIFT (\num{2} each) are captured as well, all at offsets below \num{16} frames
and misaligned by \num{0.9} to \SI{1.6}{px} (CTE$_{50}$ of the captured), with few of their inliers on the
objects found to have moved, a share that undercounts (\Cref{sec:capture-measure}); offline no
other pipeline is. No masked configuration of any pipeline is captured in either regime: the
largest capture fraction any masked estimate reaches is \num{0.05} in real time and \num{0.03}
offline, below the identity's \num{0.07}, against \num{6.03} and \num{0.36} for unmasked ORB. A
higher threshold reduces the count: at
$\lambda \geq 0.25$ unmasked ORB keeps \num{19} captures in real time and \num{4} offline, at
$\lambda \geq 0.5$ \num{11} and none, and at $\lambda \geq 1$ \num{4} and none (released
files \suppfile{realtime_masking_capture} and \suppfile{offline_masking_capture}); it drops
partial captures first, so that most of what remains at the top are the breakdowns described below.

Capture follows the survey's moving-foreground axis. A group of four or more objects that moved
together occurs only in the four high-moving cases and, on \num{13} pairs, in the stationary
control, never in the four low-moving cases, and unmasked ORB is captured in real time in all five
cases that have one
(released files \suppfile{offline_masking_capture_by_case} and
\suppfile{realtime_masking_capture_by_case}; a case is named here by the hover that ends its
name in \Cref{tab:survey-cases}): S2 holds \num{22} of ORB's real-time captures and \num{4} of its
offline ones, U1 and F2 two each in real time, B2 and the control one each, and the control the other two
offline. The control is still at the survey's offsets; it holds a group only several seconds after
the reference, once part of its standing foreground has started to move together, and its captures
lie there. The coherence axis does
not order them, since S2 lies in the mixed cell; the survey averages coherence over its offsets,
while a capture needs one coherent group at one offset. By offset, \num{17} of the \num{28}
real-time captures of ORB lie within the survey's offsets of \num{1} to \num{31} frames, \num{3} of
the \num{4} pairs at one frame among them, and the rest are spread up to the longest bin
(\Cref{fig:capture-scatter,fig:capture-by-offset}); offline, four of six lie within them. Within the survey's offsets
every capture is a pull toward the group: its error at the group lies within \SI{17}{\degree} of the
group's displacement, ORB's within \SI{11}{\degree} (\SI{5}{\degree} offline), up to a capture fraction of \num{0.96}
(\Cref{fig:teaser}; files \suppfile{realtime_masking_capture_populations} and
\suppfile{offline_masking_capture_populations}), an estimate that registers the group back onto where it stood. Beyond them the
real-time captures are of another kind. Apart from one in the control, under a pixel at the group,
they are breakdowns: corner errors of \num{153} to \SI{7085}{px} on at most \num{143} inliers, among
them the four at $\lambda \geq 1$, up to \num{6.03}. A breakdown is a failure of its own: unmasked
real-time ORB's corner error reaches \SI{100}{px} on \num{42} of its \num{1215} pairs, \num{25} of them without a group, and no
other configuration's does in either regime (files \suppfile{realtime_masking_capture} and
\suppfile{offline_masking_capture}).

\begin{figure}[htbp]\centering
  \includegraphics[width=\textwidth]{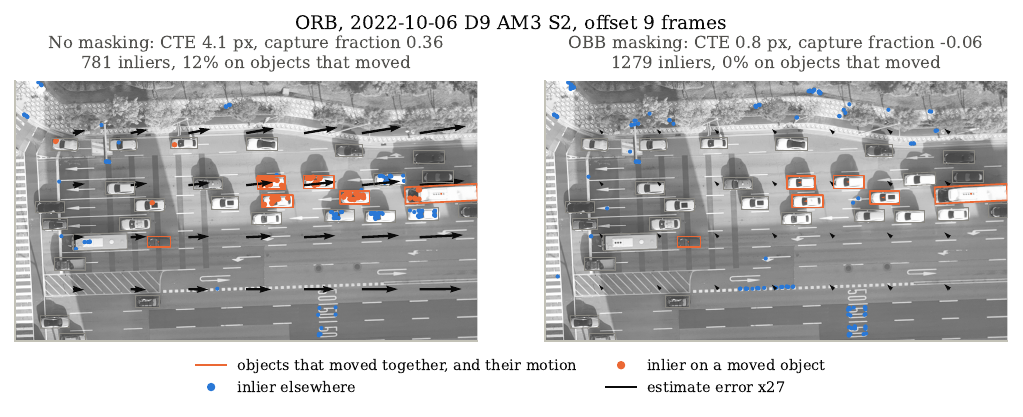}
  \caption{The offline counterpart of \Cref{fig:teaser}: the captured unmasked measurement of the
  offline regime with the largest capture fraction within the survey's offsets of \num{1} to
  \num{31} frames whose registration reproduces its stored estimate, beside the same pipeline under
  \ac{OBB} masking, drawn as in \Cref{fig:teaser}.}\label{fig:capture-example}
\end{figure}

\paragraph{Examples of captured estimates}
Both examples are unmasked ORB on S2. In real time, at an offset of ten frames
(\Cref{fig:teaser}), the estimate follows the group almost completely, with a capture fraction of
\num{0.96}, and its corner error of \SI{10.0}{px} against the reference is not confined to the
group: the error arrows carry the group's motion across the whole crop, so every position measured
in the frame inherits it. Offline, at nine frames (\Cref{fig:capture-example}), the pull is partial, a
capture fraction of \num{0.36}, and the error \SI{4.1}{px}, on \num{781} inliers, \SI{12}{\percent}
on objects that moved, against \SI{0.8}{px} on \num{1279} inliers under masking. In both the masked
estimate has more inliers, because ORB spends the share of its budget the mask frees on the ground
(\Cref{sec:pipelines}).

\paragraph{Error magnitude}
No pipeline fails on any pair in either regime. Against the \acp{GCP}, every
configuration's median corner error lies between \num{0.21} and \SI{0.57}{px} offline, below the
\ac{GCP} repeatability, and its \num{90}th percentile at most \SI{1.15}{px}, reached by unmasked
ORB. In real time the medians lie between \num{0.51} and \SI{1.67}{px}; every configuration but
ORB stays within \SI{1.94}{px} at the \num{90}th percentile, ORB reaches \num{2.03} and
\SI{2.07}{px} with \ac{HBB} and \ac{OBB} masks, and \SI{152.57}{px} without
(\Cref{tab:offline_masking-gcp,tab:realtime_masking-gcp}). That value is set by one frame: the percentile
is taken over each case's eight audited frames and averaged over the cases, and on one audited
frame of S2, a pair without a moving group, unmasked ORB's estimate broke down by \SI{4469.4}{px}
(CTE$^{\mathrm{GCP}}_{\max}$). Where a captured estimate falls on an audited frame, the \acp{GCP} confirm the capture independently of the reference estimator: the four such
estimates of unmasked ORB in real time lie a median of \SI{4.3}{px} from them, where masked ORB's
median is \SI{1.17}{px}, and the one offline \SI{4.7}{px}, against \SI{0.43}{px} masked (Audited
and CTE$^{\mathrm{GCP}}_{50}$ in \Cref{tab:realtime_masking-capture} and file \suppfile{offline_masking_capture}).
The captured SIFT, RootSIFT and BRISK estimates on audited frames lie \num{1.0} to \SI{1.2}{px}
from them, at the repeatability, which cannot resolve a pull of that size. Against the reference
estimator, unmasked ORB's \num{90}th percentile is \SI{163.44}{px} in real time
(\Cref{tab:realtime_masking-dynamic}). For scale, the unstabilized frame lies \SI{6.67}{px} from the
reference at the median (\Cref{tab:offline_masking-dynamic}).

\paragraph{Direction of the error}
The identity transform's motion alignment is \num{0.478}. In real time, unmasked ORB's is
\num{0.579} against \num{0.542} under \ac{OBB} masking, and the planned contrast of $A$ without
masking against \ac{OBB} masking is \num{+0.035} for ORB, with a \SI{95}{\percent} interval of
\num{+0.014} to \num{+0.064} and a positive value in five of the six cases where the axis is
defined; Holm's adjustment over the ten pipelines leaves it short of the rule. For the average
pipeline it is \num{+0.005} in real time and \num{-0.003} offline
(\Cref{fig:motion-alignment}; files \suppfile{offline_masking_mechanism} and
\suppfile{realtime_masking_mechanism}). $A$
pools every pair, and capture touches a few percent of them, so it moves little; the capture
fraction, which reads the group itself, is the sharper of the two instruments.

\begin{figure}[htbp]\centering
  \includegraphics[width=0.9\textwidth]{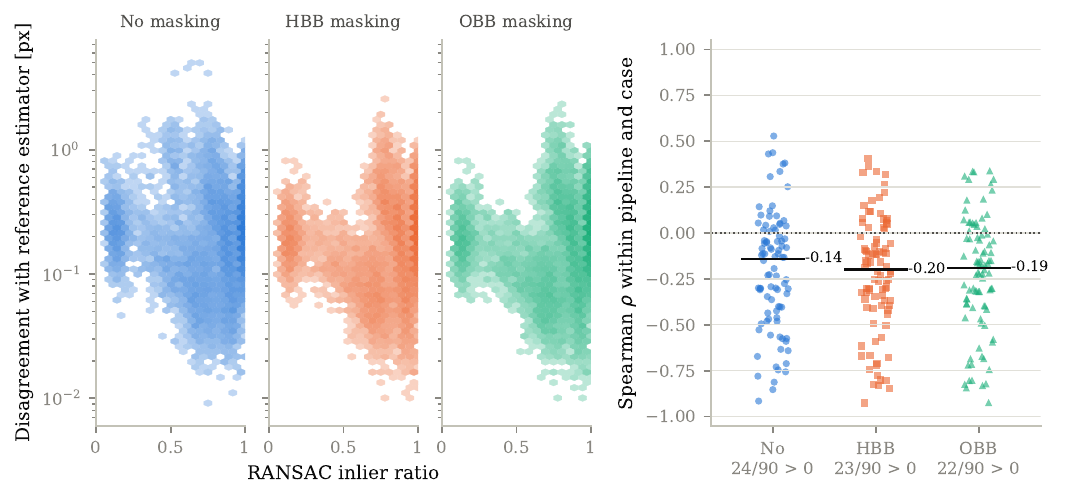}\\[0.5em]
  \includegraphics[width=0.9\textwidth]{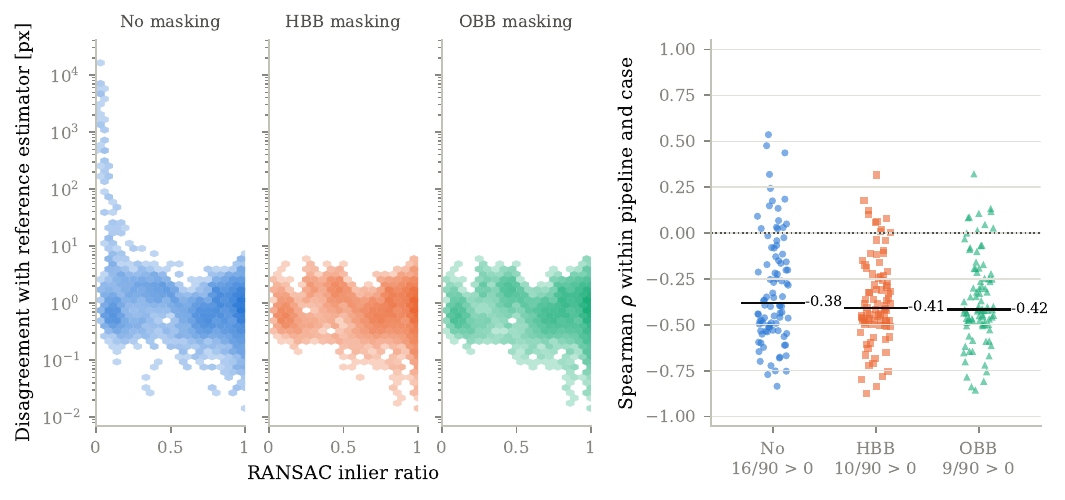}
  \caption{Inlier ratio against the disagreement with the reference estimator, offline (top) and
  in real time (bottom). In each row, left, the density of frame pairs and pipelines, one panel per
  masking mode, darker where more of them fall, on a logarithmic scale of the count; right, the
  Spearman coefficient between the two within each pipeline and case, one point per group, with
  the median printed beside it and the number of positive coefficients under each masking mode.
  A negative coefficient means that a higher inlier ratio goes with a smaller
  error.}\label{fig:inlier-vs-quality}
\end{figure}

\paragraph{Inlier statistics do not reveal capture}
Captured estimates are not exposed by their inliers. Pooled over offsets, unmasked ORB's captured
real-time estimates have a median of \num{290.5} inliers against \num{504.5} for its other pairs with a
group, but inliers fall with offset, and ranked against ORB's own estimates on the pairs of the same
case and offset bin they sit at \num{0.58} in both count and ratio; BRISK's rank
\num{0.81} in count and offline ORB's \num{0.93} in both
(\Cref{tab:realtime_masking-capture}; file \suppfile{offline_masking_capture}). The sharper
comparison is with the same case at the nearest offsets, two on either side, whose estimates are
neither captured nor broken down. Against those, unmasked ORB's pulls carry \num{1.08} times the inliers at
\num{1.08} times the ratio in real time, and \num{1.03} and \num{1.04} times offline, at a median
disagreement from the reference of \num{9.7} and \SI{4.4}{px} (\num{18.8} and \SI{4.1}{px} over all
captures) where unmasked ORB's over all pairs is \num{2.42} and \SI{0.54}{px}
(\Cref{tab:realtime_masking-dynamic,tab:offline_masking-dynamic}):
nothing in their inlier statistics sets them apart. The breakdowns carry half (\num{0.51} and \num{0.50} times), so theirs do. A
captured fit is, by construction, the fit the consensus preferred. Within each pipeline and case the
inlier ratio does order estimates by accuracy on the whole: the median Spearman coefficient between
ratio and disagreement is \num{-0.38} without masking in real time and \num{-0.14} offline,
positive in \num{16} and \num{24} of the \num{90} pipeline and case groups
(\Cref{fig:inlier-vs-quality}). The pulls are the exception that this ordering cannot
reveal. Across pipelines the inlier ratio records the estimator and its threshold rather than
accuracy, as in Part~I's estimator sweep (\Cref{fig:estimator-thresholds}), so choosing a pipeline,
which is a comparison between pipelines, cannot rest on it even where it tracks accuracy within one.

\subsection{The two regimes}
\label{sec:ranking-results}

\begin{figure}[htbp]\centering
  \includegraphics[width=\textwidth]{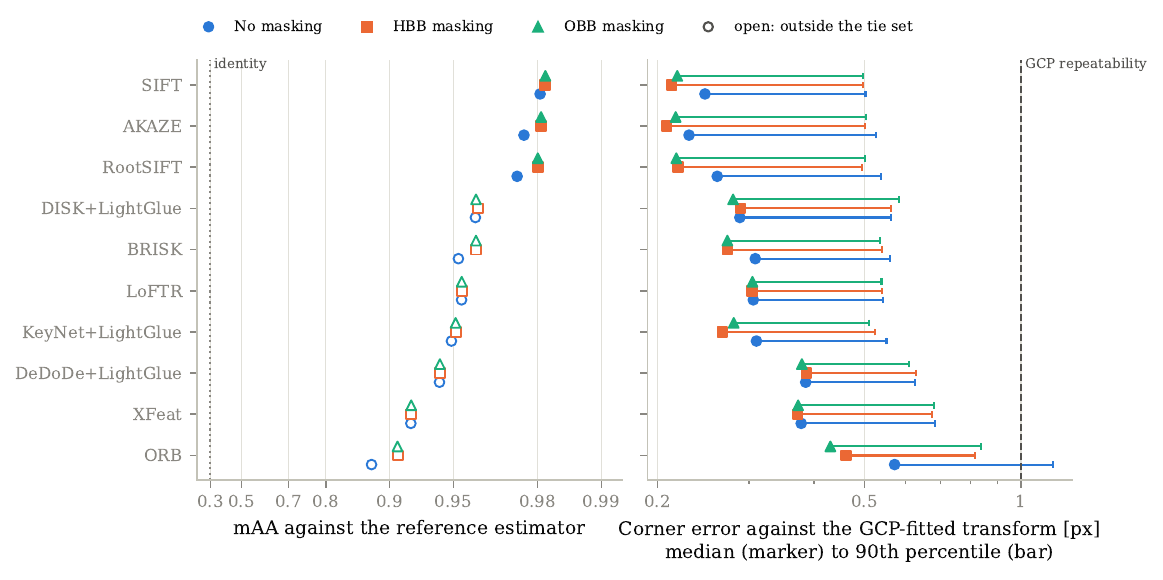}\\[0.6em]
  \includegraphics[width=\textwidth]{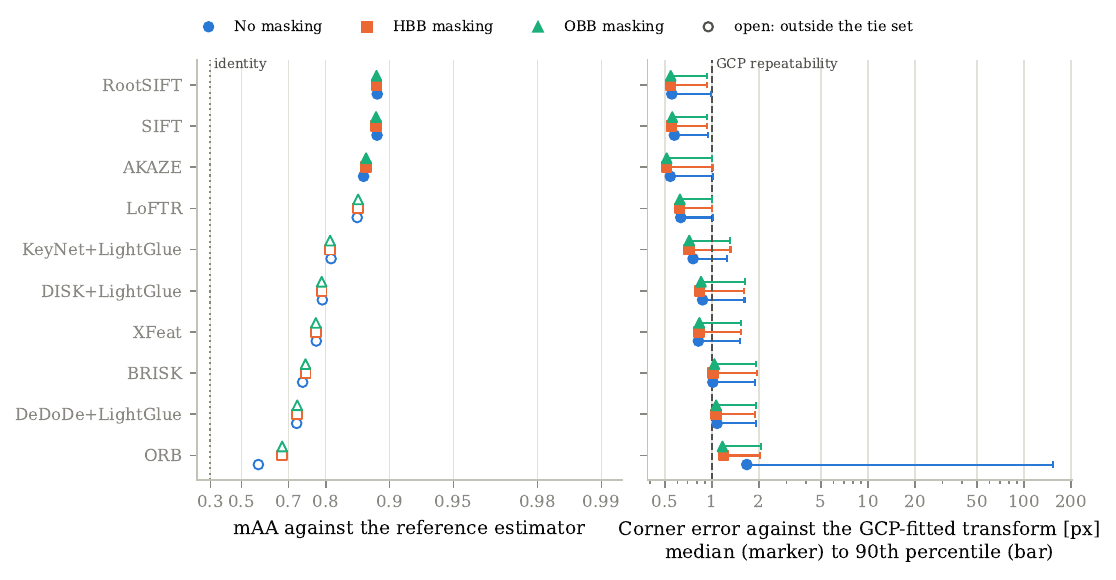}
  \caption{Every candidate in the offline (top) and real-time (bottom) regime, pooled over the nine
  cases with equal weight per case. Left, mAA over $[0, 5]$\,px against the reference estimator, on
  a logarithmic scale of $1 - \text{mAA}$, the identity dotted; right, the \ac{CTE} against the homography fitted to the
  \acp{GCP} of the audited frames, the marker at the median and the bar ending at the
  \num{90}th percentile, with the \ac{GCP} repeatability dashed. A marker is filled when the
  configuration is in the column's tie set under the analysis plan (\Cref{sec:analysis-plan}) and
  open otherwise. Unmasked ORB's real-time bar is set by one audited frame on which its estimate
  broke down (\Cref{sec:capture-results}). Full rows in
  \Cref{tab:offline_masking-dynamic,tab:realtime_masking-dynamic,tab:offline_masking-gcp,tab:realtime_masking-gcp}.}\label{fig:regime-summary}
\end{figure}

\begin{table}[htbp]
  \centering
  \caption{Part II (offline regime at downsample ratio 0.5 and 2000 features; dynamic benchmark, 9 cases): every candidate against the reference estimator. mAA, integrated over $[0, 5]$ px, and CTE are reference-relative; the largest of the cases' audited residuals, 0.65 px, is the equivalence margin of the reference-relative pixel columns, below which a difference is not resolvable, and mAA$_{\varepsilon_{\mathrm{floor}}}$ integrates from each case's own residual. SSIM, ZNCC, MI and GOS are reference-free, over the static background; IoU and OCE compare the current frame's boxes as placed by the candidate and by the reference estimator. Align (0.5 for isotropic error) and the inlier ratio are diagnostics, never bolded. Identity is the unstabilized frame, also what a failed pipeline delivers. The RootSIFT and SIFT rows share the reference estimator's keypoint detector (SIFT, with OBB masking), so their reference-relative columns are the least independent of it. Bold marks the best value of each column and every value not distinguishable from it: a paired difference to the best, over the 9 cases, whose 95\% cluster-bootstrap interval includes zero or whose estimate is below the equivalence margin (0.01 for mAA, 0.65 px for the reference-relative pixel columns). With 9 cases the smallest attainable sign-flip $p$ is 0.0039. The reference-free columns are computed over a median 56.6\% to 56.7\% of the frame. No configuration failed on any pair. Every row pools 9 cases.}
  \label{tab:offline_masking-dynamic}
  \small
  \setlength{\tabcolsep}{4pt}
  \begin{adjustbox}{max width=\textwidth}
  \begin{tabular}{llrrrrrrrrrrrrr}
    \toprule
    Pipeline & Masking & mAA & mAA$_{\varepsilon_{\mathrm{floor}}}$ & CTE$_{50}$ [px] & CTE$_{90}$ [px] & SSIM & ZNCC & MI [bits] & GOS & IoU & OCE [px] & Align & Inlier ratio & Time [s] \\
    \midrule
    Reference & -- & 1.0000 & 1.0000 & 0.00 & 0.00 & 0.9505 & 0.9890 & 3.280 & 0.9286 & 1.0000 & 0.00 & -- & -- & 0.000 \\
    SIFT & OBB & \textbf{0.9816} & 0.9996 & \textbf{0.08} & \textbf{0.16} & \textbf{0.9503} & 0.9890 & \textbf{3.277} & \textbf{0.9282} & \textbf{0.9983} & \textbf{0.04} & 0.618 & 0.641 & 0.069 \\
    SIFT & HBB & \textbf{0.9816} & 0.9997 & \textbf{0.08} & \textbf{0.16} & \textbf{0.9503} & 0.9890 & \textbf{3.277} & \textbf{0.9282} & \textbf{0.9983} & \textbf{0.04} & 0.616 & 0.640 & 0.071 \\
    AKAZE & OBB & \textbf{0.9807} & 0.9993 & \textbf{0.08} & \textbf{0.17} & \textbf{0.9503} & \textbf{0.9890} & \textbf{3.278} & \textbf{0.9282} & \textbf{0.9981} & \textbf{0.04} & 0.616 & 0.681 & 0.100 \\
    AKAZE & HBB & \textbf{0.9807} & 0.9993 & \textbf{0.08} & \textbf{0.18} & \textbf{0.9503} & \textbf{0.9890} & \textbf{3.278} & \textbf{0.9282} & \textbf{0.9981} & \textbf{0.04} & 0.617 & 0.680 & 0.101 \\
    SIFT & none & \textbf{0.9805} & 0.9996 & \textbf{0.08} & \textbf{0.16} & 0.9503 & 0.9890 & \textbf{3.276} & 0.9281 & 0.9982 & \textbf{0.04} & 0.612 & 0.622 & 0.079 \\
    RootSIFT & HBB & \textbf{0.9801} & 0.9995 & \textbf{0.09} & \textbf{0.17} & \textbf{0.9503} & \textbf{0.9890} & \textbf{3.278} & \textbf{0.9282} & 0.9982 & \textbf{0.04} & 0.628 & 0.914 & 0.085 \\
    RootSIFT & OBB & \textbf{0.9800} & 0.9995 & \textbf{0.09} & \textbf{0.17} & \textbf{0.9503} & \textbf{0.9890} & \textbf{3.278} & \textbf{0.9282} & 0.9982 & \textbf{0.04} & 0.631 & 0.914 & 0.087 \\
    AKAZE & none & \textbf{0.9768} & 0.9985 & \textbf{0.10} & \textbf{0.21} & \textbf{0.9503} & \textbf{0.9890} & \textbf{3.277} & \textbf{0.9282} & 0.9979 & \textbf{0.05} & 0.610 & 0.690 & 0.097 \\
    RootSIFT & none & \textbf{0.9750} & 0.9982 & \textbf{0.10} & \textbf{0.21} & 0.9500 & 0.9890 & 3.271 & 0.9277 & 0.9978 & \textbf{0.05} & 0.620 & 0.891 & 0.085 \\
    DISK+LightGlue & HBB & 0.9618 & 0.9961 & \textbf{0.16} & \textbf{0.30} & 0.9487 & 0.9886 & 3.246 & 0.9259 & 0.9957 & \textbf{0.09} & 0.559 & 0.989 & 0.071 \\
    DISK+LightGlue & OBB & 0.9609 & 0.9956 & \textbf{0.16} & \textbf{0.33} & 0.9486 & 0.9885 & 3.245 & 0.9257 & 0.9956 & \textbf{0.09} & 0.560 & 0.989 & 0.073 \\
    BRISK & OBB & 0.9609 & 0.9960 & \textbf{0.19} & \textbf{0.30} & 0.9498 & 0.9889 & 3.265 & 0.9275 & 0.9961 & \textbf{0.08} & 0.606 & 0.178 & 0.303 \\
    BRISK & HBB & 0.9609 & 0.9960 & \textbf{0.18} & \textbf{0.30} & 0.9498 & 0.9889 & 3.265 & 0.9275 & 0.9961 & \textbf{0.08} & 0.605 & 0.178 & 0.301 \\
    DISK+LightGlue & none & 0.9607 & 0.9953 & \textbf{0.16} & \textbf{0.34} & 0.9486 & 0.9885 & 3.244 & 0.9257 & 0.9955 & \textbf{0.10} & 0.558 & 0.983 & 0.077 \\
    LoFTR & HBB & 0.9544 & 0.9950 & \textbf{0.22} & \textbf{0.33} & 0.9482 & 0.9884 & 3.228 & 0.9251 & 0.9937 & \textbf{0.13} & 0.567 & 0.990 & 0.448 \\
    LoFTR & OBB & 0.9544 & 0.9950 & \textbf{0.22} & \textbf{0.33} & 0.9482 & 0.9884 & 3.228 & 0.9251 & 0.9937 & \textbf{0.13} & 0.567 & 0.990 & 0.449 \\
    LoFTR & none & 0.9543 & 0.9950 & \textbf{0.22} & \textbf{0.33} & 0.9482 & 0.9884 & 3.227 & 0.9251 & 0.9937 & \textbf{0.13} & 0.568 & 0.983 & 0.446 \\
    BRISK & none & 0.9527 & 0.9923 & \textbf{0.22} & \textbf{0.36} & 0.9494 & 0.9888 & 3.256 & 0.9268 & 0.9954 & \textbf{0.10} & 0.596 & 0.170 & 0.314 \\
    KeyNet+LightGlue & HBB & 0.9515 & 0.9928 & \textbf{0.23} & \textbf{0.36} & 0.9494 & 0.9888 & 3.256 & 0.9268 & 0.9949 & \textbf{0.11} & 0.634 & 0.998 & 0.141 \\
    KeyNet+LightGlue & OBB & 0.9512 & 0.9927 & \textbf{0.23} & \textbf{0.37} & 0.9494 & 0.9888 & 3.257 & 0.9268 & 0.9949 & \textbf{0.11} & 0.633 & 0.998 & 0.143 \\
    KeyNet+LightGlue & none & 0.9489 & 0.9910 & \textbf{0.24} & \textbf{0.38} & 0.9491 & 0.9887 & 3.251 & 0.9263 & 0.9946 & \textbf{0.12} & 0.629 & 0.985 & 0.143 \\
    DeDoDe+LightGlue & HBB & 0.9422 & 0.9854 & \textbf{0.25} & \textbf{0.50} & 0.9451 & 0.9875 & 3.189 & 0.9206 & 0.9924 & \textbf{0.16} & 0.565 & 0.876 & 0.205 \\
    DeDoDe+LightGlue & OBB & 0.9421 & 0.9854 & \textbf{0.25} & \textbf{0.49} & 0.9451 & 0.9875 & 3.189 & 0.9207 & 0.9924 & \textbf{0.16} & 0.566 & 0.876 & 0.206 \\
    DeDoDe+LightGlue & none & 0.9419 & 0.9850 & \textbf{0.25} & \textbf{0.50} & 0.9451 & 0.9876 & 3.189 & 0.9206 & 0.9923 & \textbf{0.16} & 0.563 & 0.868 & 0.211 \\
    XFeat & OBB & 0.9209 & 0.9717 & \textbf{0.37} & \textbf{0.62} & 0.9441 & 0.9874 & 3.173 & 0.9188 & 0.9898 & \textbf{0.21} & 0.550 & 0.999 & 0.029 \\
    XFeat & HBB & 0.9208 & 0.9715 & \textbf{0.36} & \textbf{0.63} & 0.9441 & 0.9874 & 3.173 & 0.9187 & 0.9898 & \textbf{0.21} & 0.549 & 0.999 & \textbf{0.026} \\
    XFeat & none & 0.9206 & 0.9715 & \textbf{0.37} & \textbf{0.62} & 0.9441 & 0.9874 & 3.172 & 0.9188 & 0.9898 & \textbf{0.22} & 0.547 & 0.999 & 0.030 \\
    ORB & HBB & 0.9089 & 0.9604 & \textbf{0.39} & \textbf{0.79} & 0.9379 & 0.9851 & 3.129 & 0.9092 & 0.9889 & \textbf{0.24} & 0.581 & 0.787 & 0.033 \\
    ORB & OBB & 0.9082 & 0.9597 & \textbf{0.39} & \textbf{0.79} & 0.9376 & 0.9850 & 3.128 & 0.9086 & 0.9889 & \textbf{0.24} & 0.580 & 0.787 & 0.035 \\
    ORB & none & 0.8783 & 0.9323 & \textbf{0.54} & 1.03 & 0.9363 & 0.9849 & 3.107 & 0.9067 & 0.9878 & \textbf{0.27} & 0.590 & 0.710 & 0.031 \\
    Identity & -- & 0.2957 & 0.3145 & 6.67 & 22.05 & 0.6053 & 0.7975 & 1.794 & 0.3192 & 0.7995 & 6.01 & 0.478 & -- & 0.000 \\
    \bottomrule
  \end{tabular}
  \end{adjustbox}
\end{table}

\begin{table}[htbp]
  \centering
  \caption{Part II (real-time regime at downsample ratio 0.18 and 1000 features; dynamic benchmark, 9 cases): every candidate against the reference estimator. mAA, integrated over $[0, 5]$ px, and CTE are reference-relative; the largest of the cases' audited residuals, 0.65 px, is the equivalence margin of the reference-relative pixel columns, below which a difference is not resolvable, and mAA$_{\varepsilon_{\mathrm{floor}}}$ integrates from each case's own residual. SSIM, ZNCC, MI and GOS are reference-free, over the static background; IoU and OCE compare the current frame's boxes as placed by the candidate and by the reference estimator. Align (0.5 for isotropic error) and the inlier ratio are diagnostics, never bolded. Identity is the unstabilized frame, also what a failed pipeline delivers. The RootSIFT and SIFT rows share the reference estimator's keypoint detector (SIFT, with OBB masking), so their reference-relative columns are the least independent of it. Bold marks the best value of each column and every value not distinguishable from it: a paired difference to the best, over the 9 cases, whose 95\% cluster-bootstrap interval includes zero or whose estimate is below the equivalence margin (0.01 for mAA, 0.65 px for the reference-relative pixel columns). With 9 cases the smallest attainable sign-flip $p$ is 0.0039. The reference-free columns are computed over a median 56.5\% to 56.7\% of the frame. No configuration failed on any pair. Every row pools 9 cases.}
  \label{tab:realtime_masking-dynamic}
  \small
  \setlength{\tabcolsep}{4pt}
  \begin{adjustbox}{max width=\textwidth}
  \begin{tabular}{llrrrrrrrrrrrrr}
    \toprule
    Pipeline & Masking & mAA & mAA$_{\varepsilon_{\mathrm{floor}}}$ & CTE$_{50}$ [px] & CTE$_{90}$ [px] & SSIM & ZNCC & MI [bits] & GOS & IoU & OCE [px] & Align & Inlier ratio & Time [s] \\
    \midrule
    Reference & -- & 1.0000 & 1.0000 & 0.00 & 0.00 & 0.9505 & 0.9890 & 3.280 & 0.9286 & 1.0000 & 0.00 & -- & -- & 0.000 \\
    RootSIFT & none & \textbf{0.8856} & 0.9409 & \textbf{0.54} & \textbf{0.89} & \textbf{0.9392} & \textbf{0.9861} & \textbf{3.120} & \textbf{0.9108} & \textbf{0.9876} & \textbf{0.26} & 0.620 & 0.760 & 0.020 \\
    SIFT & none & \textbf{0.8853} & 0.9409 & \textbf{0.55} & \textbf{0.89} & \textbf{0.9391} & \textbf{0.9860} & \textbf{3.119} & \textbf{0.9106} & \textbf{0.9875} & \textbf{0.26} & 0.622 & 0.753 & 0.020 \\
    RootSIFT & OBB & \textbf{0.8847} & 0.9396 & \textbf{0.55} & \textbf{0.91} & 0.9388 & 0.9859 & \textbf{3.116} & 0.9101 & 0.9873 & \textbf{0.27} & 0.617 & 0.763 & 0.022 \\
    RootSIFT & HBB & \textbf{0.8846} & 0.9395 & \textbf{0.55} & \textbf{0.90} & 0.9387 & 0.9859 & \textbf{3.116} & 0.9101 & 0.9873 & \textbf{0.27} & 0.617 & 0.763 & 0.021 \\
    SIFT & HBB & \textbf{0.8842} & 0.9393 & \textbf{0.56} & \textbf{0.91} & 0.9388 & 0.9859 & \textbf{3.116} & 0.9102 & 0.9872 & \textbf{0.27} & 0.618 & 0.758 & 0.021 \\
    SIFT & OBB & \textbf{0.8841} & 0.9393 & \textbf{0.56} & \textbf{0.91} & \textbf{0.9389} & \textbf{0.9859} & \textbf{3.116} & \textbf{0.9103} & 0.9873 & \textbf{0.27} & 0.618 & 0.757 & 0.022 \\
    AKAZE & OBB & \textbf{0.8708} & 0.9256 & \textbf{0.63} & \textbf{1.02} & \textbf{0.9352} & \textbf{0.9845} & \textbf{3.091} & \textbf{0.9039} & 0.9852 & \textbf{0.31} & 0.573 & 0.181 & 0.020 \\
    AKAZE & HBB & \textbf{0.8706} & 0.9255 & \textbf{0.63} & \textbf{1.02} & \textbf{0.9351} & \textbf{0.9845} & \textbf{3.091} & \textbf{0.9036} & 0.9852 & \textbf{0.31} & 0.571 & 0.181 & 0.019 \\
    AKAZE & none & \textbf{0.8673} & 0.9223 & \textbf{0.66} & \textbf{1.04} & \textbf{0.9348} & \textbf{0.9844} & \textbf{3.089} & \textbf{0.9031} & 0.9851 & \textbf{0.31} & 0.578 & 0.182 & 0.018 \\
    LoFTR & HBB & 0.8592 & 0.9155 & \textbf{0.69} & \textbf{1.06} & 0.9243 & 0.9816 & 2.963 & 0.8879 & 0.9788 & \textbf{0.45} & 0.584 & 0.981 & 0.049 \\
    LoFTR & OBB & 0.8591 & 0.9154 & \textbf{0.69} & \textbf{1.06} & 0.9243 & 0.9816 & 2.962 & 0.8877 & 0.9787 & \textbf{0.45} & 0.584 & 0.981 & 0.051 \\
    LoFTR & none & 0.8578 & 0.9142 & \textbf{0.69} & \textbf{1.07} & 0.9238 & 0.9815 & 2.958 & 0.8870 & 0.9785 & \textbf{0.46} & 0.586 & 0.975 & 0.057 \\
    KeyNet+LightGlue & none & 0.8110 & 0.8653 & \textbf{0.94} & \textbf{1.42} & 0.9180 & 0.9792 & 2.929 & 0.8759 & 0.9763 & \textbf{0.49} & 0.536 & 0.986 & 0.056 \\
    KeyNet+LightGlue & OBB & 0.8090 & 0.8628 & \textbf{0.94} & \textbf{1.43} & 0.9170 & 0.9788 & 2.928 & 0.8738 & 0.9758 & \textbf{0.51} & 0.537 & 0.991 & 0.051 \\
    KeyNet+LightGlue & HBB & 0.8085 & 0.8623 & \textbf{0.94} & \textbf{1.44} & 0.9166 & 0.9787 & 2.926 & 0.8731 & 0.9757 & \textbf{0.51} & 0.536 & 0.991 & 0.052 \\
    DISK+LightGlue & none & 0.7922 & 0.8443 & \textbf{0.99} & 1.73 & 0.9181 & 0.9798 & 2.931 & 0.8772 & 0.9773 & \textbf{0.48} & 0.564 & 0.956 & 0.024 \\
    DISK+LightGlue & OBB & 0.7908 & 0.8428 & \textbf{1.00} & 1.74 & 0.9179 & 0.9798 & 2.932 & 0.8767 & 0.9774 & \textbf{0.48} & 0.560 & 0.960 & 0.025 \\
    DISK+LightGlue & HBB & 0.7905 & 0.8423 & \textbf{1.00} & 1.73 & 0.9177 & 0.9798 & 2.930 & 0.8765 & 0.9773 & \textbf{0.48} & 0.560 & 0.961 & 0.024 \\
    XFeat & none & 0.7780 & 0.8291 & \textbf{1.09} & 1.68 & 0.8978 & 0.9742 & 2.797 & 0.8432 & 0.9695 & \textbf{0.66} & 0.540 & 0.823 & 0.011 \\
    XFeat & OBB & 0.7770 & 0.8277 & \textbf{1.08} & 1.71 & 0.8991 & 0.9746 & 2.804 & 0.8455 & 0.9700 & \textbf{0.65} & 0.542 & 0.821 & 0.012 \\
    XFeat & HBB & 0.7768 & 0.8275 & \textbf{1.09} & 1.72 & 0.8988 & 0.9744 & 2.803 & 0.8449 & 0.9700 & \textbf{0.65} & 0.543 & 0.821 & \textbf{0.011} \\
    BRISK & HBB & 0.7506 & 0.8004 & \textbf{1.20} & 2.00 & 0.9090 & 0.9774 & 2.873 & 0.8616 & 0.9751 & \textbf{0.54} & 0.574 & 0.908 & 0.083 \\
    BRISK & OBB & 0.7504 & 0.8002 & \textbf{1.20} & 2.01 & 0.9092 & 0.9775 & 2.874 & 0.8621 & 0.9751 & \textbf{0.54} & 0.574 & 0.908 & 0.085 \\
    BRISK & none & 0.7426 & 0.7928 & 1.21 & 2.04 & 0.9064 & 0.9768 & 2.853 & 0.8573 & 0.9745 & \textbf{0.55} & 0.573 & 0.905 & 0.086 \\
    DeDoDe+LightGlue & HBB & 0.7272 & 0.7753 & 1.31 & 2.22 & 0.8898 & 0.9725 & 2.775 & 0.8295 & 0.9693 & \textbf{0.67} & 0.527 & 0.985 & 0.035 \\
    DeDoDe+LightGlue & OBB & 0.7270 & 0.7752 & 1.30 & 2.25 & 0.8899 & 0.9724 & 2.775 & 0.8297 & 0.9693 & \textbf{0.67} & 0.528 & 0.985 & 0.036 \\
    DeDoDe+LightGlue & none & 0.7253 & 0.7734 & 1.31 & 2.25 & 0.8890 & 0.9723 & 2.769 & 0.8281 & 0.9694 & \textbf{0.67} & 0.522 & 0.979 & 0.035 \\
    ORB & HBB & 0.6787 & 0.7238 & 1.51 & 2.53 & 0.8614 & 0.9608 & 2.632 & 0.7758 & 0.9621 & \textbf{0.83} & 0.542 & 0.449 & 0.012 \\
    ORB & OBB & 0.6785 & 0.7235 & 1.53 & 2.57 & 0.8633 & 0.9617 & 2.638 & 0.7798 & 0.9625 & \textbf{0.82} & 0.542 & 0.448 & 0.013 \\
    ORB & none & 0.5835 & 0.6209 & 2.42 & 163.44 & 0.8175 & 0.9310 & 2.460 & 0.7040 & 0.9226 & 31.73 & 0.579 & 0.352 & \textbf{0.010} \\
    Identity & -- & 0.2957 & 0.3145 & 6.67 & 22.05 & 0.6053 & 0.7975 & 1.794 & 0.3192 & 0.7995 & 6.01 & 0.478 & -- & 0.000 \\
    \bottomrule
  \end{tabular}
  \end{adjustbox}
\end{table}

\begin{table}[htbp]
  \centering
  \caption{Part II (offline regime at downsample ratio 0.5 and 2000 features) candidates against the homography fitted to each audited frame's ground control points, which involves no reference estimator: CTE is the corner transfer error, IoU and OCE compare the frame's object boxes as placed by the candidate and by the fitted transform. CTE$_{50}$ and CTE$_{90}$ are per case and averaged over the cases; CTE$_{\max}$ is the largest over every audited frame. Every row is scored on 72 audited frames and 1064 ground control points. Bold marks the best value of each column and every value not distinguishable from it: a paired difference to the best, over the 9 cases, whose 95\% cluster-bootstrap interval includes zero or whose estimate is below the equivalence margin (1.00 px, the GCP repeatability, for the pixel columns, none for IoU). With 9 cases the smallest attainable sign-flip $p$ is 0.0039.}
  \label{tab:offline_masking-gcp}
  \begin{adjustbox}{max width=\textwidth}
  \begin{tabular}{llrrrrr}
    \toprule
    Pipeline & Masking & CTE$^{\mathrm{GCP}}_{50}$ [px] & CTE$^{\mathrm{GCP}}_{90}$ [px] & CTE$^{\mathrm{GCP}}_{\max}$ [px] & IoU$^{\mathrm{GCP}}$ & OCE$^{\mathrm{GCP}}$ [px] \\
    \midrule
    AKAZE & HBB & \textbf{0.21} & \textbf{0.50} & 1.38 & \textbf{0.9953} & \textbf{0.10} \\
    SIFT & HBB & \textbf{0.21} & \textbf{0.50} & 1.29 & \textbf{0.9954} & \textbf{0.10} \\
    AKAZE & OBB & \textbf{0.22} & \textbf{0.50} & 1.38 & \textbf{0.9954} & \textbf{0.10} \\
    RootSIFT & OBB & \textbf{0.22} & \textbf{0.50} & 1.27 & \textbf{0.9954} & \textbf{0.10} \\
    SIFT & OBB & \textbf{0.22} & \textbf{0.50} & 1.29 & \textbf{0.9954} & \textbf{0.10} \\
    RootSIFT & HBB & \textbf{0.22} & \textbf{0.49} & 1.27 & \textbf{0.9954} & \textbf{0.10} \\
    AKAZE & none & \textbf{0.23} & \textbf{0.53} & 1.44 & \textbf{0.9952} & \textbf{0.11} \\
    SIFT & none & \textbf{0.25} & \textbf{0.50} & 1.27 & \textbf{0.9952} & \textbf{0.11} \\
    RootSIFT & none & \textbf{0.26} & \textbf{0.54} & 1.44 & 0.9949 & \textbf{0.12} \\
    KeyNet+LightGlue & HBB & \textbf{0.27} & \textbf{0.52} & 1.16 & 0.9938 & \textbf{0.14} \\
    BRISK & OBB & \textbf{0.27} & \textbf{0.54} & 1.38 & 0.9944 & \textbf{0.12} \\
    BRISK & HBB & \textbf{0.27} & \textbf{0.54} & 1.38 & 0.9943 & \textbf{0.12} \\
    DISK+LightGlue & OBB & \textbf{0.28} & \textbf{0.58} & 1.38 & 0.9932 & \textbf{0.15} \\
    KeyNet+LightGlue & OBB & \textbf{0.28} & \textbf{0.51} & 1.16 & 0.9938 & \textbf{0.14} \\
    DISK+LightGlue & none & \textbf{0.29} & \textbf{0.56} & 1.42 & 0.9930 & \textbf{0.15} \\
    DISK+LightGlue & HBB & \textbf{0.29} & \textbf{0.56} & 1.39 & 0.9933 & \textbf{0.14} \\
    LoFTR & HBB & \textbf{0.30} & \textbf{0.54} & 1.45 & 0.9918 & \textbf{0.18} \\
    LoFTR & OBB & \textbf{0.30} & \textbf{0.54} & 1.45 & 0.9917 & \textbf{0.18} \\
    LoFTR & none & \textbf{0.31} & \textbf{0.54} & 1.46 & 0.9916 & \textbf{0.18} \\
    BRISK & none & \textbf{0.31} & \textbf{0.56} & 1.44 & 0.9939 & \textbf{0.13} \\
    KeyNet+LightGlue & none & \textbf{0.31} & \textbf{0.55} & 1.31 & 0.9933 & \textbf{0.15} \\
    XFeat & HBB & \textbf{0.37} & \textbf{0.67} & 1.22 & 0.9901 & \textbf{0.21} \\
    XFeat & OBB & \textbf{0.37} & \textbf{0.68} & 1.22 & 0.9901 & \textbf{0.21} \\
    XFeat & none & \textbf{0.38} & \textbf{0.68} & 1.21 & 0.9901 & \textbf{0.21} \\
    DeDoDe+LightGlue & OBB & \textbf{0.38} & \textbf{0.61} & 1.25 & 0.9904 & \textbf{0.21} \\
    DeDoDe+LightGlue & none & \textbf{0.39} & \textbf{0.62} & 1.21 & 0.9900 & \textbf{0.22} \\
    DeDoDe+LightGlue & HBB & \textbf{0.39} & \textbf{0.63} & 1.42 & 0.9902 & \textbf{0.21} \\
    ORB & OBB & \textbf{0.43} & \textbf{0.84} & 1.62 & 0.9875 & \textbf{0.27} \\
    ORB & HBB & \textbf{0.46} & \textbf{0.81} & 1.63 & 0.9875 & \textbf{0.27} \\
    ORB & none & \textbf{0.57} & \textbf{1.15} & 4.75 & 0.9862 & \textbf{0.31} \\
    \bottomrule
  \end{tabular}
  \end{adjustbox}
\end{table}

\begin{table}[htbp]
  \centering
  \caption{Part II (real-time regime at downsample ratio 0.18 and 1000 features) candidates against the homography fitted to each audited frame's ground control points, which involves no reference estimator: CTE is the corner transfer error, IoU and OCE compare the frame's object boxes as placed by the candidate and by the fitted transform. CTE$_{50}$ and CTE$_{90}$ are per case and averaged over the cases; CTE$_{\max}$ is the largest over every audited frame. Every row is scored on 72 audited frames and 1064 ground control points. Bold marks the best value of each column and every value not distinguishable from it: a paired difference to the best, over the 9 cases, whose 95\% cluster-bootstrap interval includes zero or whose estimate is below the equivalence margin (1.00 px, the GCP repeatability, for the pixel columns, none for IoU). With 9 cases the smallest attainable sign-flip $p$ is 0.0039.}
  \label{tab:realtime_masking-gcp}
  \begin{adjustbox}{max width=\textwidth}
  \begin{tabular}{llrrrrr}
    \toprule
    Pipeline & Masking & CTE$^{\mathrm{GCP}}_{50}$ [px] & CTE$^{\mathrm{GCP}}_{90}$ [px] & CTE$^{\mathrm{GCP}}_{\max}$ [px] & IoU$^{\mathrm{GCP}}$ & OCE$^{\mathrm{GCP}}$ [px] \\
    \midrule
    AKAZE & HBB & \textbf{0.51} & \textbf{1.01} & 1.86 & \textbf{0.9874} & \textbf{0.27} \\
    AKAZE & OBB & \textbf{0.51} & \textbf{1.00} & 1.86 & \textbf{0.9874} & \textbf{0.27} \\
    AKAZE & none & \textbf{0.54} & \textbf{1.01} & 1.87 & \textbf{0.9871} & \textbf{0.28} \\
    RootSIFT & HBB & \textbf{0.54} & \textbf{0.92} & 1.90 & \textbf{0.9880} & \textbf{0.25} \\
    RootSIFT & OBB & \textbf{0.54} & \textbf{0.93} & 1.90 & \textbf{0.9880} & \textbf{0.25} \\
    SIFT & HBB & \textbf{0.55} & \textbf{0.93} & 1.75 & \textbf{0.9880} & \textbf{0.25} \\
    RootSIFT & none & \textbf{0.55} & \textbf{0.98} & 1.95 & \textbf{0.9878} & \textbf{0.26} \\
    SIFT & OBB & \textbf{0.56} & \textbf{0.93} & 1.60 & \textbf{0.9880} & \textbf{0.25} \\
    SIFT & none & \textbf{0.57} & \textbf{0.95} & 1.56 & \textbf{0.9875} & \textbf{0.27} \\
    LoFTR & HBB & \textbf{0.62} & \textbf{1.00} & 1.52 & 0.9814 & \textbf{0.40} \\
    LoFTR & OBB & \textbf{0.62} & \textbf{1.00} & 1.52 & 0.9813 & \textbf{0.40} \\
    LoFTR & none & \textbf{0.63} & \textbf{1.01} & 1.51 & 0.9810 & \textbf{0.41} \\
    KeyNet+LightGlue & HBB & \textbf{0.71} & \textbf{1.31} & 1.94 & 0.9806 & \textbf{0.42} \\
    KeyNet+LightGlue & OBB & \textbf{0.71} & \textbf{1.30} & 1.88 & 0.9805 & \textbf{0.41} \\
    KeyNet+LightGlue & none & \textbf{0.76} & \textbf{1.24} & 1.98 & 0.9805 & \textbf{0.42} \\
    XFeat & none & \textbf{0.82} & \textbf{1.52} & 3.30 & 0.9749 & \textbf{0.54} \\
    XFeat & HBB & \textbf{0.83} & \textbf{1.54} & 2.93 & 0.9755 & \textbf{0.53} \\
    XFeat & OBB & \textbf{0.83} & \textbf{1.52} & 3.16 & 0.9753 & \textbf{0.53} \\
    DISK+LightGlue & HBB & \textbf{0.83} & \textbf{1.61} & 2.99 & 0.9800 & \textbf{0.42} \\
    DISK+LightGlue & OBB & \textbf{0.85} & \textbf{1.64} & 2.81 & 0.9800 & \textbf{0.42} \\
    DISK+LightGlue & none & \textbf{0.87} & \textbf{1.61} & 2.80 & 0.9796 & \textbf{0.43} \\
    BRISK & none & \textbf{1.01} & \textbf{1.87} & 2.94 & 0.9769 & \textbf{0.51} \\
    BRISK & HBB & \textbf{1.02} & \textbf{1.94} & 2.77 & 0.9773 & \textbf{0.49} \\
    BRISK & OBB & \textbf{1.04} & \textbf{1.91} & 2.82 & 0.9773 & \textbf{0.49} \\
    DeDoDe+LightGlue & OBB & \textbf{1.06} & \textbf{1.91} & 3.18 & 0.9743 & \textbf{0.55} \\
    DeDoDe+LightGlue & HBB & \textbf{1.06} & \textbf{1.88} & 3.20 & 0.9743 & \textbf{0.55} \\
    DeDoDe+LightGlue & none & \textbf{1.08} & \textbf{1.92} & 3.07 & 0.9742 & \textbf{0.56} \\
    ORB & OBB & \textbf{1.17} & 2.07 & 2.96 & 0.9704 & \textbf{0.66} \\
    ORB & HBB & \textbf{1.19} & 2.03 & 3.02 & 0.9690 & \textbf{0.69} \\
    ORB & none & \textbf{1.67} & 152.57 & 4469.43 & 0.9516 & \textbf{8.59} \\
    \bottomrule
  \end{tabular}
  \end{adjustbox}
\end{table}

\paragraph{Offline}
SIFT, RootSIFT and AKAZE form the tie set on mAA under every masking mode, led by SIFT with
\ac{OBB} masks at \num{0.9816} and ending with unmasked RootSIFT at \num{0.9750}; their median
disagreements with the reference, \num{0.08} to \SI{0.10}{px}, lie far below the audited floor, and
their $\mathrm{mAA}[\eps_{\mathrm{floor}}, 5]$ is \num{0.998} or more. DISK with LightGlue and BRISK
follow at about \num{0.96}, and every pipeline's median stays below \SI{0.6}{px}
(\Cref{tab:offline_masking-dynamic}). The \acp{GCP} separate almost nothing: all thirty
configurations are tied against them in corner and center error (\Cref{tab:offline_masking-gcp}). Three of Part~I's four leaders lead again; BRISK, the
fourth, falls level with DISK with LightGlue.

\paragraph{Real time}
The same three pipelines form the tie set, from unmasked RootSIFT at \num{0.8856} to unmasked AKAZE
at \num{0.8673}, at median registration times of \num{18} to \SI{22}{ms}, inside one frame
interval. LoFTR follows at \num{0.858} to \num{0.859} at \num{49} to \SI{57}{ms}, then KeyNet and
DISK with LightGlue, XFeat, BRISK and DeDoDe with LightGlue between \num{0.811} and \num{0.725};
masked ORB scores \num{0.679}, unmasked ORB \num{0.584}, and the identity \num{0.296}
(\Cref{tab:realtime_masking-dynamic}). Against the \acp{GCP} every configuration is tied
at the median, and at the \num{90}th percentile all but ORB's (\Cref{tab:realtime_masking-gcp}). SIFT and RootSIFT share the reference's detector, which
makes their reference-relative scores the least independent of it, so their standing is read
against the \acp{GCP} as well.

\subsection{Masking}
\label{sec:masking-results}

\paragraph{The primary contrast}
For the average pipeline, \ac{OBB} masking improves every outcome in most cases: in real time
ZNCC and GOS in eight of nine, the three outcomes against the \acp{GCP} in seven;
offline the corner error against them in all nine and the others in five or six. No outcome meets
the rule of \Cref{sec:analysis-plan} in either regime, and the primary contrast is not claimed.
\ac{HBB} masks against none read the same way, and \ac{OBB} against \ac{HBB} masks differ on no
outcome in either regime
(\Cref{tab:offline_masking-contrasts,tab:realtime_masking-contrasts}). Four of the nine cases
never present a moving group, so an average over cases and pipelines dilutes an effect confined to
a few pairs of a few cases and of one pipeline.

\begin{table}[htbp]
  \centering
  \caption{Part II (offline regime at downsample ratio 0.5 and 2000 features): the planned masking contrasts over 9 cases. Each cell is the Hodges--Lehmann estimate of the per-case paired benefit of the first mode over the second (positive favors the first; pixels for CTE and OCE) and the number of cases in which it is positive. CTE, IoU and OCE are against the ground control points on the audited frames, ZNCC and GOS over every pair. A contrast is read over all pipelines (Holm across the outcomes), then per pipeline (Holm across the pipelines). Bold: a benefit with a 95\% cluster-bootstrap interval above zero, an estimate reaching the margin (the GCP repeatability, 1.00 px for CTE and OCE) and a Holm-adjusted $p \leq 0.05$; underlined: the same in favor of the second mode at the unadjusted $p$. The smallest attainable $p$ is 0.0039. The primary contrast (OBB against none, all pipelines) is claimed when an outcome is bold and none is underlined. Intervals and $p$-values: \texttt{offline\_masking\_contrasts.csv}.}
  \label{tab:offline_masking-contrasts}
  \footnotesize
  \setlength{\tabcolsep}{3pt}
  \begin{adjustbox}{max width=\textwidth}
  \begin{tabular}{lrrrrrc}
    \toprule
    Pipeline & CTE$^{\mathrm{GCP}}_{50}$ [px] & ZNCC [$10^{-3}$] & GOS [$10^{-3}$] & IoU$^{\mathrm{GCP}}$ [$10^{-3}$] & OCE$^{\mathrm{GCP}}$ [px] & Claim \\
    \midrule
    \multicolumn{7}{l}{\textit{OBB against none (primary)}} \\
    \textit{All pipelines} & +0.02 (9/9) & +0.02 (5/9) & +0.32 (5/9) & +0.36 (6/9) & +0.01 (6/9) &  \\
    \cmidrule(lr){2-7}
    ORB & +0.11 (9/9) & +0.17 (5/9) & +1.25 (5/9) & +1.50 (7/9) & +0.03 (7/9) & -- \\
    SIFT & +0.01 (6/9) & +0.00 (6/9) & +0.02 (8/9) & +0.07 (3/9) & +0.00 (6/9) & -- \\
    RootSIFT & +0.02 (7/9) & +0.04 (8/9) & \textbf{+0.27} (9/9) & +0.34 (6/9) & +0.01 (6/9) & -- \\
    BRISK & +0.03 (6/9) & \textbf{+0.09} (9/9) & +0.58 (8/9) & +0.47 (8/9) & +0.01 (8/9) & -- \\
    AKAZE & +0.01 (6/9) & +0.01 (6/9) & +0.04 (7/9) & +0.22 (7/9) & +0.01 (6/9) & -- \\
    XFeat & -0.00 (4/9) & -0.01 (3/9) & -0.02 (4/9) & -0.04 (3/9) & +0.00 (3/9) & -- \\
    DISK+LightGlue & +0.01 (5/9) & -0.01 (4/9) & -0.01 (4/9) & +0.03 (5/9) & +0.00 (5/9) & -- \\
    DeDoDe+LightGlue & +0.00 (5/9) & -0.01 (5/9) & +0.04 (5/9) & +0.30 (4/9) & +0.01 (4/9) & -- \\
    KeyNet+LightGlue & +0.02 (6/9) & +0.08 (7/9) & +0.52 (7/9) & +0.55 (5/9) & +0.01 (5/9) & -- \\
    LoFTR & +0.00 (6/9) & +0.00 (5/9) & +0.03 (6/9) & +0.13 (5/9) & +0.00 (4/9) & -- \\
    \midrule
    \multicolumn{7}{l}{\textit{HBB against none (secondary)}} \\
    \textit{All pipelines} & +0.02 (8/9) & +0.03 (5/9) & +0.38 (5/9) & +0.34 (6/9) & +0.01 (6/9) & -- \\
    \cmidrule(lr){2-7}
    ORB & +0.10 (8/9) & +0.11 (4/9) & +1.74 (4/9) & +0.55 (5/9) & +0.02 (6/9) & -- \\
    SIFT & +0.01 (7/9) & +0.00 (6/9) & +0.02 (7/9) & +0.05 (4/9) & +0.00 (6/9) & -- \\
    RootSIFT & +0.02 (6/9) & +0.04 (8/9) & \textbf{+0.27} (9/9) & +0.34 (6/9) & +0.01 (6/9) & -- \\
    BRISK & +0.03 (6/9) & \textbf{+0.09} (9/9) & +0.58 (8/9) & +0.45 (8/9) & +0.01 (7/9) & -- \\
    AKAZE & +0.02 (8/9) & +0.01 (6/9) & +0.05 (7/9) & +0.13 (6/9) & +0.01 (7/9) & -- \\
    XFeat & -0.00 (3/9) & -0.01 (4/9) & -0.03 (4/9) & -0.07 (3/9) & +0.00 (3/9) & -- \\
    DISK+LightGlue & -0.00 (5/9) & +0.01 (5/9) & +0.13 (5/9) & +0.23 (5/9) & +0.00 (4/9) & -- \\
    DeDoDe+LightGlue & +0.00 (5/9) & -0.01 (4/9) & -0.04 (4/9) & +0.32 (5/9) & +0.01 (5/9) & -- \\
    KeyNet+LightGlue & +0.05 (6/9) & +0.07 (7/9) & +0.42 (7/9) & +0.47 (6/9) & +0.01 (7/9) & -- \\
    LoFTR & +0.00 (6/9) & +0.00 (5/9) & +0.04 (6/9) & +0.15 (5/9) & +0.00 (4/9) & -- \\
    \midrule
    \multicolumn{7}{l}{\textit{OBB against HBB (secondary)}} \\
    \textit{All pipelines} & +0.00 (6/9) & +0.01 (5/9) & +0.02 (5/9) & -0.02 (3/9) & -0.00 (3/9) & -- \\
    \cmidrule(lr){2-7}
    ORB & +0.01 (6/9) & +0.03 (5/9) & -0.15 (5/9) & -0.19 (4/9) & -0.00 (4/9) & -- \\
    SIFT & -0.00 (5/9) & -0.00 (4/9) & -0.00 (4/9) & -0.02 (2/9) & -0.00 (2/9) & -- \\
    RootSIFT & +0.00 (7/9) & +0.00 (5/9) & +0.00 (7/9) & +0.01 (6/9) & -0.00 (4/9) & -- \\
    BRISK & -0.00 (4/9) & -0.00 (4/9) & -0.00 (5/9) & +0.02 (6/9) & +0.00 (5/9) & -- \\
    AKAZE & -0.00 (3/9) & -0.00 (2/9) & -0.00 (3/9) & +0.03 (7/9) & +0.00 (7/9) & -- \\
    XFeat & -0.00 (1/9) & +0.00 (6/9) & +0.03 (6/9) & -0.00 (5/9) & +0.00 (5/9) & -- \\
    DISK+LightGlue & +0.00 (5/9) & -0.01 (4/9) & -0.09 (4/9) & -0.11 (3/9) & -0.00 (4/9) & -- \\
    DeDoDe+LightGlue & +0.00 (5/9) & +0.01 (5/9) & +0.04 (5/9) & +0.16 (7/9) & +0.00 (6/9) & -- \\
    KeyNet+LightGlue & -0.00 (5/9) & +0.01 (7/9) & +0.06 (7/9) & +0.04 (5/9) & +0.00 (5/9) & -- \\
    LoFTR & -0.00 (3/9) & +0.00 (6/9) & +0.00 (5/9) & -0.01 (3/9) & -0.00 (3/9) & -- \\
    \bottomrule
  \end{tabular}
  \end{adjustbox}
\end{table}

\begin{table}[htbp]
  \centering
  \caption{Part II (real-time regime at downsample ratio 0.18 and 1000 features): the planned masking contrasts over 9 cases. Each cell is the Hodges--Lehmann estimate of the per-case paired benefit of the first mode over the second (positive favors the first; pixels for CTE and OCE) and the number of cases in which it is positive. CTE, IoU and OCE are against the ground control points on the audited frames, ZNCC and GOS over every pair. A contrast is read over all pipelines (Holm across the outcomes), then per pipeline (Holm across the pipelines). Bold: a benefit with a 95\% cluster-bootstrap interval above zero, an estimate reaching the margin (the GCP repeatability, 1.00 px for CTE and OCE) and a Holm-adjusted $p \leq 0.05$; underlined: the same in favor of the second mode at the unadjusted $p$. The smallest attainable $p$ is 0.0039. The primary contrast (OBB against none, all pipelines) is claimed when an outcome is bold and none is underlined. Intervals and $p$-values: \texttt{realtime\_masking\_contrasts.csv}.}
  \label{tab:realtime_masking-contrasts}
  \footnotesize
  \setlength{\tabcolsep}{3pt}
  \begin{adjustbox}{max width=\textwidth}
  \begin{tabular}{lrrrrrc}
    \toprule
    Pipeline & CTE$^{\mathrm{GCP}}_{50}$ [px] & ZNCC [$10^{-3}$] & GOS [$10^{-3}$] & IoU$^{\mathrm{GCP}}$ [$10^{-3}$] & OCE$^{\mathrm{GCP}}$ [px] & Claim \\
    \midrule
    \multicolumn{7}{l}{\textit{OBB against none (primary)}} \\
    \textit{All pipelines} & +0.02 (7/9) & +0.67 (8/9) & +4.30 (8/9) & +0.82 (7/9) & +0.02 (7/9) &  \\
    \cmidrule(lr){2-7}
    ORB & +0.18 (7/9) & +5.93 (8/9) & +37.90 (8/9) & +5.14 (8/9) & +0.12 (8/9) & -- \\
    SIFT & +0.01 (5/9) & -0.09 (2/9) & -0.42 (2/9) & +0.27 (5/9) & +0.01 (5/9) & -- \\
    RootSIFT & +0.02 (5/9) & -0.13 (2/9) & -0.70 (2/9) & +0.17 (4/9) & +0.01 (4/9) & -- \\
    BRISK & -0.06 (2/9) & +0.75 (8/9) & +4.68 (8/9) & +0.23 (4/9) & +0.01 (4/9) & -- \\
    AKAZE & +0.02 (5/9) & +0.07 (7/9) & +0.51 (7/9) & +0.31 (5/9) & +0.01 (6/9) & -- \\
    XFeat & -0.01 (4/9) & +0.25 (8/9) & +1.37 (8/9) & +0.35 (7/9) & +0.01 (7/9) & -- \\
    DISK+LightGlue & +0.03 (6/9) & -0.04 (4/9) & -0.43 (3/9) & +0.41 (6/9) & +0.01 (6/9) & -- \\
    DeDoDe+LightGlue & +0.02 (5/9) & +0.12 (4/9) & +0.84 (6/9) & +0.11 (5/9) & +0.01 (6/9) & -- \\
    KeyNet+LightGlue & +0.01 (4/9) & -0.02 (6/9) & -0.83 (6/9) & -0.02 (4/9) & +0.01 (5/9) & -- \\
    LoFTR & +0.00 (5/9) & +0.13 (8/9) & +0.74 (7/9) & +0.29 (6/9) & +0.01 (6/9) & -- \\
    \midrule
    \multicolumn{7}{l}{\textit{HBB against none (secondary)}} \\
    \textit{All pipelines} & +0.03 (7/9) & +0.58 (8/9) & +3.91 (8/9) & +0.62 (6/9) & +0.02 (5/9) & -- \\
    \cmidrule(lr){2-7}
    ORB & +0.20 (8/9) & +4.97 (8/9) & +34.44 (8/9) & +3.32 (7/9) & +0.07 (7/9) & -- \\
    SIFT & +0.02 (5/9) & -0.10 (2/9) & -0.42 (2/9) & +0.37 (5/9) & +0.01 (5/9) & -- \\
    RootSIFT & +0.01 (5/9) & -0.15 (2/9) & -0.78 (2/9) & +0.24 (5/9) & +0.01 (4/9) & -- \\
    BRISK & -0.04 (3/9) & +0.68 (7/9) & +3.86 (7/9) & +0.14 (4/9) & -0.00 (4/9) & -- \\
    AKAZE & +0.03 (5/9) & +0.07 (7/9) & +0.57 (7/9) & +0.32 (5/9) & +0.01 (6/9) & -- \\
    XFeat & -0.01 (3/9) & +0.19 (6/9) & +1.31 (7/9) & +0.68 (8/9) & +0.01 (6/9) & -- \\
    DISK+LightGlue & +0.05 (6/9) & -0.06 (5/9) & -0.57 (4/9) & +0.41 (6/9) & +0.01 (7/9) & -- \\
    DeDoDe+LightGlue & +0.01 (6/9) & +0.15 (6/9) & +1.07 (6/9) & -0.09 (4/9) & -0.00 (4/9) & -- \\
    KeyNet+LightGlue & +0.02 (5/9) & -0.16 (5/9) & -1.93 (5/9) & +0.24 (4/9) & +0.00 (5/9) & -- \\
    LoFTR & +0.00 (6/9) & +0.13 (8/9) & \textbf{+0.80} (9/9) & +0.32 (8/9) & +0.01 (8/9) & -- \\
    \midrule
    \multicolumn{7}{l}{\textit{OBB against HBB (secondary)}} \\
    \textit{All pipelines} & -0.00 (4/9) & +0.13 (5/9) & +0.63 (6/9) & +0.17 (6/9) & +0.00 (6/9) & -- \\
    \cmidrule(lr){2-7}
    ORB & +0.02 (6/9) & +0.81 (5/9) & +4.35 (5/9) & +1.32 (6/9) & +0.03 (6/9) & -- \\
    SIFT & -0.01 (2/9) & +0.01 (6/9) & +0.02 (5/9) & -0.04 (4/9) & -0.00 (3/9) & -- \\
    RootSIFT & +0.00 (5/9) & +0.00 (5/9) & +0.01 (5/9) & -0.01 (5/9) & -0.00 (4/9) & -- \\
    BRISK & -0.02 (2/9) & +0.05 (5/9) & +0.07 (5/9) & -0.00 (4/9) & +0.00 (4/9) & -- \\
    AKAZE & +0.00 (6/9) & +0.01 (4/9) & +0.06 (4/9) & +0.03 (6/9) & -0.00 (4/9) & -- \\
    XFeat & -0.00 (4/9) & +0.13 (6/9) & +0.59 (6/9) & -0.29 (3/9) & -0.00 (3/9) & -- \\
    DISK+LightGlue & -0.02 (5/9) & +0.01 (4/9) & +0.39 (5/9) & -0.12 (3/9) & -0.00 (3/9) & -- \\
    DeDoDe+LightGlue & +0.01 (6/9) & -0.00 (4/9) & +0.05 (4/9) & +0.13 (6/9) & +0.00 (6/9) & -- \\
    KeyNet+LightGlue & -0.00 (3/9) & +0.09 (7/9) & +0.62 (7/9) & -0.05 (3/9) & +0.00 (5/9) & -- \\
    LoFTR & -0.01 (1/9) & -0.02 (3/9) & -0.07 (3/9) & -0.03 (5/9) & -0.00 (3/9) & -- \\
    \bottomrule
  \end{tabular}
  \end{adjustbox}
\end{table}

\paragraph{Masking benefit by pipeline}
The gains concentrate in the captured pipeline. In real time, \ac{OBB} masking improves ORB over its unmasked runs in
eight of nine cases on ZNCC, GOS, IoU and OCE against the \acp{GCP}, by
\num{0.0379} in GOS, with bootstrap intervals that exclude zero; after Holm's adjustment over the
ten pipelines none reaches the rule. SIFT and RootSIFT do not gain: masked, they score lower in
ZNCC and GOS in seven of nine cases, by at most \num{0.0007}, and neither change meets the rule. The
per-pipeline benefits that meet the rule are small and belong to pipelines that are never captured
or only marginally: offline, RootSIFT in GOS and BRISK in ZNCC, by \num[group-digits=none]{0.00027} and
\num[group-digits=none]{0.00009} in all nine cases, and in real time LoFTR under \ac{HBB} masks in GOS.

\paragraph{Moving objects or the feature budget}
Masking can help in two ways. It removes the correspondences on the
objects, from which a captured fit is built, and, for a detector that applies the mask before it
caps its keypoints (\Cref{sec:pipelines}), it hands the removed share of the budget to the ground
whether or not the objects moved. The first predicts a benefit concentrated on pairs with a moving
group, a captured share that falls under masking, and an error that loses its alignment with the
motion. The second predicts a benefit for the detectors that refill their budget, largest at small
budgets, that appears in the stationary control and on Part~I's static scenes as well. Both act on
ORB, and the two protocols give evidence of each. The budget effect is real: on Part~I's static scenes
\ac{OBB} masking raises ORB's mAA by \num{0.065} at the real-time point (\Cref{sec:static-results}),
where Part~II's real-time ORB rises from \num{0.5835} to \num{0.6785}. The captured pairs cannot
account for most of that rise. A pair weighs $1/135$ of its case and the cases weigh equally, so the \num{28}
captured pairs can move ORB's mAA by at most $28/(9 \times 135) = 0.023$, and ORB gains in eight of
the nine cases although only five ever hold a moving group (\Cref{tab:realtime_masking-contrasts});
the rest of the rise is of the size of the budget effect on static scenes. A budget effect cannot
make an estimate follow a group, however, and the capture fraction finds that only without masking.
The strata agree: in real time the reference-free benefit of the
average pipeline is largest in the high-moving mixed cell, which holds S2, and near zero in the
low-moving cells and the stationary control (file \suppfile{realtime_masking_benefit_strata}). The
mechanism contrasts point the same way for ORB in real time, with a positive coverage correlation,
a positive persistence gap and a positive alignment contrast, each with an interval above zero and
none meeting the rule after Holm's adjustment. For the average pipeline only the real-time
persistence gap in ZNCC has an interval above zero, and offline no contrast points toward capture
(files \suppfile{offline_masking_mechanism} and \suppfile{realtime_masking_mechanism}). In real time the benefit of
the average pipeline is positive in every offset bin and both persistence halves, larger for pairs
whose foreground moved, and by cell largest in the high-moving cells; offline it is smaller by an
order of magnitude and mixed between \num{16} and \num{127} frames
(files \suppfile{offline_masking_benefit_strata} and \suppfile{realtime_masking_benefit_strata}). With two cases per
cell the design reads them descriptively; the within-case contrasts, which match coverage and weigh
every case equally, are the test.

\paragraph{Box geometry and mask margin}
\ac{OBB} and \ac{HBB} masks do not differ on any outcome in either regime. In the offline sweep
the masks withhold \num{2.3} to \SI{4.0}{\percent} of the frame; every geometry and margin gains a
few $10^{-4}$ in ZNCC and GOS with intervals that include zero, and \acp{OBB} at a larger
margin do not beat horizontal ones at equal suppressed area (file \suppfile{offline_margin_benefit}).
Neither mechanism of \Cref{sec:masking} is large enough here to be detected. The survey is flown square to
an arterial grid, so most vehicles are aligned with the image axes and their \ac{HBB} is
nearly their \ac{OBB}: over the frames of the nine cases the \acp{HBB} cover a median of
\SI{2.17}{\percent} of the frame against \SI{2.03}{\percent} for the oriented ones
(file \suppfile{offline_masking_scenes}), so the road an \ac{HBB} withholds beyond the vehicle, and
the shadow it covers beyond an \ac{OBB}, are both small. At these densities and orientations
the suppressed area, not the box geometry, is what a mask changes.

\subsection{Behavior with temporal distance}
\label{sec:offset-results}

For the average pipeline the disagreement with the reference grows with the offset. In real time
it runs from \SI{0.33}{px} without masking and \SI{0.16}{px} with \ac{OBB} masks at one frame to
\num{3.17} and \SI{1.09}{px} at \num{2048} to \num{3600} frames; offline from \num{0.16} and
\SI{0.09}{px} to \num{0.30} and \SI{0.27}{px}, the order of the reference's own residual
(\Cref{tab:gcp-audit}). Relative to the unmasked median, masking matters most at the shortest
offsets, where the captures of \Cref{fig:capture-by-offset} are densest and the masked median is
about half the unmasked one in both regimes; in real time the absolute gap peaks where unmasked ORB
breaks down, at \num{64} to \num{127} frames (\num{4.53} against \SI{1.09}{px}) and at the longest
offsets (files \suppfile{offline_masking_benefit_strata} and \suppfile{realtime_masking_benefit_strata}).

\section{Discussion}
\label{sec:discussion}

\subsection{Interpretation of the capture measurement}
\label{sec:discussion-capture}
Consensus capture is the failure the dynamic protocol was built to measure. In these nine cases it
occurs without masking: for ORB at both budgets, in real time in every case that presents a
coherently moving group, and by a pixel or two, at the limit of what the \acp{GCP}
resolve, for BRISK, SIFT and RootSIFT at the real-time budget. Under masking it occurs nowhere. This
bears on the pipeline with which \citet{FONOD2025105205} extracted the
Songdo Traffic trajectories~\citep{songdo_traffic}, ORB at the offline budget under \ac{HBB}
masks: unmasked, it is captured on \num{6} pairs, masked on none, which supports masking at
detection for trajectory extraction. Three properties make this a measurement rather than an anecdote. The
instruments resolve it: a complete capture misaligns the fit where the moving group lies by a
sizable fraction of an object width, far above the audited error of the reference and the
repeatability of the \acp{GCP}, and a partial capture still raises the capture
fraction and the motion alignment. The cases were chosen to contain it: selected from the video alone,
at reference points spread over every hover of a recording day, they span the amount and
coherence of moving foreground that pool offers. And the tests were fixed before the outcome: the
primary and mechanism contrasts were specified in advance and are reported whatever their result. Capture
itself is counted descriptively, at a threshold above what the identity transform reaches, with the
counts at stricter thresholds beside it. Of the two descriptors the cases were
selected on, the moving share orders where capture occurs and coherence does not: the case with
most captures lies in a mixed cell, because the survey averages coherence over its offsets while a
capture needs one coherent group at one offset (\Cref{sec:capture-results}).

What the cases cannot say is how the effect scales with the moving share. Their moving foreground
covers \num{0.17}\% to \num{1.94}\% of the frame outside the control (\Cref{tab:survey-cases}), and
unmasked ORB keeps on average \num{949} inliers per pair offline and \num{513} in real time in the
high-moving coherent cell (released files \suppfile{offline_masking_by_scenario} and
\suppfile{realtime_masking_by_scenario}). The estimator weighs a structure by its share of the
correspondences, not of the area (\Cref{sec:why-not-inliers}), so the feature budget and the
detector's preference for the high-contrast content of vehicles (\Cref{fig:static-trial}) matter as
much as the area the vehicles cover, and a low-altitude view over congested lanes, where moving
vehicles can fill a large share of the frame, is a different regime. The protocol carries over
unchanged; the conclusion need not.

\subsection{Joint reading of the two protocols}
Part~I ranks pipelines against exact truth on static scenes and shows which of their settings
matter there: the compute budget, by a wide margin, then the estimator's threshold for estimators
that commit to one, while the matcher thresholds have little effect at the operating point and
masking costs almost nothing. Part~II asks whether that ranking survives footage in which part of
the frame moves. Capture predicts a specific pattern: a pipeline falls without masking, recovers
under masking, and has its error along the motion while unmasked, with a capture fraction near one on the
pairs that carry a moving group. The data show that pattern for ORB at the real-time budget: its
unmasked estimates follow the group with a capture fraction up to \num{0.96} within the survey's
offsets (\Cref{fig:teaser}), and its error along the motion rises; the alignment rises little,
since the captures are a few percent of the pairs. Its fall from \num{0.6785} masked to
\num{0.5835} unmasked is mostly not on the captured pairs, which bound their share of it at \num{0.023},
and the rest is of the size of the budget effect, which masking has on static scenes as well
(\Cref{sec:masking-results}). For the other pipelines the static ranking carries over: SIFT, RootSIFT and AKAZE
lead Part~II in both regimes, masked or not. The static masking cost is the control that makes
that reading possible: in Part~I the annotated objects satisfy the same homography as the
background, so suppressing them cannot improve the consensus, only change which correspondences the
budget keeps.

\subsection{Practical guidance}
No single pipeline is called best. The leading tie set on mAA is AKAZE, RootSIFT, SIFT and BRISK
under the main Part~I configuration and grows to six pipelines at deployment-scale perturbation;
beyond \SI{30}{\degree} of rotation the upright learned models drop out (\Cref{fig:rotation-sweep}).
In Part~II the tie set against the reference is SIFT, RootSIFT and AKAZE under every masking mode
in both regimes; against the \acp{GCP} every configuration ties at the median in both,
and at the \num{90}th percentile all but ORB in real time (\Cref{fig:regime-summary}). Offline,
masking offers little a practitioner could act on: every configuration lies within the repeatability
of the \acp{GCP}, and the masking benefits are tenths of a pixel or less. In real time
it makes ORB usable, at a cost on static scenes of less than \num{0.005} mAA. For a real-time pipeline the choice is open: masked ORB
registers in \num{12} to \SI{13}{ms} at a median error of \num{1.17} to \SI{1.19}{px} against the
\acp{GCP}, and SIFT, RootSIFT and AKAZE, masked or not, in \num{18} to \SI{22}{ms} at
\num{0.51} to \SI{0.57}{px}; all fit one frame interval at the median, and for the latter three masking changes
background alignment by less than \num{0.001} in either direction, which no contrast resolves
(\Cref{tab:realtime_masking-dynamic,tab:realtime_masking-gcp,tab:realtime_masking-contrasts}). Compute excludes some pipelines on
its own: LoFTR cannot run at Part~I's full resolution on the benchmark GPU, and four of the ten
(BRISK, LoFTR, and DeDoDe and KeyNet with LightGlue) need more than one frame interval at the
real-time point. \acp{HBB} and \acp{OBB} do not differ measurably for registration at
these densities and orientations, in the masking contrasts or in the margin sweep; a road grid oblique to
the image, larger vehicles at a lower altitude or long turning queues would give the geometry more
room to matter, and the protocol measures it there unchanged. \acp{OBB} remain useful elsewhere, in the
motion axis of \Cref{sec:motion-axis} and in object-level errors that respect an object's extent.
SIFT and the learned detectors apply the mask after selecting their keypoints in the library
evaluated here (\Cref{sec:pipelines}), so for them masking can only remove keypoints; selecting
after the mask is a change to the library, which the protocol would measure unchanged.

\subsection{Limitations}
\label{sec:limitations}
\paragraph{Coverage of the cases}
The cases come from one recording day at one altitude band, and their moving foreground at the
reference covers at most \num{1.94}\% of the frame, so what is found about capture holds for that
range. Each case is registered onto one reference frame, chosen to span the conditions capture
depends on; how often capture occurs over all the frames a deployed pipeline might register onto is
a different quantity, which the design does not estimate. Part~II has nine independent units at seven intersections: a
masking effect is detectable only if it has the same sign in all of them, or in eight with the
dissenting case the smallest, and the $2 \times 2$ with two cases per cell is descriptive. Taking
the intersection as the unit leaves every claim unchanged, but the per-pipeline benefits that meet
the rule fall short of it, since seven units cannot pass Holm's adjustment over ten pipelines, and
ORB's real-time coverage correlation loses its interval above zero. Cases
with a larger moving share, from lower altitudes or congested approaches, and
more sites and sessions are the most informative extension; the selection procedure is automated
so that they can be added without revisiting the design.

\paragraph{Resolution of the ground control points}
The \acp{GCP} are repeatable to \SI{1.00}{px} at the \num{95}th percentile. They
separate a captured estimate from a correct one and give a masking contrast independent of the
reference, but they cannot rank pipelines whose errors differ by less than a pixel, and the
reference-relative columns, which can, are limited by the audited residual. With eight audited
frames per case, the \acp{GCP} see only the few captured estimates that fall on one
(\Cref{sec:capture-results}); the capture fraction, read against the reference on every pair, is
what counts them. The reference estimator
is a member of the family under test, run with \ac{OBB} masking; its absolute error is measured
by the audit, but not its covariance with a candidate's, which is why the masking comparison is read
from the \acp{GCP} and the reference-free measures.

\paragraph{Budget, resolution and scene}
Part~II's two regimes change the resolution and the feature budget together; Part~I separates the
two on static scenes (\Cref{fig:budget-compute}), and the mask-margin sweep was run in the offline
regime only. Part~I's \num{29} scenes come from one session on one day, so the split prevents
tuning on the evaluation scenes but does not make the session representative of other cities,
altitudes or lighting. The registration
model is a single homography: buildings, raised structures and large tree canopy are excluded from
the reference-free measures, but poles, signs and isolated street trees are not, and their parallax
grows with the platform's drift, which within a hover is small and is the same for every candidate.
Footage from a translating platform, or with tall structure filling much of the frame, needs a
different registration model.

\paragraph{Annotations and object classes}
The \acp{OBB} come from a segmentation-based conversion (\Cref{sec:appendix-obb}); its
fallback share is zero on this benchmark's image sets, and where it is not, a comparison between the two box geometries is diluted by it. Most
of the benchmark is indifferent to what the objects are, but two parts assume vehicles: the motion
axis is recovered from the long axes of \acp{OBB}, which holds for vehicles and vessels and not
for people, and the coherence descriptor assumes a population that moves as streams. Both degrade
to an undefined axis and an unfilled cell rather than to a wrong number.

\section{Conclusion}
\label{sec:conclusion}

Coherently moving objects can capture the geometric consensus of a robust estimator, and the
statistics that describe a fit's self-consistency would then describe the wrong fit. This paper
turns that possibility into a measurement for the complete registration pipeline used in
reference-frame stabilization of hovering-drone video. A static protocol under exact ground truth
establishes what each pipeline can do. A dynamic protocol on nine intersection cases, each
registered onto a reference frame selected from the video alone, measures what the same pipelines
do when part of the frame moves, at an offline and a real-time budget, with a ground-control-point
audit that measures the error of the pseudo-ground-truth reference and scores every candidate
independently of it.

On static scenes, four tuned classical pipelines (AKAZE, RootSIFT, SIFT, BRISK) lead; the compute
budget decides accuracy far more than the matcher thresholds, which MAGSAC++ makes nearly
irrelevant, and masking costs every pipeline less than \num{0.005} mAA. On the dynamic cases
unmasked ORB is captured in real time in every case that presents a coherently moving group, on
\num{28} of \num{304} real-time pairs and \num{6} offline, and the estimates pulled toward a group
have the inlier statistics of the same pipeline at the nearest offsets. Object masking
removes capture for every pipeline at both budgets and restores ORB in real time, though the captured
pairs can account for at most a quarter of that gain; it leaves the leading pipelines (SIFT, RootSIFT, AKAZE)
unchanged, and \acp{HBB} and \acp{OBB} do not differ at these densities and orientations,
where the two withhold nearly the same pixels.

How capture scales with the moving share, from lower altitudes or over congested approaches, is the
question the benchmark is built to answer next. The protocol needs only
annotated boxes for the objects and a dozen or so \acp{GCP} on each of a
handful of frames per case. The benchmark implementation, the data every reported number is
computed from, the \ac{OBB} dataset and the human-verified reference that validates its conversion
are released so that the comparison can be extended to other pipelines, other platforms and other
classes of moving object without changing the protocol.

\section*{CRediT authorship contribution statement}
\textbf{Robert Fonod}: Writing -- review \& editing, Writing -- original draft, Visualization,
Validation, Software, Project administration, Methodology, Investigation, Funding acquisition,
Formal analysis, Data curation, Conceptualization.
\textbf{Nikolas Geroliminis}: Writing -- review \& editing, Supervision, Resources, Project
administration, Funding acquisition.

\section*{Declaration of competing interest}
The authors declare that they have no known competing financial interests or personal relationships
that could have appeared to influence the work reported in this paper.

\section*{Acknowledgments}
This work was supported by the Open Research Data (ORD) Program of the ETH Board and as a part of
NCCR Automation, a National Centre of Competence in Research, funded by the Swiss National Science
Foundation (grant number 51NF40\_225155). The funders had no role in the study design, conduct,
writing, or decision to submit. The authors thank Yanruiqi Yang (EPFL) for revising the boxes and
drawing the oriented ones of the human-verified reference (\Cref{sec:appendix-obb-gt}).

\section*{Data and code availability}

The benchmark implementation is released as \texttt{stabilo-benchmark}~\citep{stabilo_benchmark}
under the MIT license. It builds on \texttt{stabilo}~\citep{stabilo} for registration and on
\texttt{hbb2obb}~\citep{hbb2obb} for object detection and \ac{OBB} conversion, with the vehicle
detector of \texttt{geo-trax}~\citep{geotrax}, whose weights are used under a CC~BY~4.0 license.

The data from which every benchmark number in this paper is computed are released under a
CC~BY~4.0 license~\citep{stabilo_benchmark_data}: Part~I's \num{29} annotated scenes; Part~II's
evaluation and tuning cases with their annotations, \acp{GCP} and cached reference
homographies; the survey output the cases were selected from; and the result directory of every
campaign, which records its configuration, package versions and hardware
(\Cref{sec:appendix-parameters}). The CSV files named in the text and table captions are in the
record's \texttt{supplementary/} directory. Songdo Vision OBB and the human-verified reference
that validates the conversion (\Cref{sec:appendix-obb}) are a separate record under the same
license~\citep{songdo_vision_obb}.

\section*{Declaration of generative AI and AI-assisted technologies in the manuscript preparation process}
During the preparation of this work the authors used Claude (Anthropic) in order to assist with
drafting and editing the text and with checking and editing the references. After using
this tool, the authors reviewed and edited the content as needed and take full responsibility for
the content of the published article.

\newrefcontext[sorting=nyt]
\printbibliography

\appendix
\counterwithin*{table}{section}
\counterwithin*{figure}{section}
\crefalias{section}{appendix}
\crefalias{subsection}{appendix}
\crefalias{subsubsection}{appendix}

\section{Benchmark protocol parameters}
\label{sec:appendix-parameters}
Every campaign is a configuration file in the released implementation, and every result directory
records the configuration, its hash, the resolved run grid, the package versions and the
hardware it ran on. The implementation was written with the assistance of Claude Code (Anthropic)
and is covered by the test suite released with it. The values below are the ones the manuscript's campaigns use.

\begin{table}[htbp]
  \centering
  \caption{Part~I perturbation. A scalar range $v$ means uniform on $[-v, +v]$ for additive
           quantities and on $[1-v, 1+v]$ for multiplicative ones. The rotation sweep fixes the
           rotation at one magnitude per run, narrows translation to $\pm\SI{5}{\percent}$ and
           scale to $\pm\SI{2}{\percent}$ without a projective term, and disables the photometric
           chain; the deployment-regime configuration narrows the geometric ranges as stated in
           \Cref{sec:static-protocol}. The geometric ranges are those of the published campaign.}
  \label{tab:perturbation}
  \small
  \begin{tabular}{@{}lll@{}}
    \toprule
    Perturbation & Range & What it stands for \\
    \midrule
    Translation & $\pm\SI{10}{\percent}$ of $W$ and of $H$ & wind drift \\
    Rotation & $\pm\SI{15}{\degree}$ about the frame center & yaw \\
    Scale & $\pm\SI{5}{\percent}$ & altitude change \\
    Projective & $\pm\num{5e-5}$ on both projective entries & pitch and roll \\
    \midrule
    Brightness & factor in $[0.75, 1.25]$ & exposure drift, time of day, cloud shadow \\
    Saturation & factor in $[0.95, 1.05]$, in HSV & color rendition \\
    Blur & Gaussian, kernel of \num{1}, \num{3} or \num{5}\,px & defocus, residual motion blur \\
    Haze & white veil blended in, density up to \num{0.10} & haze, thin cloud, a long air column \\
    Sensor noise & additive Gaussian, $\sigma$ up to \num{3.0} gray levels & sensor noise \\
    JPEG & quality in $[70, 95]$ & recompression of the video stream \\
    \bottomrule
  \end{tabular}
\end{table}

\begin{table}[htbp]
  \centering
  \caption{Fixed pipeline parameters shared by every preset unless a campaign sweeps them, and
           the reference estimator's settings. The candidates' inlier and ratio-test thresholds
           are the tuned parameters of \Cref{tab:tuned-parameters} in Part~I and are tuned again on
           the tuning cases in Part~II (\Cref{sec:dynamic-protocol}); the reference estimator's are
           tuned the same way at its own settings (\Cref{sec:oracle}). The downsampling ratio and
           the feature budget are those of the matched operating point of
           \Cref{sec:comparability} in every Part~I ranking campaign, and points of the same budget
           grid in Part~II's two regimes (\Cref{sec:dynamic-protocol}). The mask margin $m$ scales every
           box about its center by $1 + m$ before it is rasterized.}
  \label{tab:fixed-parameters}
  \small
  \begin{tabular}{@{}lll@{}}
    \toprule
    Parameter & Candidates & Reference estimator \\
    \midrule
    Matcher & brute force, ratio test (LightGlue for its detectors; none for LoFTR) & brute force, ratio test \\
    Ratio-test threshold & tuned (\Cref{tab:tuned-parameters}) & tuned (\Cref{sec:oracle}) \\
    Inlier threshold & tuned (\Cref{tab:tuned-parameters}) & tuned (\Cref{sec:oracle}) \\
    Robust estimator & MAGSAC++ (OpenCV flag 38), projective model & same \\
    Iteration cap & \num{5000} & \num{20000} \\
    Confidence & $1 - 10^{-6}$ & $1 - 10^{-7}$ \\
    Feature budget & \num{8000}; Part~II: \num{2000} offline, \num{1000} in real time & \num{20000} \\
    Reference-frame budget & twice the feature budget & twice the feature budget \\
    Downsampling ratio & \num{1.0}; Part~II: \num{0.50} offline, \num{0.18} in real time & \num{1.0} \\
    Mask margin $m$ & \num{0.15} & \num{0.20} \\
    Masking & none, horizontal, oriented (the axis) & oriented \\
    \bottomrule
  \end{tabular}
\end{table}

\begin{table}[htbp]
  \centering
  \caption{Part~II protocol constants.}
  \label{tab:dynamic-parameters}
  \small
  \begin{tabular}{@{}ll@{}}
    \toprule
    Reference frame & a frame of the hover chosen by the case selection \\
    Candidate references & every \num{900} frames, clear of the hover's ends, with \num{3600} frames after them \\
    Selection offsets & \num{1}, \num{2}, \num{5}, \num{11}, \num{22} frames \\
    Displaced & farther than \num{0.1} median object width \\
    Moving together & within \num{0.1} median object width or \num{25}\% of the displacement, whichever is larger \\
    Regimes & offline: ratio \num{0.50}, \num{2000} features; real-time: ratio \num{0.18}, \num{1000} features \\
    Inlier threshold & in pixels of the downsampled frame, both regimes \\
    Sampling & offsets up to \num{3600} frames in twelve octave bins, up to \num{15} pairs each, \num{135} pairs \\
    Audit bins & \num{1}--\num{3}, \num{4}--\num{31}, \num{32}--\num{511}, \num{512}--\num{3600} frames, two frames each \\
    Capture & capture fraction $\lambda \geq 0.1$, on pairs whose moving group holds $\geq 4$ objects \\
    Object dilation $\tau$ & \num{10}\,px in both frames \\
    Region erosion $\rho$ & \num{8}\,px; the validity term drops a further \num{2}\,px fringe \\
    Excluded static structure & buildings, raised structures, large tree canopy; polygons on the reference frame \\
    Minimum region area & \SI{5}{\percent} of the frame, else the identity's score is recorded \\
    SSIM & $7 \times 7$ window, region eroded by its radius \\
    MI & $64 \times 64$ intensity histogram, in bits \\
    Motion-axis concentration $\kappa_{\min}$ & \num{0.3} \\
    HEA thresholds & \num{1}, \num{3}, \num{5}, \num{10}\,px; mAA over $[0, 5]$\,px \\
    GCPs & $\geq \num{12}$ per audited frame, $\geq \num{8}$ audited frames per case \\
    Repeats & \num{1} for OpenCV estimators; \num{3} for the learned pipelines on one case \\
    \bottomrule
  \end{tabular}
\end{table}

Two conventions are worth stating, because a reader comparing against published numbers for the
same detectors will otherwise be comparing different quantities. First, the inlier threshold is
expressed in pixels of the full-resolution \emph{reference} frame whatever the processed
resolution, which is the convention of the stabilization library: the pipelines estimate current
$\to$ reference, so the reprojection residual is evaluated there, and a threshold quoted in the
other frame differs by the scale factor of the transform. Part~I's ranking campaigns and
Part~II's reference estimator run at full resolution, where the processed and the full-resolution
frame coincide, and Part~I's budget sweep keeps the library's convention. Part~II's two regimes
state the threshold in pixels of the downsampled frame instead, so that Part~I's values, from which
their tuning starts, keep their size relative to keypoint localization; \SI{4}{px} is \SI{8}{px} at
full resolution in the offline regime and about \SI{22}{px} in the real-time one. Because every threshold is tuned at the operating
point it is run at, no reported number depends on a rescaling between resolutions. Second, all reported
values are for the implementations and versions recorded in each result manifest; as
\citet{barath2023homography} document, the same nominal algorithm can differ measurably between
libraries, so the implementation is part of what is being compared.

\paragraph{Repeated runs}
\citet{jin2021image} run each evaluation three times and average the results to reduce the
randomness in them, which they found negligible. We report single runs instead, having measured
how random the estimators used here are. On the OpenCV build recorded in the manifests, homography
estimation does not change its output across different global random seeds, at outlier
fractions up to \SI{75}{\percent}, and does not advance the global random stream either; each
estimator seeds an internal generator from a fixed constant.
Every method is additionally bit-identical across repeated calls on identical input, with one
exception: the parallel LO-RANSAC variant returns one of a small set of matrices depending on
thread scheduling, a race that no seed removes. Its spread is at most
$1.5 \times 10^{-3}$ px of corner transfer on a $3840 \times 2160$ frame, more than two orders
of magnitude below the smallest audited resolution floor of \Cref{tab:gcp-audit} and therefore
not a limitation on any claim made here. For the OpenCV estimators, results are consequently
reproducible exactly rather than in distribution. The learned detectors and matchers run under
a tensor library whose kernel selection on a GPU is not guaranteed to be bit-exact, so that
argument does not extend to them, and the learned pipelines of Part~II's offline regime were
repeated three times on one case instead.
The repeats agree: every learned pipeline under every masking mode returned the same estimate
on all \num{135} pairs in all three repeats, and no pair changed between success and failure
(released file \suppfile{offline_repeats_repeat_spread}).

\FloatBarrier
\section{Part I: further results}
\label{sec:appendix-static}
\begin{figure}[htbp]\centering
  \includegraphics[width=\textwidth]{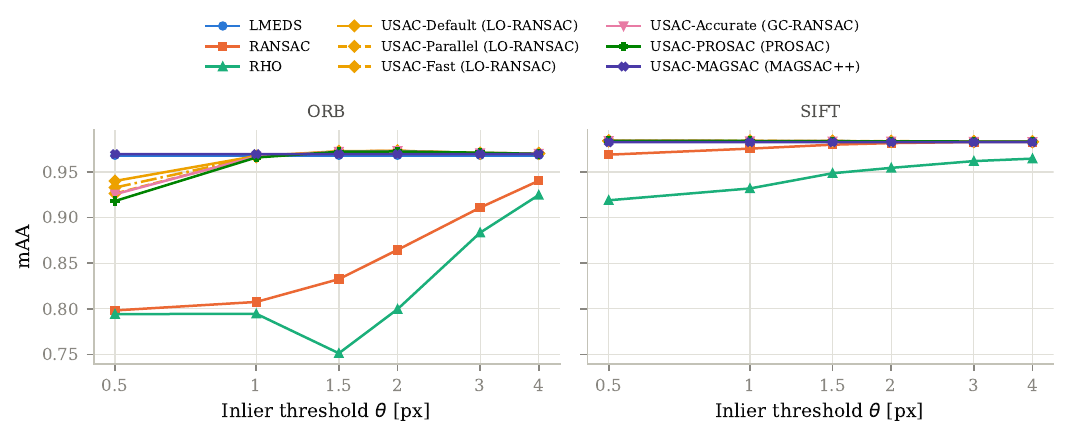}
  \caption{mAA against the inlier threshold for every robust
estimator, with the detector fixed (ORB, left; SIFT, right: one binary and one floating-point
descriptor), \ac{OBB} masking, on the evaluation scenes. Color follows the algorithm and the line
style the OpenCV flag within it, so the three LO-RANSAC configurations share a hue.}\label{fig:estimator-thresholds}
\end{figure}

\begin{figure}[htbp]\centering
  \includegraphics[width=0.6\textwidth]{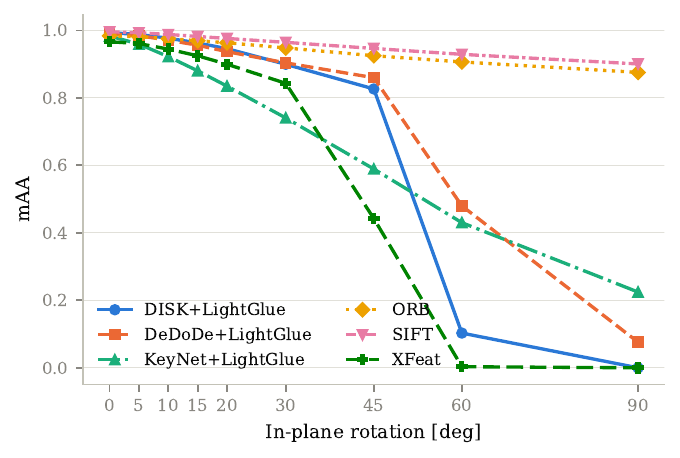}
  \caption{mAA against in-plane rotation on the evaluation
scenes, \ac{OBB} masking, \num{50} trials per scene at each of nine magnitudes. LoFTR cannot run
at full resolution on the benchmark GPU and is not shown.}\label{fig:rotation-sweep}
\end{figure}

\Cref{tab:tuned-parameters} lists the values every pipeline keeps under the rule of
\Cref{sec:comparability} on the \num{9} tuning scenes at \num{30} trials each, over ten ratios $r$
(\num{0.5}, \num{0.6}, \num{0.7}, \num{0.75}, \num{0.8}, \num{0.85}, \num{0.9}, \num{0.93},
\num{0.96}, \num{0.99}) and eight inlier thresholds $\theta$ (\num{0.25}, \num{0.5}, \num{1},
\num{1.5}, \num{2}, \num{3}, \num{4}, \SI{8}{px}), with the mAA at the untuned starting values and
the number of grid points tied with the best. Where every grid point is tied, no point of the grid
is better supported than another, which is why the preset's values are kept. LoFTR has no row,
since it cannot run at the operating point. The other tables behind \Cref{sec:static-results}
are released as files of the data record: the detector axis at the starting values on the
evaluation scenes (\suppfile{detectors_default_static}) and at deployment-scale perturbation
(\suppfile{detectors_hover_static}), the rotation sweep of \Cref{fig:rotation-sweep}
(\suppfile{rotation_sweep}) and the estimator sweep of \Cref{fig:estimator-thresholds}
(\suppfile{estimators_estimators}).

\begin{table}[htbp]
  \centering
  \caption{Part I: the joint sweep of the ratio-test threshold $r$ and the inlier threshold $\theta$ (reference-frame pixels) on the held-out tuning scenes. A pipeline keeps its preset's values unless a grid point beats them on mAA by more than the margin (0.01); Grid counts the grid points and Tied those within the margin of the best. mAA$_{\mathrm{default}}$ is the score at the untuned starting values, a ratio of 0.9 and an inlier threshold of 2 px (3 px for RootSIFT and SIFT; 2.5 px for AKAZE and BRISK), and $\Delta$mAA, never bolded, the gain from tuning. $r$ reads -- for matchers without a ratio test. No configuration failed on any trial. $^{\ddagger}$ marks a reference value off the grid, read at its nearest grid point (the better of two equidistant ones), so the gain beside it is a lower bound. $^{\S}$ marks incumbent values kept because no grid point beats them by more than the margin.}
  \label{tab:tuned-parameters}
  \small
  \setlength{\tabcolsep}{4pt}
  \begin{adjustbox}{max width=\textwidth}
  \begin{tabular}{lrrrrrrrr}
    \toprule
    Pipeline & $r$ & $\theta$ [px] & mAA & mAA$_{\mathrm{default}}$ & $\Delta$mAA & CTE$_{50}$ [px] & Grid & Tied \\
    \midrule
    AKAZE & 0.70 & 4 & \textbf{0.9856}$^{\S}$ & 0.9844$^{\ddagger}$ & +0.0011 & \textbf{0.066} & 80 & 80 \\
    SIFT & 0.70 & 0.5 & 0.9840$^{\S}$ & 0.9830 & +0.0010 & 0.076 & 80 & 80 \\
    RootSIFT & 0.70 & 0.5 & 0.9839$^{\S}$ & 0.9829 & +0.0010 & 0.077 & 80 & 80 \\
    BRISK & 0.70 & 1.5 & 0.9811$^{\S}$ & 0.9802$^{\ddagger}$ & +0.0009 & 0.089 & 80 & 80 \\
    DISK+LightGlue & -- & 2 & 0.9708$^{\S}$ & 0.9708 & +0.0000 & 0.126 & 8 & 8 \\
    ORB & 0.85 & 4 & 0.9708$^{\S}$ & 0.9706 & +0.0002 & 0.137 & 80 & 80 \\
    DeDoDe+LightGlue & -- & 1 & 0.9625$^{\S}$ & 0.9624 & +0.0001 & 0.177 & 8 & 8 \\
    XFeat & 0.70 & 2 & 0.9470$^{\S}$ & 0.9407 & +0.0062 & 0.254 & 80 & 57 \\
    KeyNet+LightGlue & -- & 4 & 0.9214$^{\S}$ & 0.9214 & +0.0000 & 0.382 & 8 & 8 \\
    \bottomrule
  \end{tabular}
  \end{adjustbox}
\end{table}

\paragraph{The estimator sweep}
The estimator axis is swept for two detectors rather than ten. The estimator sees only the
correspondences, and the two cover both descriptor families the classical pipelines use: ORB, the
fastest, with a binary descriptor, and SIFT, the reference estimator's detector, with a
floating-point one. Nine estimators at six thresholds already make \num{108} runs of \num{2000}
trials each; all ten detectors would make five times as many, the learned ones among the slowest.
With the detector fixed and \ac{OBB} masking, the estimator matters less than the threshold it runs
at (\Cref{fig:estimator-thresholds}; values in file \suppfile{estimators_estimators}). With SIFT,
every estimator but RHO reaches an mAA of \num{0.983} to \num{0.984} at its best threshold, all
within the equivalence margin of one another; RHO peaks at \num{0.965} and is the only estimator
that fails (\num{6} of its \num{12000} trials) or leaves the integration interval (\num{9}). With
ORB, the three LO-RANSAC configurations, GC-RANSAC and PROSAC reach \num{0.972} to \num{0.973} at
$\theta = \SI{2}{px}$, MAGSAC++ and LMEDS lie within the margin at \num{0.970} and \num{0.968}, and
plain RANSAC and RHO fall outside it at \num{0.940} and \num{0.925}. MAGSAC++ and LMEDS move by less
than \num{0.0001} over the whole range of $\theta$ for both detectors, since neither commits to one
threshold (LMEDS has none), whereas with ORB plain RANSAC loses \num{0.14} and RHO \num{0.13}
between their best threshold and $\theta = \SI{0.5}{px}$. The inlier ratio records the threshold
rather than the estimate: at a given $\theta$ it is nearly the same for every estimator but RHO,
it rises from about \num{0.17} to \num{0.81} for ORB and from \num{0.77} to \num{0.93} for SIFT
between $\theta = \SI{0.5}{px}$ and \SI{4}{px} whether mAA rises, falls or stays constant, and the
highest inlier ratio in the sweep belongs to neither detector's best configuration.

\paragraph{The deployment regime}
At deployment-scale perturbation (file \suppfile{detectors_hover_static}) every pipeline scores higher
and the pipelines move closer together: unmasked mAA spans \num{0.968} to \num{0.994}, against \num{0.919} to \num{0.984}
in the main configuration. The leading tie set grows from four pipelines to six. AKAZE stays at the top at
\num{0.994}, SIFT and RootSIFT follow at \num{0.993}, and DISK with LightGlue, DeDoDe with
LightGlue and BRISK lie within the equivalence margin of it, while ORB, KeyNet with LightGlue and
XFeat remain outside it. No pipeline moves by more than two places, and no trial fails or leaves
the integration interval under any masking strategy. The wider configuration therefore separates
the pipelines more than a hovering platform would, without changing which three pipelines lead,
which is why the ranking is read from it.

\paragraph{Rotation limits of upright models}
Several learning-based detectors do not estimate keypoint orientation, and
\Cref{fig:rotation-sweep} (values in file \suppfile{rotation_sweep}) shows where that starts to
matter. Up to \SI{30}{\degree} these pipelines lose accuracy gradually, as the classical ones do:
at \SI{30}{\degree} XFeat, DISK and DeDoDe with LightGlue score \num{0.843}, \num{0.899} and
\num{0.904}, against \num{0.964} for SIFT. Beyond it they collapse. XFeat falls to \num{0.442} at
\SI{45}{\degree} and \num{0.003} at \SI{60}{\degree}; DISK and DeDoDe with LightGlue hold
\num{0.826} and \num{0.860} at \SI{45}{\degree}, fall to \num{0.103} and \num{0.480} at
\SI{60}{\degree}, and reach \num{0.000} and \num{0.076} at \SI{90}{\degree}. KeyNet with
LightGlue, which estimates orientation, degrades steadily instead, to \num{0.224} at
\SI{90}{\degree}, and SIFT and ORB stay above \num{0.87} over the whole range. The
\SI{15}{\degree} of the main configuration lies well inside the range where no pipeline has
collapsed. The limit is a property of the pretrained weights, so it is reported rather than
compensated for.

\FloatBarrier
\section{Part II: further results}
\label{sec:appendix-dynamic}
This appendix gives the tables and figures that the main tables of \Cref{sec:dynamic} summarize. Every one is
written by \texttt{report} from the result directories, and the file each caption names holds the
rows a table summarizes; the remaining tables are released as files of the data record, written
by the same run.

\paragraph{Tuning}
\Cref{tab:offline_tuning-tuned-parameters,tab:realtime_tuning-tuned-parameters} list the thresholds
selected in each regime on the five tuning cases, with Part~I's values as the incumbents.

\begin{table}[htbp]
  \centering
  \caption{Part II (offline regime at downsample ratio 0.5 and 2000 features): the joint sweep of $r$ and $\theta$ (pixels of the downsampled image) on 5 tuning cases from the sites of Part I's tuning scenes, unmasked, against the reference estimator, each case weighing equally. A pipeline keeps its Part I values unless a grid point beats them on mAA by more than the margin (0.01); Grid counts the grid points and Tied those within the margin of the best. mAA$_{\mathrm{default}}$ is the score at the Part I values and $\Delta$mAA, never bolded, the gain from tuning. $r$ reads -- for matchers without a ratio test. No configuration failed on any trial. $^{\dagger}$ marks a selected value at an end of its swept range, where the optimum may lie outside the grid. $^{\S}$ marks incumbent values kept because no grid point beats them by more than the margin.}
  \label{tab:offline_tuning-tuned-parameters}
  \small
  \setlength{\tabcolsep}{4pt}
  \begin{adjustbox}{max width=\textwidth}
  \begin{tabular}{lrrrrrrrr}
    \toprule
    Pipeline & $r$ & $\theta$ [px] & mAA & mAA$_{\mathrm{default}}$ & $\Delta$mAA & CTE$_{50}$ [px] & Grid & Tied \\
    \midrule
    SIFT & 0.40$^{\dagger}$ & 0.1$^{\dagger}$ & \textbf{0.9812} & 0.9710 & +0.0102 & 0.084 & 182 & 85 \\
    AKAZE & 0.40$^{\dagger}$ & 0.1$^{\dagger}$ & 0.9812 & 0.9681 & +0.0131 & \textbf{0.081} & 182 & 65 \\
    RootSIFT & 0.70 & 0.5 & 0.9716$^{\S}$ & 0.9716 & +0.0000 & 0.131 & 182 & 98 \\
    LoFTR & -- & 2 & 0.9597$^{\S}$ & 0.9597 & +0.0000 & 0.195 & 13 & 13 \\
    BRISK & 0.40$^{\dagger}$ & 0.1$^{\dagger}$ & 0.9571 & 0.9415 & +0.0156 & 0.197 & 182 & 51 \\
    DISK+LightGlue & -- & 2 & 0.9559$^{\S}$ & 0.9559 & +0.0000 & 0.191 & 13 & 8 \\
    KeyNet+LightGlue & -- & 4 & 0.9522$^{\S}$ & 0.9522 & +0.0000 & 0.230 & 13 & 12 \\
    DeDoDe+LightGlue & -- & 1 & 0.9380$^{\S}$ & 0.9380 & +0.0000 & 0.274 & 13 & 13 \\
    XFeat & 0.45 & 8 & 0.9336 & 0.9197 & +0.0138 & 0.313 & 182 & 59 \\
    ORB & 0.85 & 4 & 0.9004$^{\S}$ & 0.9004 & +0.0000 & 0.435 & 182 & 18 \\
    \bottomrule
  \end{tabular}
  \end{adjustbox}
\end{table}

\begin{table}[htbp]
  \centering
  \caption{Part II (real-time regime at downsample ratio 0.18 and 1000 features): the joint sweep of $r$ and $\theta$ (pixels of the downsampled image) on 5 tuning cases from the sites of Part I's tuning scenes, unmasked, against the reference estimator, each case weighing equally. A pipeline keeps its Part I values unless a grid point beats them on mAA by more than the margin (0.01); Grid counts the grid points and Tied those within the margin of the best. mAA$_{\mathrm{default}}$ is the score at the Part I values and $\Delta$mAA, never bolded, the gain from tuning. $r$ reads -- for matchers without a ratio test. No configuration failed on any trial. $^{\dagger}$ marks a selected value at an end of its swept range, where the optimum may lie outside the grid. $^{\S}$ marks incumbent values kept because no grid point beats them by more than the margin.}
  \label{tab:realtime_tuning-tuned-parameters}
  \small
  \setlength{\tabcolsep}{4pt}
  \begin{adjustbox}{max width=\textwidth}
  \begin{tabular}{lrrrrrrrr}
    \toprule
    Pipeline & $r$ & $\theta$ [px] & mAA & mAA$_{\mathrm{default}}$ & $\Delta$mAA & CTE$_{50}$ [px] & Grid & Tied \\
    \midrule
    RootSIFT & 0.70 & 0.5 & \textbf{0.9095}$^{\S}$ & 0.9095 & +0.0000 & \textbf{0.437} & 182 & 150 \\
    SIFT & 0.70 & 0.5 & 0.9088$^{\S}$ & 0.9088 & +0.0000 & 0.454 & 182 & 145 \\
    AKAZE & 0.50 & 0.1$^{\dagger}$ & 0.9064 & 0.8963 & +0.0100 & 0.452 & 182 & 88 \\
    LoFTR & -- & 2 & 0.8639$^{\S}$ & 0.8639 & +0.0000 & 0.703 & 13 & 13 \\
    KeyNet+LightGlue & -- & 4 & 0.8326$^{\S}$ & 0.8326 & +0.0000 & 0.816 & 13 & 4 \\
    XFeat & 0.80 & 6 & 0.8175 & 0.7669 & +0.0506 & 0.920 & 182 & 49 \\
    DeDoDe+LightGlue & -- & 4 & 0.8018 & 0.6769 & +0.1249 & 0.981 & 13 & 6 \\
    DISK+LightGlue & -- & 3 & 0.7958 & 0.7590 & +0.0368 & 1.000 & 13 & 2 \\
    BRISK & 0.70 & 1.5 & 0.7955$^{\S}$ & 0.7955 & +0.0000 & 0.998 & 182 & 68 \\
    ORB & 0.99$^{\dagger}$ & 3 & 0.6707 & 0.6593 & +0.0114 & 1.488 & 182 & 6 \\
    \bottomrule
  \end{tabular}
  \end{adjustbox}
\end{table}

\paragraph{Case selection and the reference estimator}
\Cref{tab:survey_validation} checks the survey's descriptors against the published trajectories,
file \suppfile{offline_masking_scenes} characterizes the frames of the nine cases,
\Cref{tab:oracle_tuning-tuned-parameters} reports the tuning of the reference estimator,
\Cref{tab:gcp-proposals} how every \ac{GCP} of the audited frames was placed, and
\Cref{tab:gcp-proposal-validation} the agreement of the proposals with unaided clicks on two cases.

\begin{table}[htbp]
  \centering
  \caption{The survey's 831 candidate reference frames against the published vehicle trajectories of the same hovers, which the survey never reads. Each descriptor is recomputed from the trajectories with the survey's own definitions (ground coordinates, thresholds in median vehicle widths, the survey's offsets). Areas are in square meters there and frame fractions in the survey, so agreement is stated as correlation. Cell agreement is the share of candidates placed in the same moving $\times$ coherence cell by both, among those the survey placed, by their margin from the split boundaries; the selection takes the two of largest margin per cell.}
  \label{tab:survey_validation}
  \small
  \begin{tabular}{lrrr}
    \toprule
    Descriptor & Candidates & Pearson $r$ & Spearman $\rho$ \\
    \midrule
    Static foreground & 831 & 0.972 & 0.955 \\
    Moving (displaced) foreground & 831 & 0.974 & 0.953 \\
    Coherently moving foreground & 831 & 0.970 & 0.933 \\
    Coherence & 726 & 0.898 & 0.892 \\
    Stationarity & 831 & 0.926 & 0.918 \\
    \midrule
    Cell agreement & Candidates & \multicolumn{2}{r}{Share} \\
    \midrule
    all placed & 703 & \multicolumn{2}{r}{0.81} \\
    top 50\% by margin & 352 & \multicolumn{2}{r}{0.94} \\
    top 25\% by margin & 176 & \multicolumn{2}{r}{0.97} \\
    selected cases & 9 & \multicolumn{2}{r}{9/9} \\
    \bottomrule
  \end{tabular}
\end{table}

\begin{table}[htbp]
  \centering
  \caption{Part I: the joint sweep of the ratio-test threshold $r$ and the inlier threshold $\theta$ (reference-frame pixels) on the held-out tuning scenes. A pipeline keeps its preset's values unless a grid point beats them on mAA by more than the margin (0.01); Grid counts the grid points and Tied those within the margin of the best. mAA$_{\mathrm{default}}$ is the score at the preset's own values (it has no separate starting values), and $\Delta$mAA, never bolded, the gain from tuning. $r$ reads -- for matchers without a ratio test. No configuration failed on any trial. $^{\S}$ marks incumbent values kept because no grid point beats them by more than the margin.}
  \label{tab:oracle_tuning-tuned-parameters}
  \small
  \setlength{\tabcolsep}{4pt}
  \begin{adjustbox}{max width=\textwidth}
  \begin{tabular}{lrrrrrrrr}
    \toprule
    Pipeline & $r$ & $\theta$ [px] & mAA & mAA$_{\mathrm{default}}$ & $\Delta$mAA & CTE$_{50}$ [px] & Grid & Tied \\
    \midrule
    Reference estimator & 0.70 & 0.5 & \textbf{0.9856}$^{\S}$ & 0.9856 & +0.0000 & \textbf{0.073} & 70 & 70 \\
    \bottomrule
  \end{tabular}
  \end{adjustbox}
\end{table}

\begin{table}[htbp]
  \centering
  \caption{Part II: how the ground control points of the audited frames were placed, per case. A local image fit proposes each point and the annotator reviews the standing proposals: Accepted counts those kept as they were, Corrected those moved, by the median Corr.$_{50}$ and the largest Corr.$_{\max}$ distance, and By hand the points without a standing proposal, placed in the annotation panel. The reference frame's points, all placed by hand, are not counted. The last row pools every case.}
  \label{tab:gcp-proposals}
  \small
  \setlength{\tabcolsep}{4pt}
  \begin{adjustbox}{max width=\textwidth}
  \begin{tabular}{llrrrrrr}
    \toprule
    Case & Cell & Points & Accepted & Corrected & Corr.$_{50}$ [px] & Corr.$_{\max}$ [px] & By hand \\
    \midrule
    2022-10-06\_D1\_PM1\_B2 & high moving / mixed & 120 & 117 & 0 & -- & -- & 3 \\
    2022-10-06\_D10\_PM5\_U1 & high moving / coherent & 119 & 85 & 5 & 0.35 & 0.50 & 29 \\
    2022-10-06\_D3\_AM1\_M1 & low moving / coherent & 119 & 113 & 0 & -- & -- & 6 \\
    2022-10-06\_D3\_PM3\_F2 & high moving / coherent & 119 & 116 & 1 & 0.79 & 0.79 & 2 \\
    2022-10-06\_D7\_PM1\_P1 & low moving / coherent & 118 & 116 & 0 & -- & -- & 2 \\
    2022-10-06\_D7\_PM5\_P2 & low moving / mixed & 117 & 116 & 0 & -- & -- & 1 \\
    2022-10-06\_D8\_PM3\_M1 & low moving / mixed & 116 & 110 & 0 & -- & -- & 6 \\
    2022-10-06\_D9\_AM1\_R1 & stationary control & 119 & 112 & 0 & -- & -- & 7 \\
    2022-10-06\_D9\_AM3\_S2 & high moving / mixed & 117 & 111 & 1 & 0.50 & 0.50 & 5 \\
    all cases &  & 1064 & 996 & 7 & 0.35 & 0.79 & 61 \\
    \bottomrule
  \end{tabular}
  \end{adjustbox}
\end{table}

\begin{table}[htbp]
  \centering
  \caption{Click noise against estimator error on 2 cases (2022-10-06\_D10\_PM5\_U1 and 2022-10-06\_D9\_AM3\_S2), whose audited frames were also annotated entirely by hand, each point clicked in a magnified panel opened near the reference estimator's prediction under a hidden offset. Over the points that have an unaided click, a standing proposal and a prediction, the distance between each pair of positions in current-frame pixels: median and 95th percentile; the last row pools every point.}
  \label{tab:gcp-proposal-validation}
  \small
  \begin{adjustbox}{max width=\textwidth}
  \begin{tabular}{lrrrrrrr}
    \toprule
     & & \multicolumn{2}{c}{Click -- proposal} & \multicolumn{2}{c}{Click -- estimator} & \multicolumn{2}{c}{Proposal -- estimator} \\
    \cmidrule(lr){3-4} \cmidrule(lr){5-6} \cmidrule(lr){7-8}
    Case & Points & Median & 95th & Median & 95th & Median & 95th \\
    \midrule
    2022-10-06\_D10\_PM5\_U1 & 90 & 0.39 & 0.75 & 0.37 & 0.76 & 0.04 & 0.09 \\
    2022-10-06\_D9\_AM3\_S2 & 112 & 0.47 & 0.98 & 0.52 & 1.08 & 0.11 & 0.40 \\
    \midrule
    Pooled & 202 & 0.44 & 0.82 & 0.46 & 0.91 & 0.06 & 0.32 \\
    \bottomrule
  \end{tabular}
  \end{adjustbox}
\end{table}

\begin{figure}[htbp]\centering
  \includegraphics[width=\textwidth]{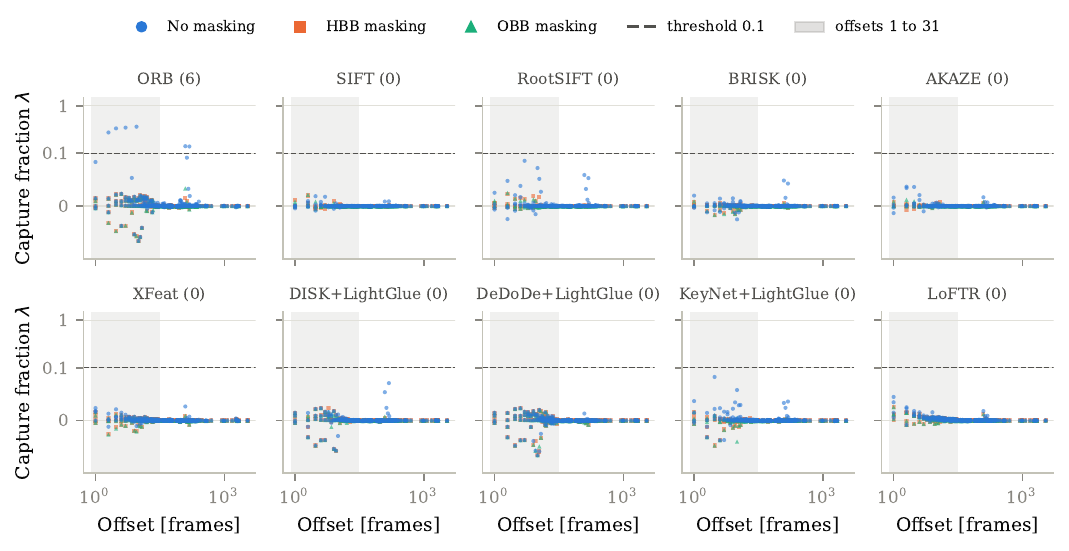}
  \caption{The capture fraction of every offline measurement on a pair with a moving group, drawn as
  \Cref{fig:capture-scatter}.}\label{fig:capture-scatter-offline}
\end{figure}

\begin{figure}[htbp]\centering
  \includegraphics[width=0.7\textwidth]{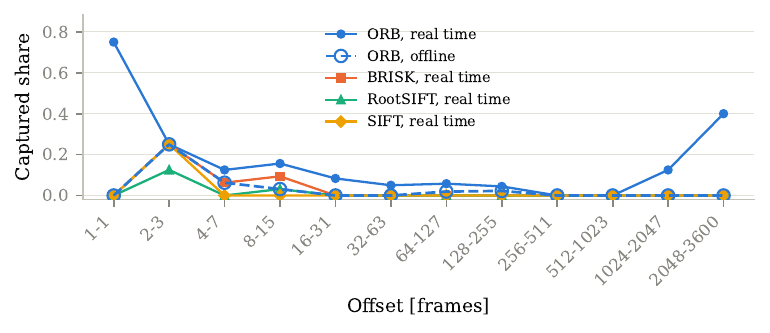}
  \caption{Share of the pairs with a moving group on which the
estimate is captured (capture fraction $\lambda \geq 0.1$, \Cref{eq:capture-fraction}), by offset
bin, for every configuration captured at least once in either regime, all without
masking.}\label{fig:capture-by-offset}
\end{figure}

\begin{figure}[htbp]\centering
  \includegraphics[width=0.49\textwidth]{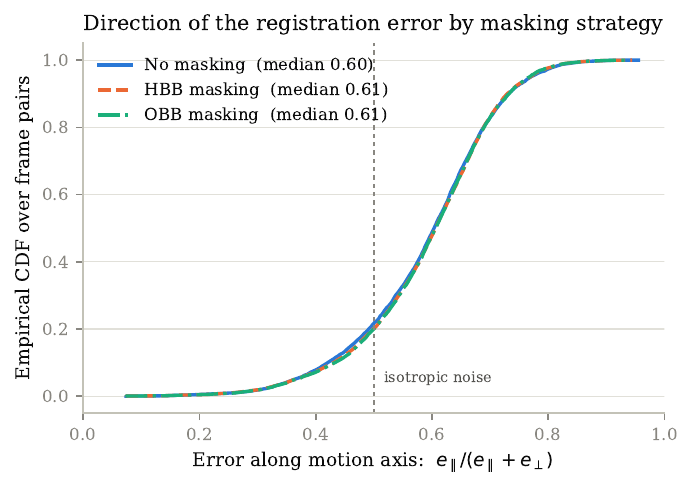}\hfill
  \includegraphics[width=0.49\textwidth]{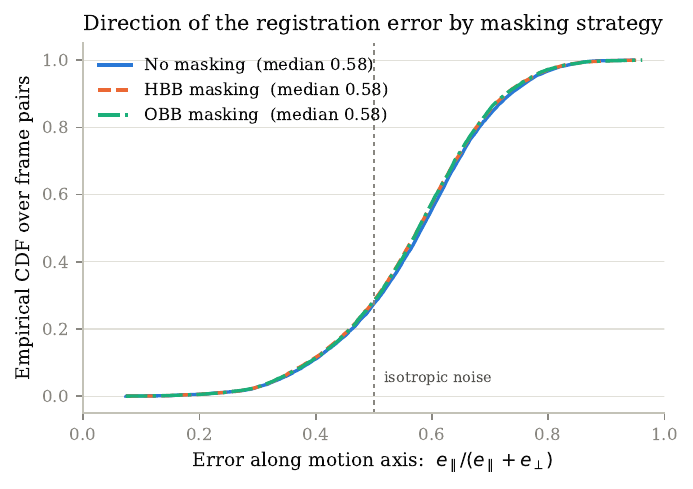}
  \caption{Direction of the registration error: the share of the corner error lying along the
  dominant motion axis, $A$ of \Cref{eq:motion-alignment}, by masking mode, over the frame pairs
  whose axis is defined ($\kappa \geq \kappa_{\min}$), every pipeline pooled. Left, offline
  regime; right, real-time regime. Capture would move the unmasked distribution toward \num{1},
  away from the masked ones.}\label{fig:motion-alignment}
\end{figure}

\paragraph{Capture}
\Cref{fig:capture-scatter-offline} is the offline counterpart of \Cref{fig:capture-scatter}, and
file \suppfile{offline_masking_capture} that of \Cref{tab:realtime_masking-capture};
\Cref{fig:capture-by-offset} gives the captured share per offset bin in both regimes and
\Cref{fig:motion-alignment} the direction of the registration error.

\paragraph{Masking}
The tables behind \Cref{sec:masking-results} are released as files: for the mechanism contrasts,
\suppfile{offline_masking_mechanism} and \suppfile{realtime_masking_mechanism}; for the benefit by
stratum, \suppfile{offline_masking_benefit_strata} and \suppfile{realtime_masking_benefit_strata};
and for the mask-margin sweep, \suppfile{offline_margin_benefit}.

\FloatBarrier
\section{The analysis plan applied}
\label{sec:appendix-plan}

The rule of \Cref{sec:analysis-plan} decides what a results table bolds, and the bold is a
verdict rather than evidence. The comparisons behind it are released for every bolded column, one
row per configuration and outcome: for Part~I in file \suppfile{detectors_comparisons} and for
Part~II, one file per regime, in \suppfile{offline_masking_comparisons} and
\suppfile{realtime_masking_comparisons}; the other campaigns' files are named in their captions.
Each row gives the Hodges--Lehmann estimate of the paired difference to the best configuration, its
\SI{95}{\percent} cluster bootstrap interval, and the exact sign-flip $p$ adjusted by Holm within
the outcome. A reader who disagrees with the equivalence margin can therefore reread every
conclusion at a margin of their own, which a table of bolded means alone would not allow.

\FloatBarrier
\section{Horizontal-to-oriented box conversion}
\label{sec:appendix-obb}

The \acp{OBB} are converted from the \acp{HBB} by \texttt{hbb2obb}~\citep{hbb2obb}. This appendix
summarizes the procedure, selection and accuracy, which Songdo Vision OBB~\citep{songdo_vision_obb}
documents in full with the commands that reproduce them.

\paragraph{Procedure}
Each \ac{HBB}, its sides moved outward by a scale factor $s$ of its width and height (each
dimension grows by $2s$), prompts one or more SAM-family
models~\citep{kirillov2023sam,ravi2025sam2}, and an ensemble's masks are combined by majority vote.
The mask is cropped to the scaled prompt and refined by a signed morphological
step~\citep{soille2004morphological}, its square kernel a set fraction of the shorter side of the
largest contour's bounding rectangle: a positive kernel opens it, removing a
thin protrusion such as an overhead sign enclosed by the box, and a negative kernel closes it,
rejoining a body split by glare or an occluder. A minimum-area rotated
rectangle~\citep{freeman1975minimum} is fitted to the largest contour together with any piece
holding at least a set fraction of its area. The \ac{OBB} keeps the class of its \ac{HBB};
where no valid mask is found, the \ac{HBB} is emitted unchanged, the fallback whose
consequence is stated in \Cref{sec:masking}.

\begin{figure}[htbp]\centering
  \includegraphics[width=0.75\textwidth]{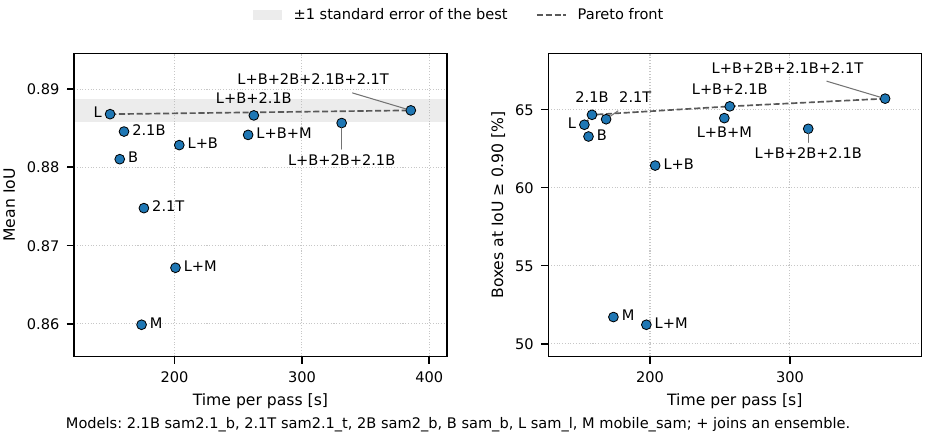}
  \caption{Accuracy against time, one point per model
configuration at its best grid point for the score shown, with the Pareto front dashed (the key
below the panels names the checkpoints; + joins an ensemble). Left, mean \ac{IoU}: the shaded band
is one standard error of the mean about the leading configuration. Right, the share of boxes at \ac{IoU} \num{0.90} or above,
which does not saturate and orders the configurations the mean cannot separate. Times are a single
pass over the \num{50} reference frames.}\label{fig:obb-pareto}
\end{figure}

\paragraph{Selection}
The ensemble, the inference resolution, the prompt scale factor and the signed kernel fraction
interact, so they were selected by grid search against the human-verified reference of
\Cref{sec:appendix-obb-gt}: eleven model configurations, each at \num{45} grid points (resolution
$\{896,\, 1024,\, 1152\}$, scale factor $\{-0.01,\, 0,\, 0.01,\, 0.02,\, 0.03\}$, kernel
$\{-0.10,\, 0,\, +0.10\}$), \num{495} conversions of the \num{50} reference frames. Mean \ac{IoU}
saturates on tightly fitted boxes: the three leading configurations lie within \num{0.44} standard
errors of one another (\Cref{tab:obb-eval,fig:obb-pareto}). Configurations within one standard
error of the leader are treated as tied, a heuristic rather than a test, and ties are broken by the share
of boxes below \ac{IoU} \num{0.75} and the \num{90}th-percentile orientation error, taking the cheapest configuration that leads on both. That
is a single \texttt{sam\_l} at resolution \num{1024}, with a scale factor of \num{0.02} and a
closing whose kernel is \num{0.1} of that side, at \SI{39}{\percent} of the inference cost of the
leading ensemble.

\paragraph{Accuracy}
On the reference, the selected configuration reaches a mean \ac{IoU} of \num{0.88679} (standard
error \num{0.00143}), a median of \num{0.91759}, \SI{92.1}{\percent} of boxes above \ac{IoU}
\num{0.75} and \SI{61.5}{\percent} above \num{0.90}, and a \num{90}th-percentile orientation error
of \SI{3.50}{\degree}, measured on the frames it was selected on, with exact prompts (\Cref{sec:appendix-obb-gt}). With
perturbed copies it loses at most \num{0.002} of mean \ac{IoU} while the expanded prompt still covers
the vehicle, \num{0.027} under a \SI{2}{px} center jitter and \num{0.043} once the prompt falls short. On the \num{1345} vehicles turned
off the image axes it raises mean \ac{IoU} over the \ac{HBB} itself from \num{0.6032} to
\num{0.8039} and improves on it for \SI{91.9}{\percent} of them; over the whole set the
\num{90}th-percentile orientation error falls from \SI{22.69}{\degree} to \SI{3.50}{\degree}.
Segmentation found a usable mask inside all \num{4245} reference boxes and all \num{274190} boxes of
the release, so neither carries a fallback.

\begin{table}[htbp]
  \centering
  \caption{\ac{OBB} conversion against the \num{4245} boxes of the human-verified reference, one
  row per model configuration at its grid point of highest mean \ac{IoU}. Models lists the segmentation checkpoints of an ensemble; Size is the inference
  resolution, Scale the prompt expansion and Kernel the morphological setting (negative closes).
  Mean \ac{IoU} is given with its standard error over boxes; configurations within one standard error of the
  leader are tied with it and set in bold. IoU$\geq$0.90 is the share of boxes at or above that
  rung, and Angle$_{90}$ the 90th percentile of the orientation error.}
  \label{tab:obb-eval}
  \small
    \begin{adjustbox}{max width=\textwidth}\begin{tabular}{lrrrrrrr}
  \toprule
  Models & Size & Scale & Kernel & Mean IoU & IoU$\geq$0.90 & Angle$_{90}$ [$^\circ$] & Time [s] \\
  \midrule
  sam\_l + sam\_b + sam2\_b + sam2.1\_b + sam2.1\_t & 1152 & 0.03 & $-$0.1 & \textbf{0.88728} $\pm$ 0.00150 & 64.1\% & 4.38 & 385.5 \\
  sam\_l & 1024 & 0.02 & $-$0.1 & \textbf{0.88679} $\pm$ 0.00143 & 61.5\% & 3.50 & 149.4 \\
  sam\_l + sam\_b + sam2.1\_b & 1024 & 0.02 & $-$0.1 & \textbf{0.88663} $\pm$ 0.00147 & 63.2\% & 3.85 & 262.2 \\
  sam\_l + sam\_b + sam2\_b + sam2.1\_b & 1024 & 0.03 & $-$0.1 & 0.88566 $\pm$ 0.00148 & 61.2\% & 3.96 & 331.2 \\
  sam2.1\_b & 1152 & 0.03 & $-$0.1 & 0.88457 $\pm$ 0.00158 & 64.0\% & 4.56 & 160.3 \\
  sam\_l + sam\_b + mobile\_sam & 1024 & 0.02 & $-$0.1 & 0.88414 $\pm$ 0.00147 & 60.9\% & 3.97 & 257.8 \\
  sam\_l + sam\_b & 896 & 0.02 & $-$0.1 & 0.88284 $\pm$ 0.00152 & 61.4\% & 4.06 & 203.7 \\
  sam\_b & 1024 & 0.02 & $-$0.1 & 0.88104 $\pm$ 0.00149 & 59.8\% & 4.18 & 156.9 \\
  sam2.1\_t & 1152 & 0.02 & $-$0.1 & 0.87478 $\pm$ 0.00171 & 63.3\% & 6.04 & 175.9 \\
  sam\_l + mobile\_sam & 1024 & 0.03 & $-$0.1 & 0.86716 $\pm$ 0.00141 & 43.4\% & 4.03 & 200.7 \\
  mobile\_sam & 1152 & 0.02 & $-$0.1 & 0.85988 $\pm$ 0.00149 & 41.9\% & 5.63 & 174.1 \\
  \bottomrule
\end{tabular}\end{adjustbox}
\end{table}

\subsection{The human-verified reference}
\label{sec:appendix-obb-gt}

The reference is \num{4245} \acp{OBB} over \num{50} frames from six recordings. A detector
supplied the \acp{HBB}; one annotator went over every frame and redrew as \acp{OBB} the
\num{1345} vehicles not square to the image, leaving the other \num{2900} on their detected geometry,
and the set was restricted to the four vehicle classes, rectified to true rectangles, and
reconciled one-to-one with the \acp{HBB} that prompt the conversion, each the exact axis-aligned
envelope of its \ac{OBB}. The unrevised boxes are about \SI{5.8}{\percent} wider across the
vehicle than hand-drawn ones, so on the axis-aligned majority mean \ac{IoU} mixes reference slack with
conversion error. An independent
DOTA-pretrained oriented detector (\texttt{YOLO11x-OBB}~\citep{yolo11_ultralytics}, run tiled and
scored class-agnostically), which shares neither architecture nor training data with the conversion,
places the axes of these vehicles within a median \SI{0.82}{\degree} of the reference boxes, the scale
against which the conversion's orientation error is read.

\end{document}